%% file: Paper.tex
\RequirePackage{fix-cm}                                                                                                                         
\documentclass[smallextended]{svjour3}       
\pdfoutput=1 
\usepackage[backref]{hyperref}
\usepackage[dvipsnames]{xcolor}
\usepackage{algorithmic}
\usepackage{amsmath,amssymb,amsfonts}
\usepackage{bm}
\usepackage{booktabs}
\usepackage{caption}
\usepackage{cite}
\usepackage{color}
\usepackage{diagbox}
\usepackage{float}
\usepackage{graphicx}
\usepackage{longtable}
\usepackage{lscape}
\usepackage{multirow}
\usepackage[figuresright]{rotating}
\usepackage{stfloats}
\usepackage{subfigure}
\usepackage{textcomp}
\usepackage{verbatim}
\usepackage{makecell}
\usepackage{longtable}
\usepackage{verbatim}
\usepackage{enumitem}

\smartqed  

\begin{document}

\title{A State-of-the-art Survey of Object Detection Techniques in Microorganism Image Analysis: From Classical Methods to Deep Learning Approaches}


\author{Pingli Ma, Chen Li, Md Mamunur Rahaman, Yudong Yao, Jiawei Zhang, Shuojia Zou, 
Xin Zhao, Marcin Grzegorzek}


\institute{Pingli Ma, Chen Li, Md Mamunur Rahaman, Jiawei Zhang, Shuojia Zou and Xin Zhao \at
              Microscopic Image and Medical Image Analysis Group, 
              Medicine and Biological Information Engineering College, Northeastern University, China \\
            Corresponding author: Chen Li, \email{lichen201096@hotmail.com}           
           \and
           Yudong Yao \at
              Department of Electrical and Computer Engineering, 
              Stevens Institute of Technology, USA
              \and Marcin Grzegorzek \at
              Biomedical Information College, University of Luebeck, Germany
}

\date{Received: date / Accepted: date}

\titlerunning{A Survey for Microorganism Image Detection Method}
\authorrunning{Pingli Ma, et.al.}
\maketitle

\begin{abstract}
Microorganisms play a vital role in human life. Therefore, microorganism detection is of great significance to human beings. However, the traditional manual microscopic detection methods have the disadvantages of long detection cycle, low detection accuracy in large orders, and great difficulty in detecting uncommon microorganisms. Therefore, it is meaningful to apply computer image analysis technology to the field of microorganism detection. Computer image analysis can realize high-precision and high-efficiency detection of microorganisms. In this review, first,we analyse the existing microorganism detection methods in chronological order, from traditional image processing and traditional machine learning to deep learning methods. Then, we analyze and summarize these existing methods and introduce some potential methods, including visual transformers. In the end, the future development direction and challenges of microorganism detection are discussed. In general, we have summarized 142 related technical papers from 1985 to the present. This review will help researchers have a more comprehensive understanding of the development process, research status, and future trends in the field of microorganism detection and provide a reference for researchers in other fields.

\keywords{Microorganisms Images \and Microscopic Images \and Image Analysis \and Object Detection \and Machine Learning \and  Visual Transformer} 
\end{abstract}

\input{Introduction}

\input{Overview}

\input{Classical}

\input{Traditional}
\input{Deep}

\input{Methodology}
\input{Conclusion}

\section*{Acknowledgements}
This work is supported by the ``National Natural Science Foundation of China'' (No.61806047). We also thank Miss Zixian Li and Mr. Guoxian Li for their important discussion.

\section*{Conflict of Interest}
The authors declare that they have no conflict of interest.

\bibliographystyle{unsrt} 
\bibliography{Reference} 

\end{document}

%% file: Introduction.tex
\section{Introduction}
\label{Sect:1}
\subsection{Knowledge of Microorganism}
Microorganisms have a unique ability to adapt to extreme conditions. They are found 
in every environment imaginable. Microorganisms refer to tiny organisms with independent 
living functions, which means they can absorb energy, grow and reproduce by themselves. 
The modern classification of microorganisms are in a three-domain system. The system 
consists of Archaea, Eucarya, and Bacteria~\cite{Pepper-2011-EM}. Bacteria, along 
with actinomycetes and cyanobacteria (blue-green algae) belongs to the prokaryotes. 
Eukaryotes or eukarya includes fungi protozoa, algae, plant, and tiny animal. Viruses 
are obligate intracellular parasites that belong to neither of these two 
groups~\cite{Bitton-2005-WM}. Microorganisms play a very important role in human life, 
for better or worse. For example, the presence of plant growth-promoting rhizobacteria 
can encourage beneficial effects on plant health and growth. It can also suppress 
disease-causing microbes~\cite{Babalola-2010-BBOA}. Deleterious rhizobacteria can 
inhibit plants growth through the production of phytotoxins~\cite{Nehl-1997-DRB}. 
Lactic acid bacteria can affect humans in many aspects, such as the regulation of 
gastrointestinal flora, the normal operation of human metabolism, and the inhibition 
of the reproduction of harmful bacteria~\cite{Masood-2011-BEO}. Severe acute respiratory 
syndrome coronavirus 2 (SARS-CoV-2) can cause fever, discomfort, dry cough, shortness 
of breath and even death in severe case~\cite{Hui-2020-TC2}. In summary, the work of 
microorganism analysis is of great significance for the human being.

Observation under a microscope is a common and important method in microorganism analysis. 
A stereo scan electron microscopy uses microorganism in soil analysis~\cite{Gray-1967-SEM}. 
In~\cite{Daley-1975-DCO}, a modified epifluorscence technique based on a microscope is 
used for aquatic bacteria counting. In~\cite{Collins-1993-AOE}, environmental scanning 
electron microscopy is used for microorganism analysis. However, those microscopic methods 
have some disadvantages. Firstly, there are many types of microorganisms. Estimated 
in~\cite{Locey-2016-SLP}, Earth is inhabited by $10^{11}$-$10^{12}$ microbial species. 
Thus the knowledge of experts is always inadequate. When experts use this method for 
microorganism analysis, they often have to consult a great deal of literature. Secondly, 
the training cycle for researchers and overall detection time need much time. For example, 
the counting of phytoplankton with traditional microscopic method requires much time. 
Besides, an operator needs to master a great deal of professional 
knowledge~\cite{Embleton-2003-ACO}. Thirdly, the microorganisms to be analyzed are often 
of large orders of magnitude. It is difficult for microscopic methods to deal with the 
analysis problem with a large amount of data. Large sample size will affect the analysis 
accuracy of operators~\cite{Van-2002-OTC}. Because of these disadvantages of the 
microscopic method, we need a more efficient method for microorganism analysis. For 
example, computer image analysis is a feasible method.

\subsection{Motivation}
Computer image analysis is part of both computer vision and image processing. Typically, 
image analysis is used for gaining insight into the raw image and extracting the 
information we need~\cite{Umbaugh-2005-CID}. 
In terms of microorganism analysis, using image analysis has many advantages over the 
microscopic method. Firstly, image analysis does not care about the number of microbial 
species. Computers can store and remember more information about different kinds of 
microorganisms than experts. The Ribosomal Database Project stores 2460 different species. 
As information is entered, the number will continue to increase~\cite{Maidak-2000-TRD}. 
Secondly, image analysis takes less time. With the help of image analysis, 
phytoplankton analysis just needs 30 to 40 minutes~\cite{Embleton-2003-ACO}. Thirdly, 
for image analysis, the task of microorganism analysis of large orders of magnitude 
does not pose any difficulties. Fourthly, the operation of image analysis is simple. 
When using image analysis, necessary operations usually only include import data and 
a few simple steps, such as the Center for Microbial Ecology Image Analysis 
System~\cite{Dazzo-2015-UOC}. Based on the above reasons, it is feasible to apply 
computer-aided image analysis in microorganism analysis.

At present, many people are conducting researches in this field, such as microorganism 
segmentation~\cite{Kulwa-2019-ASO}, microorganism clustering~\cite{Li-2020-AROC}, 
microorganism classification~\cite{Li-2019-ASF}, microorganism counting~\cite{Li-2021-ACR}, 
etc. However, there is no specific review of microorganism detection. To have a 
comprehensive overview of the existing microorganism detection research, we discuss many 
materials. According to our search, some reviews involve microorganism detection. 
In~\cite{Benfield-2007-RRO}, the development of plankton-imaging systems and advances in 
extracting information from image data sets timely are summarized, where twelve papers 
involve microorganism detection among all 56 references. In~\cite{Schaap-2012-LOA}, 
a review of the current status of lab on a chip technologies in the context of algae 
detection and monitoring is presented. In this work, there are twenty four papers related to algae detection mentioned among all 81 references. In \cite{Gopinath-2014-BDF}, a review for existing bacterial detection methods is proposed, where methods from manual microscope detection to smartphone-based detection are analysed. Twelve papers in this survey are related to bacterial detection among all 126 references. One review focuses on using computer image analysis to examine the microscopic objects \cite{Puchkov-2016-IAI}. In this work, twenty-two papers are related to microorganism detection among all 102 references. In \cite{Li-2019-ASF}, the development history of microorganism classification using content-based microscopic image analysis approaches is reviewed, where twenty-four papers relate to microorganism detection among all 317 references. In \cite{Zhou-2020-RAI}, the development of diatom testing over the decades is reviewed and a new method of deep learning for diatom detection is discussed. In this work, two papers are related to diatom detection among all 34 references. Hence, none of these surveys conducts a comprehensive study on object detection in microorganism image analysis. In fact, the analysis of microorganism detection methods can facilitate development in other microorganism image analysis fields, such as microorganism image classification \cite{Zhao-2021-ACO,Kulwa-2021-ANP,Li-2021-EEM,Xu-2020-AEF,Li-2019-ASF,Kosov-2018-EMC,Li-2016-EMA,Li-2015-AOC}, microorganism image segmentation \cite{Zhang-2021-ACR,Zhang-2021-LAN,Zhang-2020-AMC,Kulwa-2019-ASS} and microorganism image retrieval \cite{Zou-2016-EMI,Zou-2017-CIR}.

\begin{figure}[http!]
\centering
\includegraphics[width=0.95\linewidth]{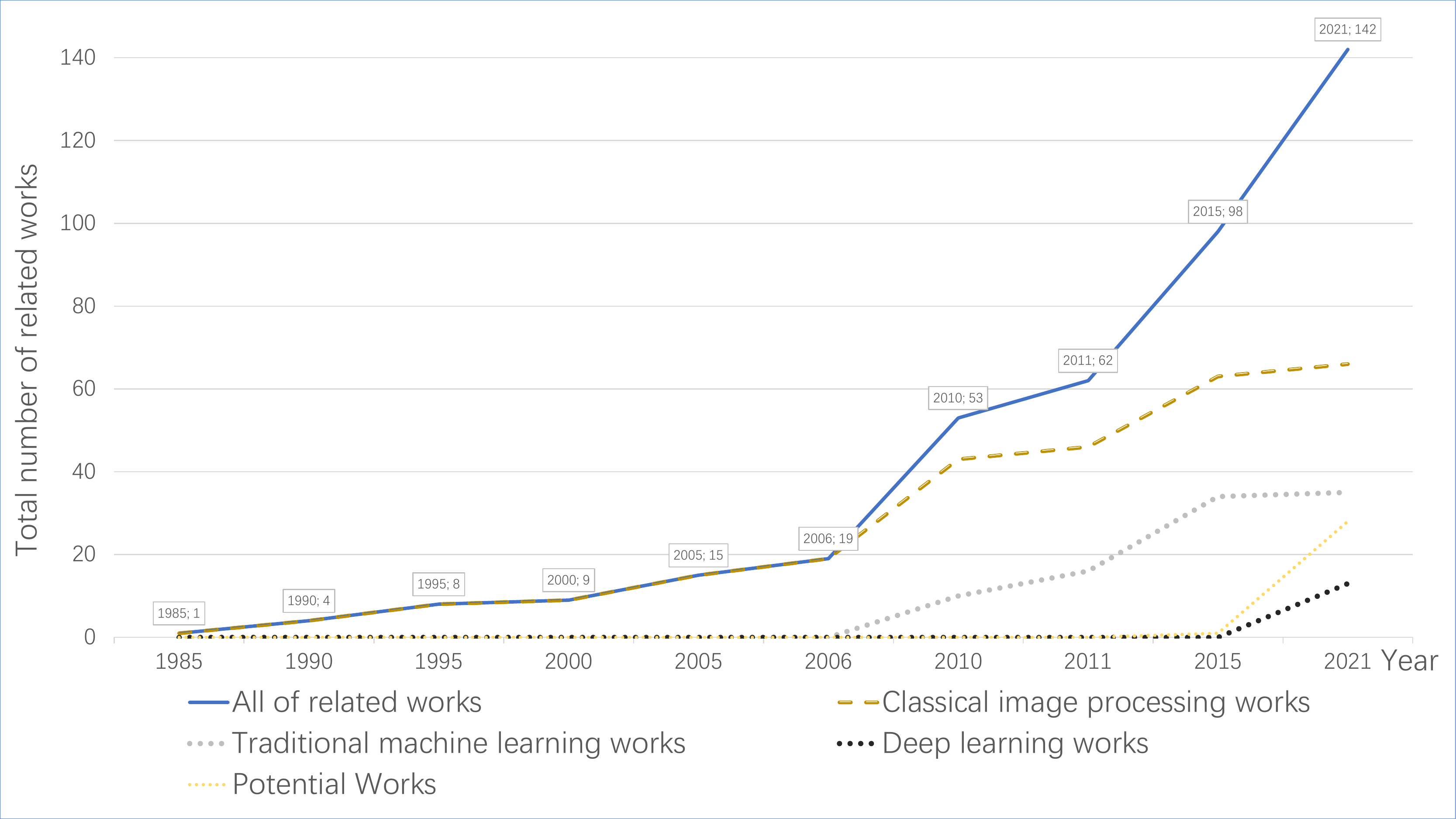}
\caption{The total number of related works on microorganism detection.}
\label{fig:1} 
\end{figure}

\begin{figure}[http!]
\centering
\includegraphics[width=1\linewidth]{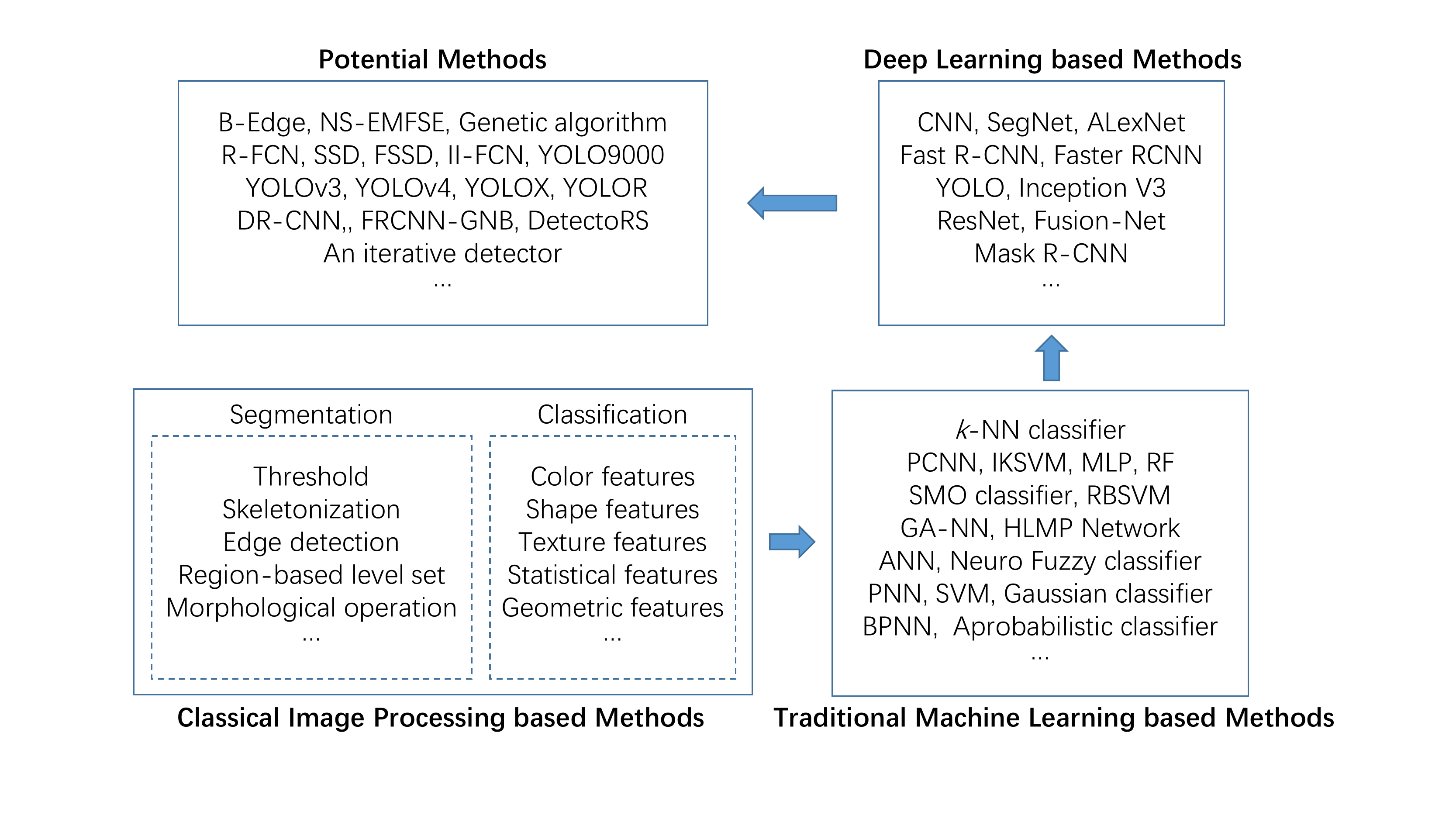}
\caption{Microorganism image detection methods mentioned in this survey.}
\label{fig:methods} 
\end{figure}

As shown in Fig.\ref{fig:1}, for microorganism detection, the early work is mainly based on classsical image processing methods, such as image segmentation~\cite{Fukuda-1989-ESD} and binarization~\cite{Bloem-1995-FAD}. With the development of the field of machine learning, researches used traditional machine learning methods appear gradually, such as artificial neural network (ANN) \cite{Widmer-2005-UOA}, support vector machine (SVM)~\cite{Lenseigne-2007-SVM} and genetic algorithm-neural network (GA-NN)~\cite{Osman-2010-AGA}. In recent years, deep learning-based methods become very popular, such as faster region-convolutional neural networks (Faster R-CNN) \cite{Viet-2019-PWE} and Mask R-CNN~\cite{Ruiz-2020-SVI}. In addition, many new methods can be applied in microorganism detection, such as single shot detector (SSD) \cite{Liu-2016-SSM}. Moreover, visual transformer-based methods have a more robust global information representation capability than the current famous CNNs, which means they can eliminate the problem of microorganism structure described in the complete image. Therefore, the field of visual transformer has developed rapidly in recent years. In fact, visual transformer has been applied to cell detection with satisfactory results~\cite{Aubreville-2017-AGS}. However, visual transformer have not yet been applied to microorganism detection, although   visual transformer based methods show great potential for application in microorganism detection, such as squeezeand-excitation (SE)~\cite{Hu-2018-SN}, Vision Transformer-Faster RCNN (ViT-FRCNN) \cite{Beal-2020-TTO}.
The number of related works is rising steadily. The data in Fig.~\ref{fig:1} shows that microorganism object detection has a good development trend and a great development potential. Thus we decide writing a survey of object detection technologies for microorganism image analysis, where we summarize about 142 related works for a comprehensive survey. Methods mentioned in these works are mainly grouped into classical image processing based methods, traditional machine learning based methods, deep learning based methods and potential methods, as shown in Fig.~\ref{fig:methods}. 

\subsection{Paper Retrieval and Screening}
We specifically searched for papers in the field of microbiology. Therefore, according to the definition of microorganisms (tiny organisms with independent living), some excellent articles are not analyzed and discussed in this article due to the non-compliant research fields, such as \cite{Aubreville-2017-AGS} for histology cell, \cite{Chen-2021-GAM} for gastric histopathology image, \cite{Liu-2021-ITA} for cytopathology cell and \cite{Hu-2022-GAN} for gastric histopathology image.

\begin{figure}[htbp!]
\centering
\includegraphics[width=1\linewidth]{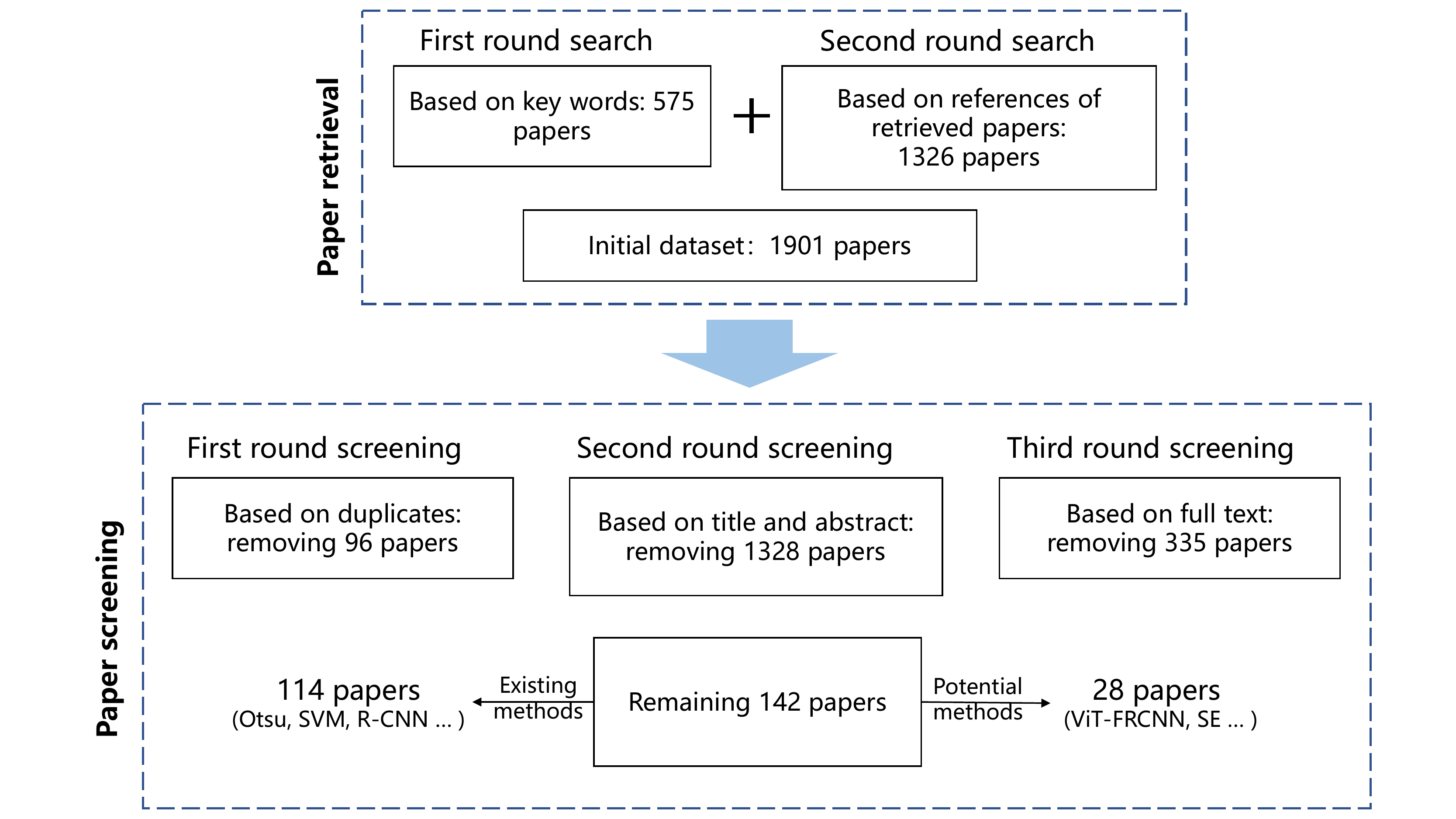}
\caption{The retrieval and screening flow of involved papers.}
\label{fig:0} 
\end{figure}

The retrieval and screening flow of papers involved in this review is shown in Fig.\ref{fig:0}. In retrieving stage, we first search for random combinations of the following phrases in Google Scholar: microbe, detection, recognition, classification, image analysis, image processing, where 575 papers are collected. Then, based on the references of collected papers, we accumulated extra 1326 papers. Therefore, a total 1901 papers are obtained in the retrieving stage. In screening stage, we mainly removing papers by three steps. Firstly, we screen paper by checking whether the paper is duplicated. 96 papers are removed. Among retained papers, 1325 papers' abstract or title are not related to our review and then removed. At last, 335 of remaining papers are removed after reading through them carefully. In brief, 142 papers are remained and analysed in our review. Among them, 65 papers are related to classical image processing; 34 papers are related to traditional machine learning; 13 papers are related to deep learning; one is related to both classical image processing and traditional machine learning; one is related to both traditional machine learning and deep learning; 28 papers are related to potential methods, including visual transformer-based methods.

\subsection{Structure of This Review}
In this paper, microorganism detection is summarized according to different detection methods. Firstly, the existing microorganism detection methods are classified. Then, each class is described and analyzed respectively. In addition, the research motivation, basic knowledge of object detection and evaluation criteria of different methods are introduced. For some critical methods, flow charts or sample diagrams are shown to deepen understanding.

The rest of this review is structured as follows: In Sect.\ref{Sect:2}, we have introduced the basic knowledge of object detection and some commonly used evaluation indicators. Then, in Sect.\ref{Sect:3}, Sect.\ref{Sect:4} and Sect.\ref{Sect:5} we have introduced classical image processing-based methods, traditional machine learning-based methods and deep learning-based methods,respectively. After, the three categories of methods mentioned above are summarized in Sect.\ref{Sect:6}. In addition, potential methods are analysed. Finally, Sect.\ref{Sect:7} summarizes the whole work with potential future work. With the structure, relevant researchers can have a better understand of the related works.

%% file: Overview.tex
\section{Overview of Microorganism Detection}
\label{Sect:2}
Prior to an overview of microorganism detection approaches, we review the basic knowledge of object detection. Next, depending on the form of the test results  detection results, the history of microorganism detection is grouped into the early phase and current phase. We will outline the early and current phases of microorganism detection development in chronological order.

\subsection{Basic Knowledge of Object Detection} 
\label{Sect:2.1}
\begin{figure}[htbp!]
\centering
\includegraphics[width=0.95\linewidth]{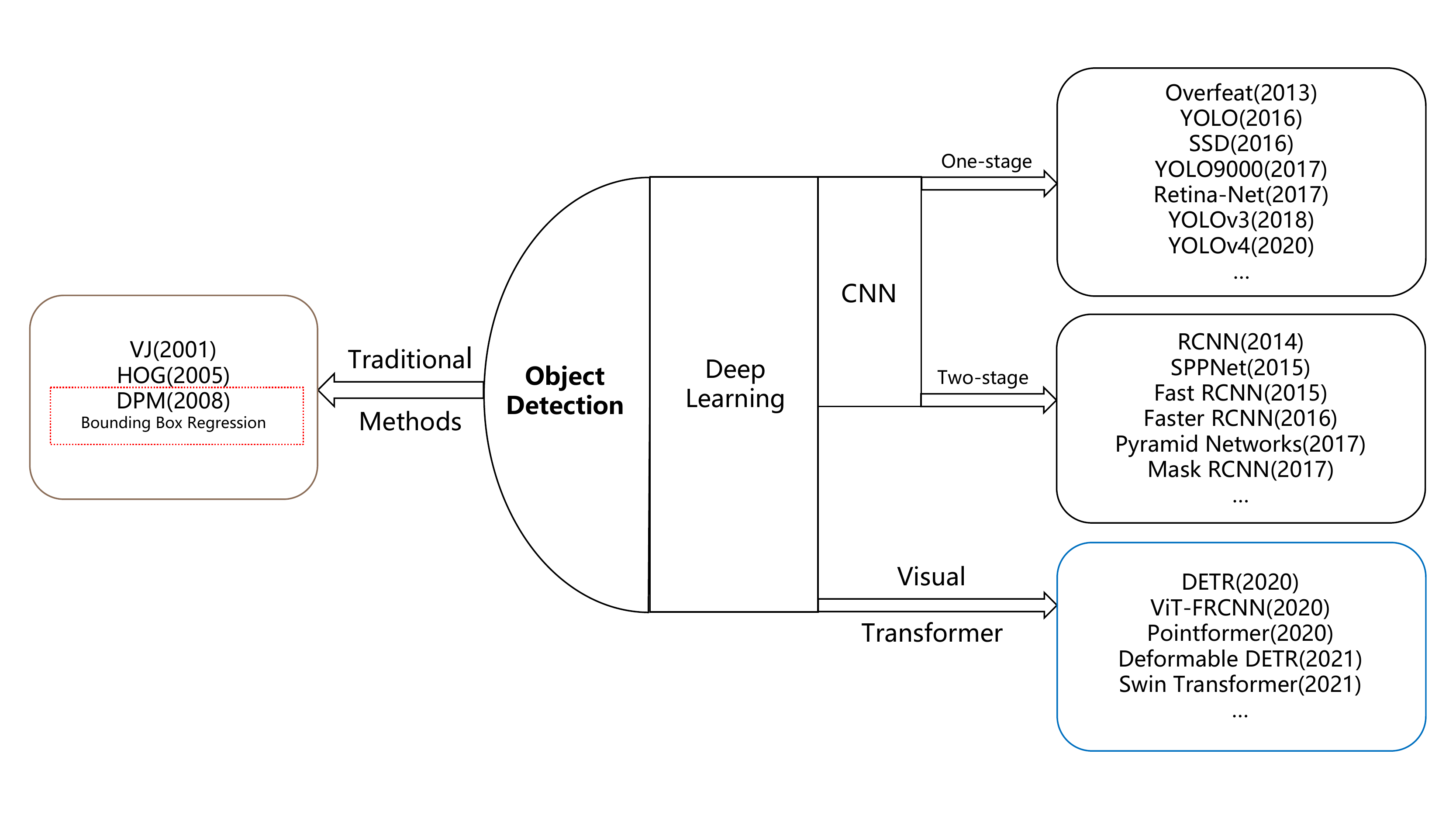}
\caption{The development of detection methods over the past two decades.}
\label{fig:2} 
\end{figure}

Object detection is a long-term and challenging task in computer vision \cite{Fischler-1973-TRA}. As one of the primary tasks of computer vision, the ultimate goal of object detection is to give the classes and locations of objects. Therefore, object detection is the basis for many other computer vision tasks, such as object tracking \cite{Son-2017-MTW} and scene understanding \cite{Pan-2017-SAD}. Detection methods are evolving all the time since object detection is proposed. In this section, we comprehensively summarize the development of object detection methods over the past twenty years and group object detection methods into traditional manual features-based methods and deep learning-based methods \cite{Zou-2019-ODI}. Among them, deep learning-based methods include one-stage methods and two-stage methods according to different processing flows. Fig.\ref{fig:2} shows the development of detection methods over the past two decades. Methods involved in Fig.\ref{fig:2} are as follows: Viola-Jones(VJ) \cite{Viola-2001-ROD}, Histograms of Oriented Griented gradients(HOG) \cite{Dalal-2005-HOO}, Deformable Parts Model (DPM) \cite{Felzenszwalb-2008-ADT}, R-CNN \cite{Girshick-2014-RFH}, Spatial Pyramid Pooling Networks(SPPNet) \cite{He-2015-SPP}, Fast R-CNN \cite{Girshick-2015-FR}, Faster R-CNN\cite{Ren-2016-FRT}, Pyramid Networks \cite{Lin-2017-FPN}, Mask R-CNN \cite{He-2017-MR}, OverFeat \cite{Sermanet-2013-OIR}, You Only Look Once (YOLO) \cite{Redmon-2016-YOLO}, Single Shot MultiBox Detector (SSD) \cite{Liu-2016-SSS}, YOLO9000 \cite{Redmon-2017-YBF}, Retina-Net \cite{Lin-2017-FLF}, YOLOv3 \cite{Redmon-2018-YAI}, YOLOv4 \cite{Bochkovskiy-2020-YOS}, Vision Transformer-Faster RCNN (ViT-FRCNN) \cite{Beal-2020-TTO}, DEtection TRansformer (DETR) \cite{Carion-2020-EOD}, Pointformer \cite{Pan-2020-3OD}, Deformable DETR \cite{Zhu-2021-DDD} and Swin Transformer \cite{Liu-2021-STH}.

\subsection{The Early Phase of Microorganism Detection}
\begin{figure}[htbp!]
\centering
\includegraphics[width=0.95\linewidth]{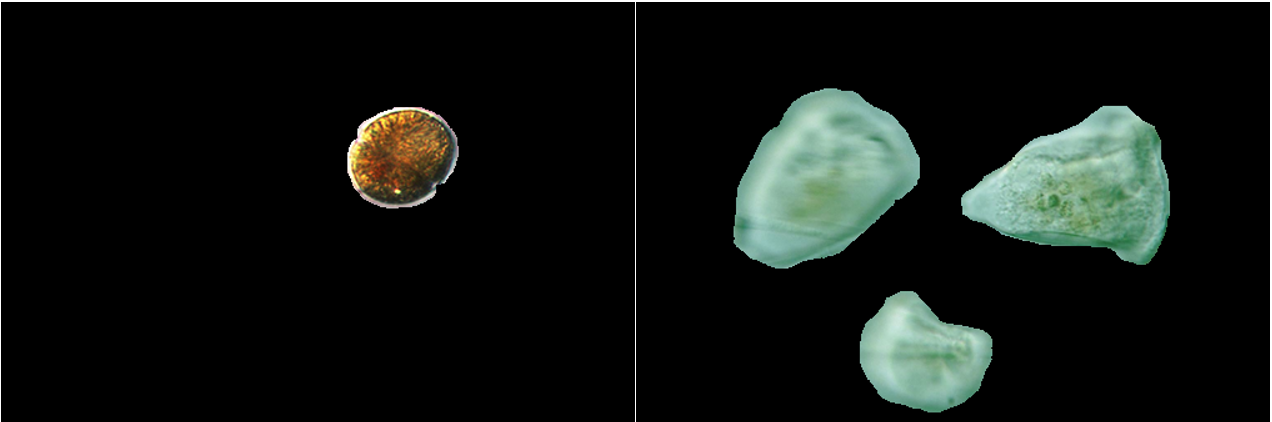}
\caption{Examples of segmentation-based detection.}
\label{fig:3} 
\end{figure}

In the early phase of microorganism detection, the primary purpose is to determine whether the object exists or not. At this phase, microorganism detection is closely related to microorganism segmentation. Fig.\ref{fig:3} shows an example of segmentation-based detection. In \cite{Bloem-1995-FAD}, the detection of soil bacterium is based on the combination of segmented binary and grayscale. In \cite{Rachna-2013-DOT}, $k$-means clustering and Otsu threshold are respectively employed to segment mycobacterium tuberculosis images for mycobacterium tuberculosis detection. In addition, microorganism detection is also closely related to classification. In \cite{Lenseigne-2007-SVM}, the automatic detection of mycobacterium tuberculosis is based on SVM classifier. In \cite{Verikas-2014-AIA}, SVM as well as random forest (RF) classifiers is used for distinguishing prorocentrum minimum (P. minimum) cells and other objects. 

\subsection{The Current Phase of Microorganism Detection}
\begin{figure}[htbp!]
\centering
\includegraphics[width=0.95\linewidth]{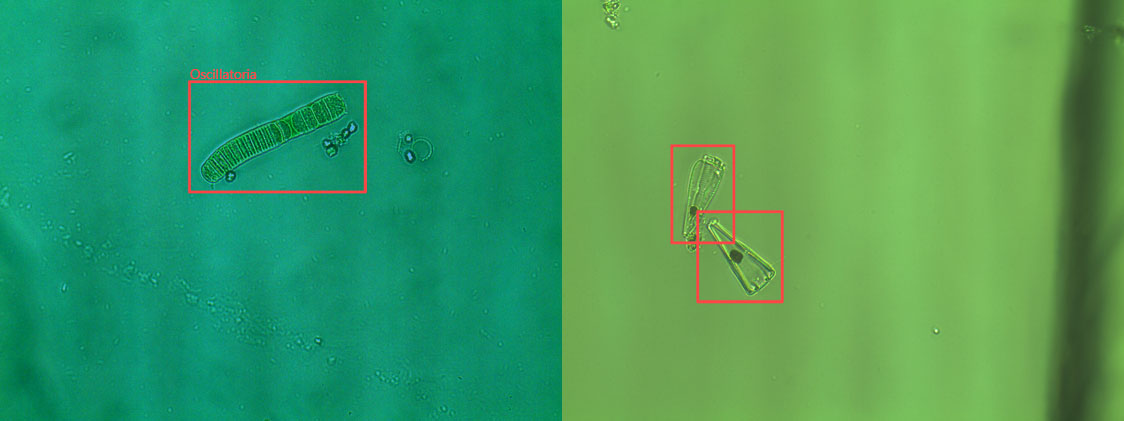}
\caption{Example of deep learning-based detection.}
\label{fig:4} 
\end{figure}

Unlike the early phase of microorganism detection, the current phase of microorganism detection precisely gives the classes and locations of objects. At this stage, the detection results include several bounding boxes surrounding each detected objects with respective class labels. Microorganism detection at this stage is mainly based on deep learning methods \cite{Lecun-2015-DL}. An example of a deep learning-based detection result is shown in Fig.\ref{fig:4}. 
\par
Deep learning has a long history of development. In 1943, the emergence of McCulloch-Pitts (MCP) model, which is also regarded as original version of the artificial neural model (ANN), means the initial realization of deep learning \cite{Mcculloch-1943-ALC}. After MCP, deep learning field has a long period of stable development until the problem caused by hardware limitations is resolved. After 2014, thanks to the powerful computing power of graphics processing unit (GPU) and implementation of CNN models, deep learning has been able to develop rapidly, especially in object detection task. In fact, Faster R-CNN, a typical model of two-stage object detector not proposed until 2016. As for YOLO, a typical model of one-stage object detector, it is proposed even five months later than Faster R-CNN. It is only after the relative maturity of deep learning techniques that it is applied to microorganism detection. Therefore, while deep learning is widely used for object detection, few research studies are based on deep learning for microorganism detection. However, it can be seen from Fig.\ref{fig:1} that deep learning in the field of microorganism detection has developed rapidly, although it started late. After consulting a lot of materials, we have found  13 researches that are related to deep learning. In \cite{Hung-2017-AFR}, Faster R-CNN is firstly applied for cell detection in brightfield microscopy images of malaria-infected blood. In \cite{Ruiz-2020-SVI}, Mask R-CNN is for the first time used for detecting diatoms in images, where include many diatom shells.

\subsection{Evaluation Criteria}
In order to evaluate the effectiveness of microorganism detectors, some proper criteria are given in different related work. It is observed that there is no widely accepted evaluation criteria for evaluating detection performance. Different criteria are applied in different research scenarios. For detection work in counting, the main criterion is the correlation with the expert counting results or real number, such as \cite{Ogawa-2005-MDI}, \cite{Hung-2017-AFR}. When using this kind of criterion, the expert counting results or actual number is regarded as the baseline. Therefore, the closer the detection results are to the reference baseline, the better. In \cite{Jan-2015-DOT}, the effectiveness of detection result for tubercle bacillus(TB) is evaluated by the number of objects in sample image and the number of detected objects, shown in Equation \ref{equ:1}.
\begin{equation}
Accuracy=\rm \frac{No.objects\ in\ sample\ image}{No.detected\ TB\ bacteria}\
\label{equ:1}
\end{equation}

\begin{figure}[http]
\centering
\includegraphics[width=0.75\linewidth]{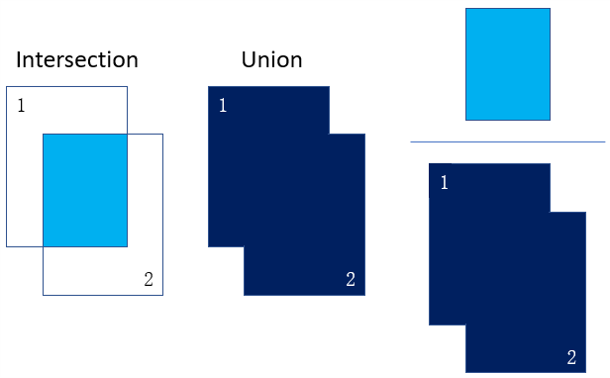}
\caption{Illustration of the calculation of IoU.}
\label{fig:IoU} 
\end{figure}

For detection work in classification, accuracy, sensitivity, specificity and precision are widely used as the criteria. To better explain how to calculate these evaluation criteria, we illustrate four fundamental concepts in Tab.\ref{tab:1}. Positive and negative represent the judgement results of the model. True and false indicate whether the model's decision is correct or not. By combining positive, negative, true and false, we can get true positive (TP), false positive (FP), true negative (TN) and false negative (FN). In addition, to judge whether the actual detection result belongs to TP, FP, TN or FN, not only the detection category but also the Intersection over Union (IoU) must be considered. In practice, IoU is used to evaluate the distance between the output box and the ground-truth. The calculation method of IoU is shown in Fig.\ref{fig:IoU}, where box 1 and box 2 respectively represent real box and prediction result, the light blue area and the dark blue area represent the intersection and union of the predicted box and the actual box in turn. Finally, IoU is calculated by intersection ratio union.  
\begin{table}[http!]
\centering
\caption{Explanatory table of TP, FN, FP, and TN.}
\begin{tabular}{|c|c|c|c|}
\hline
\multicolumn{2}{|c|}{\multirow{2}{*}{}}                                      & \multicolumn{2}{c|}{Predicted Label} \\ \cline{3-4} 
\multicolumn{2}{|c|}{}                                                       & Yes               & No               \\ \hline
\multirow{2}{*}{\begin{tabular}[c]{@{}c@{}}True \\ Label\end{tabular}} & Yes & TP                & FN               \\ \cline{2-4} 
                                                                       & No  & FP                & TN               \\ \hline
                                                                       
\end{tabular}
\label{tab:1}
\end{table}

Based on the four criteria in Tab.\ref{tab:1}, the calculation methods of accuracy, sensitivity, specificity and precision are shown in Tab.\ref{tab:2}. Some studies only used accuracy \cite{Hiremath-2010-DIA} or sensitivity \cite{Verikas-2012-AIA} as evaluation metrics. However, few works considered multiple of them, such as sensitivity and specificity \cite{Shah-2016-ADA}, and accuracy, sensitivity and specificity \cite{Osman-2010-DOM}.
\begin{table}[http!]
\centering
\caption{Formula of accuracy, sensitivity, precision and specificity.}
\resizebox{\textwidth}{9mm}{
\begin{tabular}{@{}ccccc@{}}
\toprule
        & Accuracy                                                     & Sensitivity                                        & Specificity                                        & Precision                                          \\ \midrule
Formula & \begin{tabular}[c]{@{}c@{}}$\frac{TP+TN}{TP+TN+FP+FP}$\end{tabular} & \begin{tabular}[c]{@{}c@{}}$\frac{TP}{TP+FN}$\end{tabular} & \begin{tabular}[c]{@{}c@{}}$\frac{TN}{FP+TN}$\end{tabular} & \begin{tabular}[c]{@{}c@{}}$\frac{TP}{TP+FP}$\end{tabular} \\ \bottomrule
\end{tabular}}
\label{tab:2}
\end{table}

In addition to the common criteria mentioned above, F-score, mean average precision(mAP), true accept rate (TAR), false accept rate (FAR), false reject rate (FRR) and detection accuracy are also used. Tab.\ref{tab:3} shows the mathematical expression of them. For the calculation of F-score, recall is the same as sensitive mentioned before. For mAP, AP means the mean of precision-values for each class and $m$ means how many classes detected. The value of AP is determined by the precision-sensitivity curve. 
\begin{table}[http!]
\centering
\caption{Formula of mAP, TAR, FAR , FPR and F-score.}
\resizebox{\textwidth}{9mm}{
\begin{tabular}{@{}cccccc@{}}
\toprule
        & mAP & TAR & FAR & FRR & F-score \\ \midrule
Formula &\(\frac{1}{n}*\sum\limits_{i=1}^{n}AP_{i}\)& \(\frac{TA}{TA+FA+FR}\) & \(\frac{FA}{TA+FA+FR}\) & \(\frac{FR}{TA+FA+FR}\) & \(\left ( 1+\beta ^{2} \right )*\frac{Precision*Recall}{\beta ^{2}*Precision+Recall}\) \\ \bottomrule
\end{tabular}}
\label{tab:3}
\end{table}

\subsection{Summary}
Microorganism detection methods have changed dramatically over the decades. In terms of expressions, microorganism detection develops from simple counting to the realization of localization and classification.
In implementation techniques, microorganism detection develops from direct observation with a microscope to more efficient methods using modern image processing techniques. With the continuous improvement of detection methods, the corresponding evaluation criteria are also designed.

%% file: Classical.tex
\section{Classical Image Processing Based Methods}
\label{Sect:3}
In this section, classical image processing based methods for microorganism detection are grouped into segmentation based methods and classification based methods respectively introduced in Sect.\ref{Sect:3-1}, Sect.\ref{Sect:3-2}. After that, a concise table is prepared in Sect.\ref{Sect:3-3}. In Tab.\ref{tab:4}, the publication date, literature links, research objects and its domains, main methods and evaluation indicators of related works are presented.
\subsection{Segmentation Based Methods}
\label{Sect:3-1}
In this subsection, related works that used segmentation methods for microorganism detection are introduced.
\par
In \cite{Sieracki-1985-DEA}, for alleviating subjectivity of manual counting, a novel method based on a epifluorescence microscope and image analyzer is proposed. The first step is image enhancement, including setting small bright dots next to each detected object and bright enhancement covering the entire detected area of each object. The second step is using a light pen manually for distinguishing the object from other regions. The last step is object counting with the general analysis software provided by Artek Systems Corp. The effectiveness of proposed system is analysed by comparing counts of image analysis with visual counts. Experiment result shows that the proposed system can achieve the same accuracy with visual counts in bacterial detection. 
\par
In order to analyze the characterization of mycelial morphology, in \cite{Adams-1988-TUO}, a semi-automated image analysis is introduced. The first step is preprocessing by a mean operator and an edge enhancement operator for denoising and sharpening the image. The second step is segmentation and skeletonization. After that, the light pen is used to identify the branches manually. Finally the length of the main hyphae is determine based on the software supplied with the Magiscan 2A. The experiment is performed on five samples. The result shows that compared with the traditional digitizing-table method, semi-automatic image analysis has the advantages of short time-consuming and high convenience.
\par
In \cite{Packer-1990-MMO}, a fully automatic system is developed for morphological measurements on filamentous microorganisms, in which speed is gained. The first step is to binarize the image based on the preset gray value. The second step is eliminating objects in processed image other than microorganisms such as dust, media particles by a circularity test. Each objects in the resulting image is then skeletonized. Subsequently, the processed binary image is divided into an image containing measurable microorganisms and an image containing clumped (or aggregated) material. Finally, the processed image are measured. Experiment is performed on eight samples of about 100 microorganisms. The result shows that the fully automatic method is marginally faster than the manual image analysis method and much faster than the digitizing method.
\par
In \cite{Masuko-1991-ANM}, a novel anultra-high-sensitivity TV camera based method for enumeration of bacteria is proposed. The first step is using a anultra-high-sensitivity TV camera for enumerating small objects that emit light in a photon-counting image. After colony growth, a luminous image of the membrane filter is obtained. Finally, the accuracy of counting is judged by comparing the photon-counting image with the luminous image of the membrane filter. Experiment result shows that the for proposed algorithm is reliable for detecting single bacteria with light.
\par
In \cite{Bloem-1995-FAD}, for counting cell numbers and calculating some geometric parameters of bacteria in soil smears, a fully automatic image processing method is proposed. Firstly, convolution, morphological erosion and dilation to remove noise are used for smoothing grey images. Holes in the image are then filled for background equalization. By applying two top hat transforms, the background is eliminated. At last, an image sharpening operation is done for better detention. Particles that have a higher threshold than a fixed one are detected. In addition, the cells number is determined by number of detected particles with maximum gray scale. The proposed method has high efficiency for detecting about 1500 cells within 30 min. In addition, the detection result of proposed method is similar to the manual detection result.
\par
In \cite{Baillieul-1998-ODO}, an image analysis system is described for measuring the average velocity of many simultaneously moving similar objects. First, the operator manually locates an active window on the image to focus on the area of interest. The second step is setting the object to one and the background to zero based on a fixed threshold value. At last, by collecting the trajectories in real-time over a series of frames, average velocity of the objects are calculated. Three groups of 25 daphnids are tested on this experiment. Result shows that the system is a useful tool for the objective quantification of velocity.
\par
\begin{figure}[htbp]
 \centering
 \subfigure[Germs detection]{
  \label{fig18-a}
  \rotatebox{90}{\includegraphics[width=4cm,height=5cm]{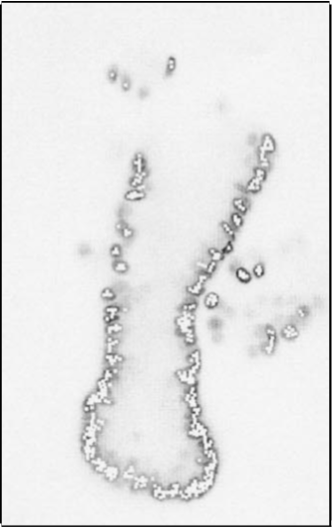}}}
 \quad
 \quad
 \subfigure[Nuclei labelling]{
  \label{fig18-b}
  \rotatebox{90}{\includegraphics[width=4cm,height=5cm]{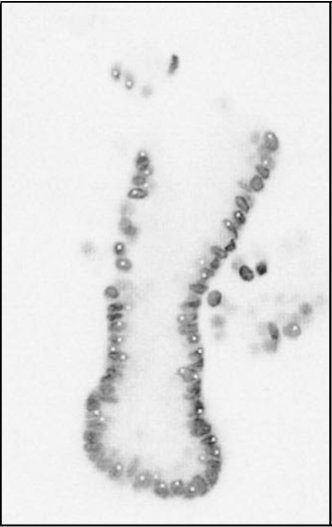}}}

 \subfigure[Volume rending of the labelling results]{
  \label{fig18-c}
  \rotatebox{90}{\includegraphics[width=4cm,height=5cm]{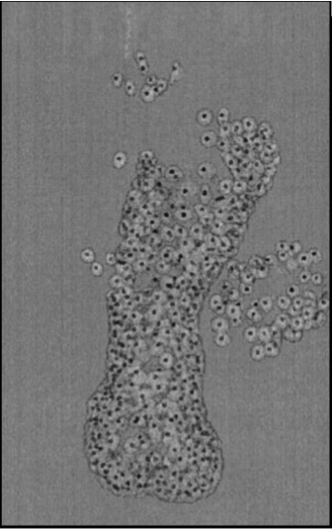}}}
   \quad
 \quad
 \subfigure[Dynamic inspection of volume rendered images with objects incorporated]{
  \label{fig18-d}
  \rotatebox{90}{\includegraphics[width=4cm,height=5cm]{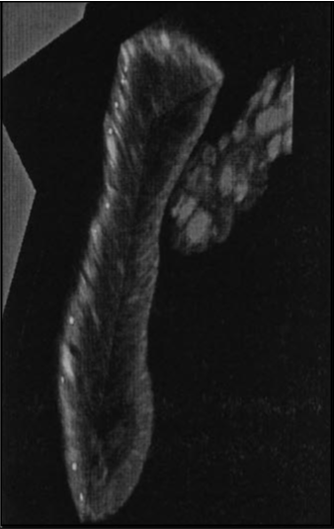}}}
 \caption{The overview of proposed method mentioned in \cite{Wang-2003-3AO}. The figure corresponds to Fig.8 in the original paper.}
  \label{fig:18}
\end{figure}

In \cite{Wang-2003-3AO}, a statistical, non-parametric framework based method is proposed for three-dimensional (3-D) object detection and labelling. The first step is extracting the nuclei from the observed image. In this step, the image is segmented into regular non overlapped cubes. Each cube is then inspected based on defined criteria. The cubes identified as the nucleus are setting to white. At last, all white cubes in the same cell are merged into a whole by an iterative merging algorithm. Fig.\ref{fig:18} indicates the results obtained based on proposed method. Experiment suggests that proposed method achieves a fast cell detection on prepared data within 148 seconds.
\par
In \cite{Wang-2003-3BO}, a low computational complexity and robust method for 3-D biological object detection and labelling is described. The first step is extracting the nuclei from the observed image. In this step, the image is divided into regular non overlapped cubes. Each cube is then inspected based on defined criteria. The cube identified as the nucleus is setting to white. At last, all white cubes in the same cell are merged into a whole by an iterative merging algorithm. Several experiments are performed on real image data to demonstrate the applicability. 439 nuclei are detected in the experiments.
\par
In \cite{Qing-2006-AOM}, a method for identifying and counting algae microscopic color images is proposed. This method sequentially performs the following operations on the image: image smoothing by median filtering, threshold segmentation based on the hue saturation and intensity (HSI) color space characteristics of the image, seed filling, noise elimination based on the noise region threshold segmentation, refinement and the final step of the total calculation quantity. The experiment compares the performance of the proposed system with manual statistics on 40 images of algae cells containing Chlamydomonas and Chlamydomonas bicuspidata. The result shows that compared with manual recognition method, the recognition rate of proposed system is greater than 90$\%$, indicating that this method is feasible. 
\par
For improving the accuracy of traditional image segmentation and image enhancement algorithms, in \cite{Li-2007-EDO}, an image edge detection method based on histogram equalization and soft mathematical morphology is proposed. This method is on the basis of combining traditional image enhancement and image segmentation. The first step is preprocessing, including histogram equalization and histogram specification. The second step is flexible morphological operations, including corrosion and expansion. The last step is edge detection based on the degree of gray value change. Based on this method, the detection accuracy can be improved and the edge details can be protected well. 
\par
In \cite{Costa-2008-AIO}, in order to reduce the labor workload of clinician while diagnosing tuberculosis and the cost of the patient, an automatic detection algorithm is proposed for TB. Firstly, a difference image (R channel minus G channel) from red green and blue (RGB) color format is obtained. Secondly, a global adaptive threshold operation is applied for TB segmentation. At last, several filtering operations are used for separating the bacilli from artifacts. About 50 images are analysed in the experiment, and the number recognized by experts is regarded as a reference standard. Experiment result shows that the proposed algorithm yields the highest sensitivity of 76.65$\%$ and the best FP rate of 12$\%$.
\par
In \cite{Zhang-2008-AAB}, for improving the counting efficiency of bacterial colonies, a fully automatic and cost-effective bacterial colony counter is introduced. First, original images are classified into color images and non-color images. For all images, Otsu threshold is applied for segmenting foreground object region, which is named as the dish/plate area. After that, for color image, Otsu threshold is applied for further objects segmenting from dish/plate area. Watershed algorithm is employed for final segmentation of object with more than one colony. For non-color image, the further objects segmenting from dish/plate area is based on a statistic method. The final detection number is calculated by adding the number detected in the color image to the number detected in the non-color image. For testing experiment, 100 chromatic and achromatic images are prepared. In addition, Otsu’s method is employed for a controlled experiment. Result shows that the presented algorithm has better performance with a satisfaction rate of 96$\%$ compared to Otsu’s method.
\par
In \cite{Wang-2008-MMT}, an automatic method is introduced to track motile microorganism. Microorganisms in motion are tracked through the movement of the XY platform. The main image processing method used in this work is multiple thresholds based object recognition. For the binary image, the centroid of the target mask is regard as the coordinates to calculate the speed and direction of the target movement in the time series image. The tracking control system is tested on the unicellular photosynthetic alga Chlamydomonas reinhardtii. Experiment result shows that combined with proposed microorganism detection method, the system can continuously track cells in motion for up to 300 seconds.
\par
In \cite{Fernandez-2006-MDI,Fernandez-2008-MVA}, an automatic machine vision based method is presented for micro-organism oocysts detection, which can not only provide satisfactory detection results, but also require very little time. Two detection methods are mentioned in this system. The first method is threshold twice. Firstly, the brighter and darker components of the image are respectively processed by threshold. After that, an or operation and denoising are performed on two images processed. At last, the objects are detected by employing Danielsson circle detection method. The other method is based on independent red color space and green color space. Based on the green plane, the wall and cell nucleus of Cryptosporidium are detected. Based on the red plane, only the nucleus are detected. The number of nuclei is obtained from the red image. The number of cell walls is obtained by subtracting the two. Experiment result shows that the proposed method achieve a successful detection rate of 100$\%$.
\par
In \cite{Coltelli-2007-RMA}, for measuring the velocity of moving objects in microscopic images, a novel digital system is proposed. The first step is a subtraction operation between two successive frames. Based on this step, moving objects in video are detected while stationary objects are eliminated. The difference images are then stored. The third step is automatic labelization of the cells moving. At last, the variation of area is calculated and regarded as a measure of speed. 10-images time sequence (400 msec) are used for test. The test data shows that the speed of \textit{Dunaliella} cells is about 100 $\rm \mu$/sec, which is similar to the speed of \textit{Dunaliella} cells of 100 $\rm \mu$/sec reported in previous studies \cite{Rose-1974-VHA}. This means the proposed system can accurately measure the speed of moving object.
\par
For providing the doctors a faster way to diagnose tuberculosis and allowing patients to spend less, in \cite{Sotaquira-2009-DAQ}, a image processing techniques based method is described. This method can perform bacterial segmentation and cluster detection for TB diagnosis purposes. First, color space of the original image in RGB is converted into YCbCr and Lab color spaces. Then Cr plane of YCbCr color space and a plane of Lab color space are then respectively segmented by threshold calculated. At last, two segmented images are combined for final detection. 1400 images of 14 patients are used for testing this method, which takes the ratio of correctly segmented objects to the total number of objects detected in the image as standard. Experiment results suggest that method proposed in this work yields a good detection performance with an average efficiency of 96.3$\%$.
\par
\begin{figure}[ht]
\centering
\includegraphics[width=0.95\linewidth]{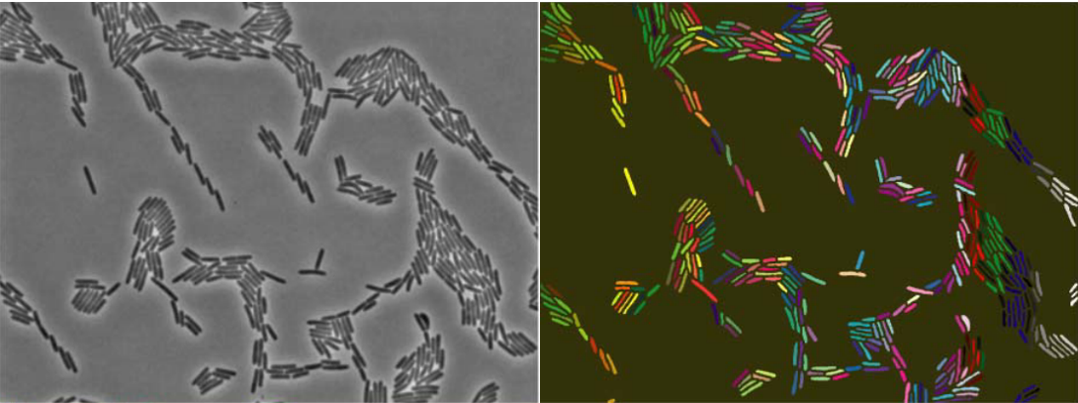}
\caption{The detection result of proposed method in \cite{Vallotton-2010-SOD}. The figure corresponds to Fig.5 in the original paper.}
\label{fig:8} 
\end{figure}

In \cite{Vallotton-2010-SOD}, a novel detection method is proposed for characterizing bacteria, that is from high resolution phase contrast image. The first step is boundary detection by Marr Hildreth edge detector. The second step is selecting features for detecting splitting, including the image contrast at a particular candidate point, a constriction at the septum, feature provided by a probabilistic model and angular feature. After that, the normalization operator is then employed in four features selected above to make them between zero and one and then multiply. Finally, segmentation is performed according to the values obtained in step three and preset threshold. Result shows that the system not only achieves slightly better results than manual counting, but also provides reliable information of the bacteria shape. Fig.\ref{fig:8} shows an example of segmentation result.
\par
In \cite{Mukti-2010-DAC}, a method for detecting and counting Mycobacterium TB is proposed. Through this method, a threshold suitable for the image can be obtained. On this basis, the main goals of color threshold and image segmentation can be achieved. The first step is image segmentation by color threshold method. Counter is then used for counting out number of black Mycobacterium TB cells based on resulting binary image. By employing the method, the patient's disease grade determined by the number of TB cells can be accurately judged. 
\par
In \cite{Verikas-2012-AIA}, in order to protect the environment better, the system of detecting P. minimum automatically is proposed, which can get the quantitative concentration estimates of object. Firstly, an object image is created by segmenting Lab image. The center and contour of circular objects are then determined, respectively. After that, circular noise is eliminated based on the area threshold. For images with more than one object, the method takes a sequential detection method that removes each detection object. The developed algorithm is tested with 114 images recorded by a color camera. According to the test results, the introduced algorithms makes a good detection performance with a detection rate of 93.25$\%$.
\par
\begin{figure}[ht]
\centering
\includegraphics[width=0.95\linewidth]{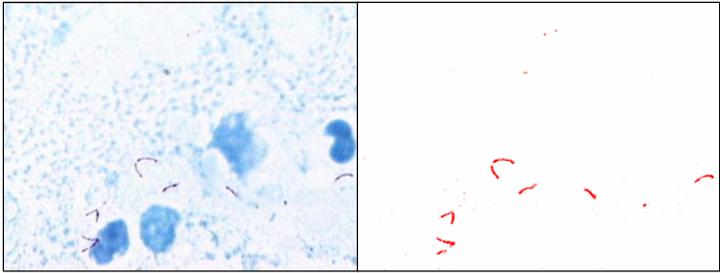}
\caption{The segmentation result of proposed method in \cite{Raof-2011-ISO}. The figure corresponds to Fig.6 in the original paper.}
\label{fig:10} 
\end{figure}

In \cite{Raof-2011-ISO}, for handling with large number of TB cases, a TB image segmentation method is proposed with the same accuracy as manual diagnosis and higher efficiently. The main idea includes two steps. One is image enhancement that involves adjusting image brightness, contrast, and color operations. The other is achieved by using color threshold technique. At this step, the color information is collected for TB pixels detection. An example of segmentation result is shown in Fig.\ref{fig:10}. The final images suggest that TB in the original images can be detected accurately based on proposed method.
\par
In \cite{Shi-2012-FBA}, to improve the efficiency of counting the total number of colonies, an automatic detection method of bacteria with high practical value is proposed. The main idea is to design a circular filter based on the features of low brightness around the live bacteria and high brightness in the center for detection. By comparing the results of manual detection and machine detection, it is shown that the recognition error of this system is within 10$\%$.
\par
In \cite{Rachna-2013-DOT}, an algorithm based on image processing is developed for identification of TB bacteria in sputum, which can reduce the workload of doctors in diagnosing tuberculosis. The first step is preprocessing. Then a global threshold is employed to remove tissue and background. After that, $k$-means clustering approach and Otsu threshold approach are respectively applied to detect TB for comparing their performances. At last, region growth is applied to mark and remove noise and large areas of excessive pollution after objects extracted by $k$-means clustering approach or Otsu threshold approach. The fully automatic TB bacilli segmentation method is tested on 25 positive tissue slides. The test result shows that the proposed method yields a satisfactory performance in TB segmentation with an accuracy of 98.00$\%$.
\par
In \cite{Matuszewski-2013-PDA}, for detection and tracking small objects in video streams, a set of spatial filters are designed. The first step is to calculate the Fourier transformation for the sample image. The second step is creating a binary filter by the values above the threshold parameter. Thirdly, the input image is transformed to the frequency domain and its spectrum is multiplied by the filter. After that, the inverse Fourier transformation of the result is calculated and a very blurred image is obtained. At last, other threshold value is applied to the output picture. Experiments show that the presented method fits for detecting and tracking known objects.
\par
In \cite{Kowalski-2014-ASL}, an effective nematode extraction method is proposed for obtaining more informations about nematode behavior. First, the grayscale image is binarized using the adaptive threshold method. Second, a globle threshold is applied for extracting reference labels from original image. The XOR operation are then employed between the binary image obtained in first step and the binary image with extracted markers. At last, the position of nematode detected is determined by the center of the largest blob. Based on this method, the location of the nematode can be accurately determined. At the same time, with the aid of this method, it is possible to better track the nematodes. 
\par
\begin{figure}[ht]
\centering
\includegraphics[width=0.95\linewidth]{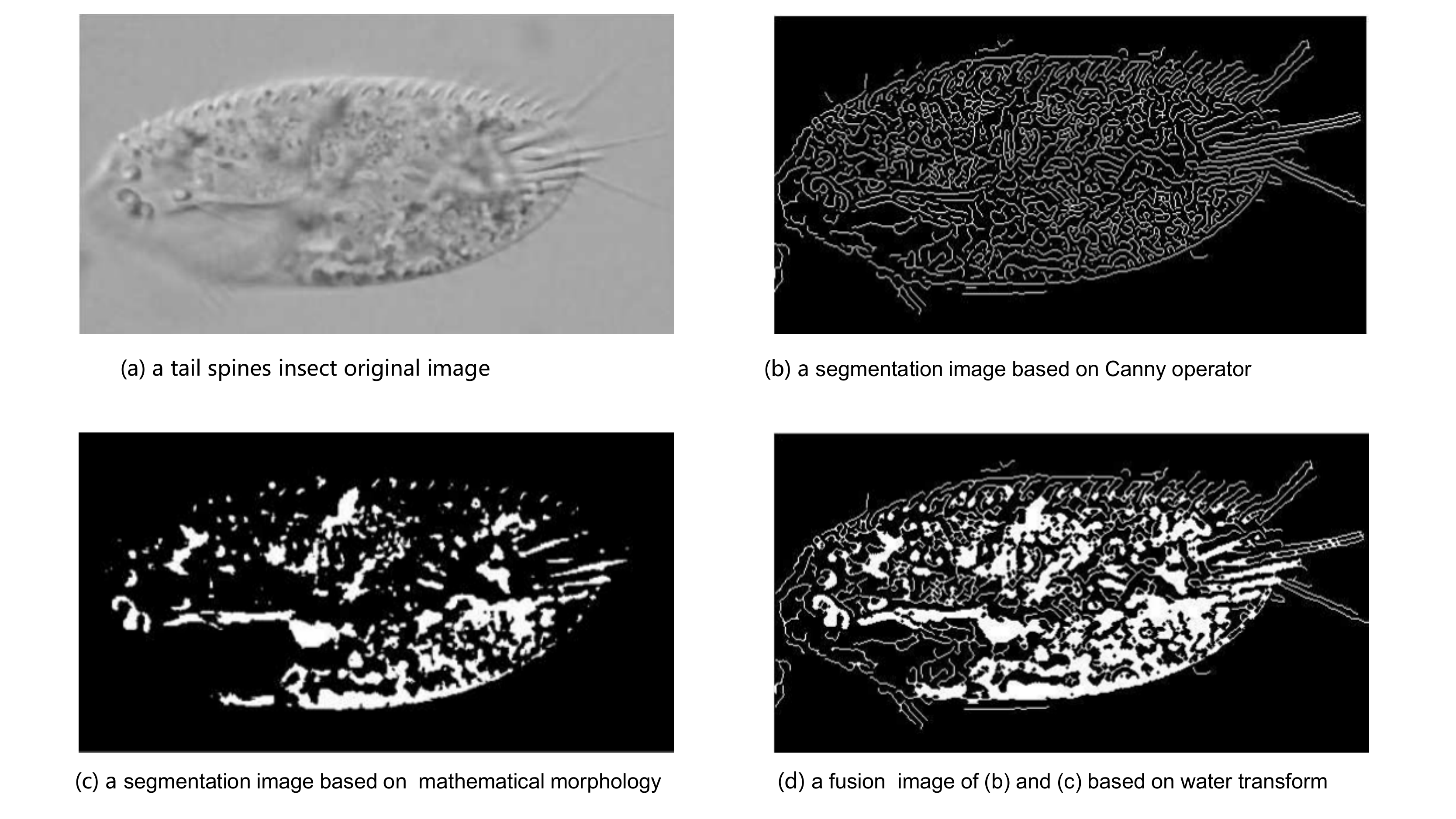}
\caption{The segmentation results of proposed method in \cite{Wang-2013-IFT}. The figure corresponds to Fig.1-Fig.4 in the original paper.}
\label{fig:11} 
\end{figure}

In \cite{Wang-2013-IFT}, in order to replace manual observation with automatic observation based on machine vision, this paper proposes an edge detection method that is more suitable for low-contrast images. Canny operator is firstly employed for segmenting the low-contrast image while mathematical morphology is also applied in the backup image for the same motivation. Then two segmented images resulting from two method respectively are fused by employing wavelet transform. Experiment shows that the segmented image after fusion has better results than the previous image segmented based on a single method. It can be seen from Fig.\ref{fig:11} that the fused segmentation image can reflect the object feature more completely and clearly. 
\par
In \cite{Kurtulmucs-2014-DOD}, a new method using computer vision to detect and count the deaths of Heterobacter elegans from microscope images is proposed for better controlling agricultural pests and improving the detection efficiency of Heterobacter elegans. The first step is obtaining smooth medial axes of the nematode worms by a median filter, Otsu threshold, a morphological operation of area opening and filling operation. The overlapped nematode worms  are then separated by a skeleton analysis. Finally, two different path analysis methods are used for detection of dead nematodes, which are both based on the standard error of the mean. A dataset containing total 685 images is prepared for verifying the performance of proposed methods, which includes 935 live worms and 780 dead one. The experiment result shows that the proposed algorithm can achieve a good performance with a detection rate of 85$\%$.
\par
In \cite{Farahi-2015-ASO}, in order to better diagnose visceral leishmaniasis, a method of segmenting leishman bodies is proposed. The first step is preprocessing by a linear contrast stretching transformation. In order to improve the performance of final segmentation, all small oval objects in image are extracted by applying morphological closing. In this step, objects, including leishman bodies and other shape-like objects are extracted. To extract boundary of objects detected, CV level set is employed. Finally, to remove the influence of other shape-like objects, a local threshold is applied. Experiment results suggest that the proposed method yields a mean of segmentation error of 9.76$\%$.
\par
In \cite{Goyal-2015-ADO}, for improving the efficiency of tuberculosis diagnosis, a reliable method is presented. The first step is converting the color space of input image into gray space. After that, in order to improve the performance of TB segmentation, tubeness filtering is employed for highlighting objects. The third step is roughly extracting bacteria based on Otsu global threshold. After labelling the extracted bacteria, all the other noise are removed. When compared to expert manual detection, the automatic detection algorithm presented in this work obtains a count of bacteria that shows a strong correlation with manual results.
\par
In \cite{Shah-2016-ADA}, to improve the sensitivity and specificity of tuberculosis testing, an automatic detection method used on smart-phones is introduced. The first step is preprocessing that involves grayscale conversion and contrast enhancement. The second step is to binarize the image based on the threshold to separate the bacteria from the background. Morphological closing and filling are then performed. In step four, artifactas are removed based on geometric features. Finally, overlapping (touching) bacilli are segmented and separated by the watershed algorithm, after labelling TB. Total 30 images from smart-phone are prepared. One half is TB positive, the other half is TB negative. Experiment result shows that the presented method achieves a sensitivity of 93.3$\%$ and a specificity of 87$\%$.
\par
In \cite{Zhou-2016-MCE}, a microbial contour extraction method is proposed to extract object contours from sewage microscopic images. First, the Sobel operator is used for to calculating the gradient of the image. Based on the calculated values, four adaptive thresholds are then employed for obtaining better object edges. After that, the extracted edges are connected based on the edge connection method designed. Finally, the outermost contour edge is then extracted from the image after processing of edge connection. Experiment shows that compared with traditional edge detection algorithms, the proposed method can extract higher quality of microbial contours. 
\par
In \cite{Payasi-2017-DAC}, for improving the efficiency of tuberculosis diagnosis, a feasible algorithm is proposed. The first step is converting the image color space from RGB to HSI. The second step is segmenting the Hue component image. After denoising based on the area threshold, the holes are also filled. Finally, object is segmented from the processed image and the area and perimeter of object are calculated. Experiment results on prepared data show that the proposed method yields a high accuracy of over 90$\%$ in TB detection.
\par
In \cite{Kemmler-2011-DOM}, for reducing the workload of labelling and counting bacteria in bright field microscopes, several semantic segmentation techniques are compared. One of introduced methods is region-based level set segmentation method. A database including five different microbe species with 40 up to 470 microbes per class is prepared for testing. The average recognition rate is employed for evaluating the detection performance. Other methods involved in this paper are explained in the corresponding chapters of their category. 
\par
\subsection{Classification Based Methods}
\label{Sect:3-2}
\begin{figure}[http]
\centering
\includegraphics[width=0.98\linewidth]{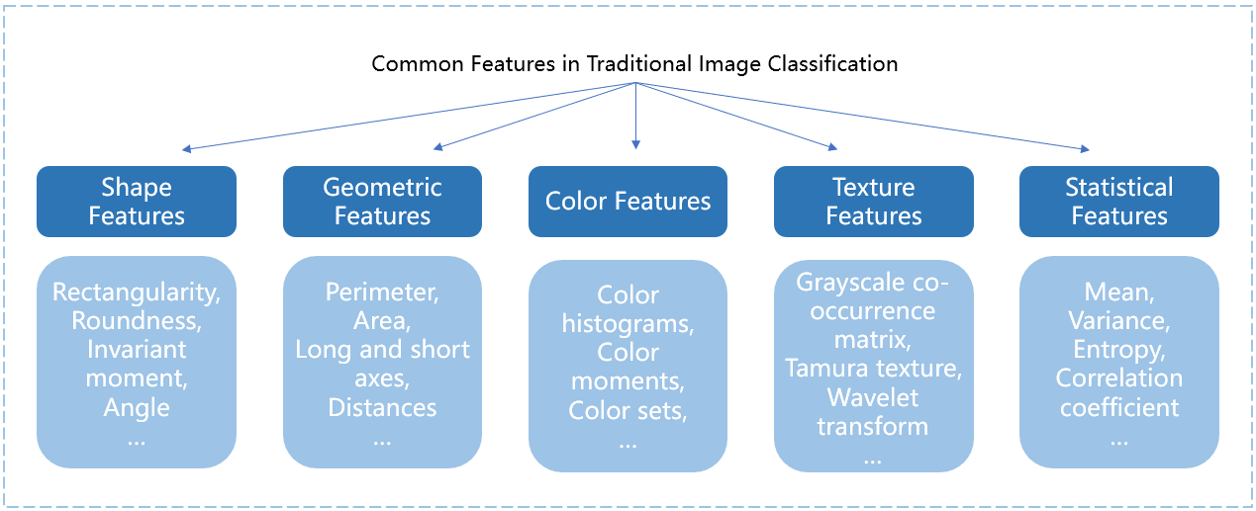}
\caption{Features commonly used in traditional image classification.}
\label{fig:3.2-1} 
\end{figure}

In this subsection, related works that use classification methods for microorganism detection are introduced. To present this works more clearly, we decide to introduce the involved methods separately according to their adopted features. Features commonly used in traditional image classification, as shown in Fig.\ref{fig:3.2-1}, include shape features, geometric features, color features, texture features, statistical features, etc.
\par
\subsubsection{Classification Based Methods with Shape Features}
\label{Sect:3-2-1}
Shape features are commonly used in traditional image detection methods, which mainly contain contour features and area features. More specifically, it include feature parameters such as squareness, angular, roundness, invariant moment, eccentricity, polygon description, and curve description.
\par
\begin{figure}[http]
\centering
\includegraphics[width=0.95\linewidth]{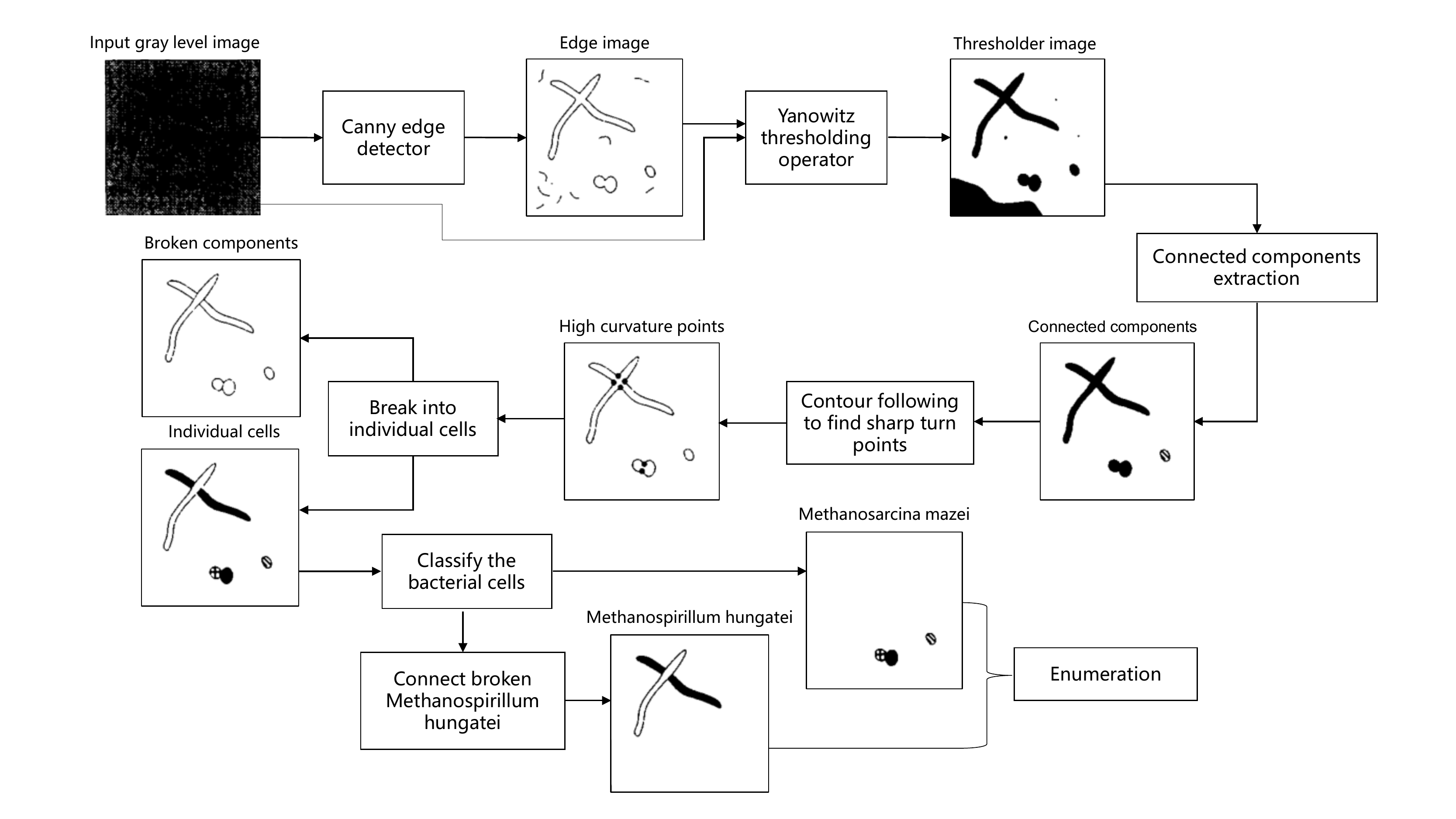}
\caption{The overview of proposed method mentioned in \cite{Dubuisson-1994-SAC}. The figure corresponds to Fig.1 in the original paper.}
\label{fig:9} 
\end{figure}
\begin{figure}[htbp]
 \centering
 \subfigure[Original image of diatom algae]{
  \label{fig19-a}
  \includegraphics[width=4.7cm,height=4cm]{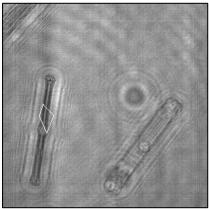}}
 \quad
 \quad
 \subfigure[Segmentation result of (a)]{
  \label{fig19-b}
  \includegraphics[width=4.7cm,height=4cm]{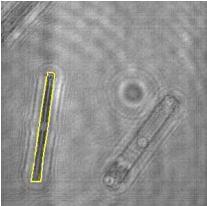}}
 \quad
\subfigure[Reference sample]{
  \label{fig19-c}
   \includegraphics[width=5cm,height=4cm]{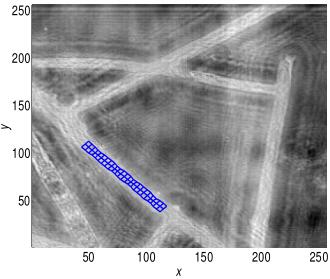}}
 \quad
 \subfigure[Recognition result of (c)]{
  \label{fig19-d}
  \includegraphics[width=5cm,height=4cm]{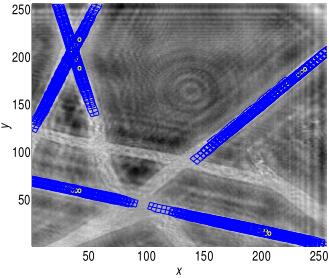}}
   \quad
 \caption{The detection results of proposed method in \cite{Javidi-2006-RA3}. The figure corresponds to Fig.6 and Fig.10 in the original paper.}
  \label{fig:19}
\end{figure}

In \cite{Dubuisson-1994-SAC,Javidi-2006-RA3,Yeom-2006-R3S,Liu-2014-AOI,Liu-2014-VID}, contour feature, as a type of shape feature, is employed for microorganism detection. In \cite{Dubuisson-1994-SAC}, contour feature is extracted by Canny edge detector and threshold segmentation. Based on contour feature, Methanospirillum hungatei and Methanosarcina mazei are successfully distinguished and detected. Specific process of described algorithm is shown in Fig.\ref{fig:9}. In \cite{Javidi-2006-RA3,Yeom-2006-R3S}, contour feature is combined with rigid graph matching (RGM) method to segment and detect biological microorganism in 3-D image. The segmentation results and detection results are respectively exhibited in Fig.\ref{fig:19}. In \cite{Liu-2014-AOI,Liu-2014-VID}, an optofluidic imaging system,which obtains contour feature and other informations by the image processing software called ImageJ, is proposed. In \cite{Liu-2014-AOI}, the system is used for detecting \textit{E. coli}, \textit{Shigella flexneri} and \textit{Vibrio cholera}. In \cite{Liu-2014-VID}, it is proposed to detect flu virus by the same research group with \cite{Liu-2014-AOI}. Experiments in \cite{Liu-2014-AOI,Liu-2014-VID} show that all the prepared microorganisms can be effectively detected and classified with proposed system.
\par
\begin{figure}[http]
\centering
\includegraphics[width=0.75\linewidth]{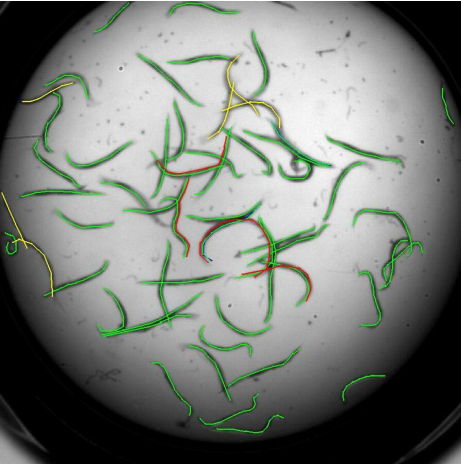}
\caption{The detection results of proposed method in \cite{Rizvandi-2008-ELB}. The figure corresponds to Fig.5 in the original paper.}
\label{fig:13} 
\end{figure}

In \cite{Rizvandi-2008-SAO,Huang-2008-ADA,Rizvandi-2008-ELB,Rizvandi-2008-AID,Zhou-2008-UMV}, some similar methods based on angular features are proposed to detect \textit{C. elegans} nematode worms. This methods contain several similar steps. The first step is preprocessing, including image binarization, small hole filling, skeleton and pruning. The second step is splitting skeleton into independent branches by processing specific pixels. Then, all the individual skeletons are reconstructed by applying branch merging method. At last, angular features extracted from proposed skeletons are used for \textit{C. elegans} nematode worms detection. An outcome of experiment in \cite{Rizvandi-2008-ELB} is shown in Fig.\ref{fig:13}. Related result shows that the best percentage of correctly detected worms is about 83$\%$. 
\par
\begin{figure}[http]
\centering
\includegraphics[width=1\linewidth]{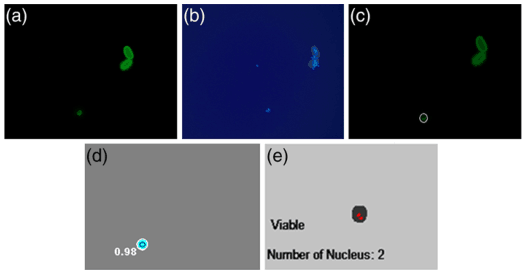}
\caption{Detection of \textit{Cryptosporidium} and \textit{Giardia (oo)cysts} in \cite{Badsha-2013-ACA}. The figure corresponds to Fig.8 in the original paper.}
\label{fig:3.2-2} 
\end{figure}

In \cite{Zhai-2010-AIO,Hiremath-2011-SAI,Badsha-2013-ACA}, roundness feature is especially employed for detecting microorganisms that are approximately circular in shape. Method presented in \cite{Zhai-2010-AIO} includes image segmentation and bacilli detection by roundness and roughness. Among them, roundness is used for distinguishing Single-bacillus objects and touching-bacillus objects. The experiment is carried out on 100 images. Result shows that 95$\%$ of total prepared samples yield a good detection accuracy of over 80$\%$. In \cite{Hiremath-2011-SAI}, after several preprocessing operator, such as grayscale conversion, image intensity values adjusting and binarized, roundness data is obtained from binarized objects. At last, all Rotavirus-A particles are identified based on roundness and pre-set rule. 50 images are prepared for test the performance of presented algorithm. Experiment result shows that the identification rate of presented algorithm is 98$\%$. In \cite{Badsha-2013-ACA}, roundness combined with eccentricity is applied for detecting \textit{Cryptosporidium} and \textit{Giardia (oo)cysts}. The corresponding detection process is shown in Fig.\ref{fig:3.2-2}. 40 images containing \textit{Cryptosporidium} and \textit{Giardia (oo)cysts} are prepared for testing the performance of described method. Performance is evaluated by nucleus counting rates. In addition, in order to evaluate the performance of the method more objectively, its results are compared with manual counts. Experiment results indicate that proposed method achieves a detection rate of 97$\%$.
\par
\subsubsection{Classification Based Methods with Geometric Features }
Geometric features generally refer to the position and direction of object in image, as well as the perimeter, area, distance and other characteristics of objects. Geometric features are more intuitive and simple, and can play a good role in object detection. 
\par
In \cite{Fang-2008-AIO,Liu-2014-AOI,Liu-2014-VID,Yu-2014-DOI}, areas of objects are chosen as an important feature for accurate detection. In \cite{Fang-2008-AIO}, area and Neural Network (NN) are applied for automatically identifying TB in Acid-Fast stain sputum smears. Experiments on 44 Acid-Fast stain microscopic images with 533 TB show that the Perceptron has 100$\%$ sensitivity and 39.8$\%$ specificity; the feed-forward backpropagation has 97.8$\%$ sensitivity and 72.4$\%$ specificity. In \cite{Liu-2014-AOI,Liu-2014-VID}, not only the shape features mentioned in the previous subsection \ref{Sect:3-2-1} is applied, but also area feature. Results indicate that method proposed in this research performs well in detecting a wide range of bacteria by combining shape features and area feature. In \cite{Yu-2014-DOI}, area is considered as an important indicator to detect the presence of bacteriophage and to estimate the number of bacteriophage. In \cite{Coltelli-2013-AAR}, centroid distance spectrum extracted from processed image is selected for microalgae detection. This methodology achieves a high accuracy of 96.6$\%$ in 3423 samples, which contain 24 kinds of microalgae. In \cite{Mader-2015-IST}, an image-analysis scheme using width information is presented to detect fungal infection. 415 images are prepared for evaluating the performance of proposed method with 194 positive samples and 221 negative one. Experimental results indicate that the proposed method yields a sensitivity of 83$\%$ and a specificity of 79$\%$ on prepared data.
\par
\begin{figure}[http]
\centering
\includegraphics[width=0.95\linewidth]{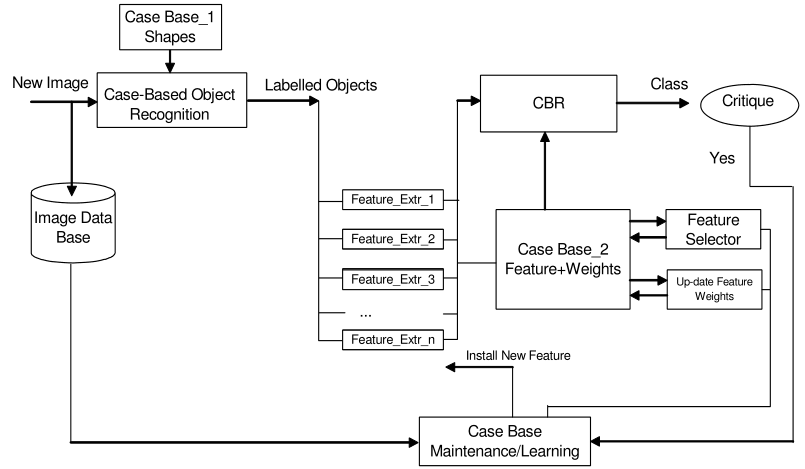}
\caption{The overview of proposed method mentioned in \cite{Sklarczyk-2007-IAA}. The figure corresponds to Fig.6 in the original paper.}
\label{fig:17} 
\end{figure}
\begin{figure}[http]
\centering
\includegraphics[width=0.82\linewidth]{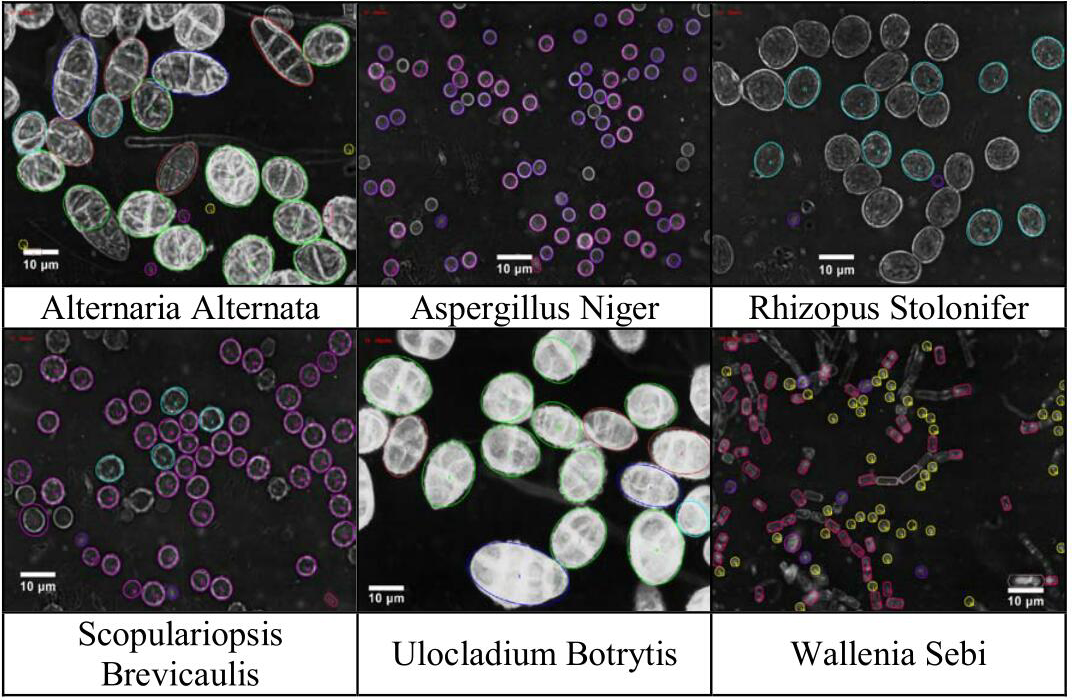}
\caption{The tested objects in \cite{Perner-2004-ROA}. The figure corresponds to Tab.2 in the original paper.}
\label{fig:16} 
\end{figure}

In fact, the combination of several suitable geometric features selected for different microorganisms can achieve better detection. In \cite{Fukuda-1989-ESD,Forero-2003-AIT,Kumar-2008-GAO,Coltelli-2013-AAR,Mader-2015-IST,Jan-2015-DOT,Perner-2004-ROA,Sklarczyk-2007-IAA}, they achieve accurate and efficient detection of different microorganisms by using different combinations of geometric features. In \cite{Fukuda-1989-ESD}, a novel approach that considers a complete microbial individual as an individual composed of several basic shapes is proposed. Based on this basic shape’s length and area features, the system matches the image with prepared data base. The identification experiment on three kind of microorganisms shows that the best detection performance is yielded in Nitzschia Fonticola with a detection rate of 90$\%$. In \cite{Forero-2003-AIT}, many geometric features are randomly combined and tested for choosing the most suitable combination of features. At last, the combination of length, width and mahalanobis distance with a classification tree achieves the best result in bacilli detection. In \cite{Perner-2004-ROA,Sklarczyk-2007-IAA}, the area-to-length ratio is applied for determining how close the target is to prepared template. The whole system architecture is shown in Fig.\ref{fig:17}. Six different airborne fungi spores (shown in  Fig.\ref{fig:16}) are tested with proposed method. The highest recognition rate can be achieved for Scopulariopsis Brevicaulis with 98.2$\%$. In \cite{Kumar-2008-GAO}, some geometric features of the region of interesting (ROI) are obtained by Image-Pro Plus for exploring a reliable and automatic microorganism detection technique, including width, length, area, perimeter and aspect ratio. In \cite{Jan-2015-DOT}, a method that applies area and bounding box is presented for greatly improving the efficiency of doctor diagnosis of tuberculosis. After color space conversion, several morphological operations and edge detection, data about area and bounding box are obtained. Experiments on 100 positive samples and 10 negative samples show that it achieves a high accuracy rate of 90$\%$.
\par
\subsubsection{Classification Based Methods with Color, Texture or Statistical Features }
Color features can describe the surface properties of the corresponding scene in the image or image area. Color features description can be divided into color histogram, color distribution, color set, etc. In \cite{Ogawa-2005-MDI}, a novel multi-colour detection method is described for bacteria detection and metabolic activity assessment. Result shows that when compared to expert manual detection, the automatic detection algorithm presented in this work gains a count of bacteria that shows a strong correlation with manual results. In \cite{Tripathi-2007-DAI}, for detection and identification of individual bacterial cells and spores, a method based on Raman chemical imaging microscopy and color features is proposed. Result shows that Raman chemical imaging microscopy can distinguish a mixture of Gram-positive \textit{Bacillus atrophaeus} spores and Gram-negative \textit{E. coli} cells. In addition, color features are usually combined with other features for better detection performance. In \cite{Kumar-2008-GAO,Coltelli-2013-AAR}, color feature and geometric feature together constitute the feature parameters of the object. In \cite{Forero-2003-AIT,Fang-2008-AIO}, color features are regarded as the criteria for the initial screening of regions.
\par
Texture features can reflect the slowly changing surface organization structure arrangement properties of the surface of objects. In \cite{Thiel-1995-TAD}, texture features are employed for detecting and distinguishing \textit{Anabaena} and \textit{Oscillatoria}. In this work, 14 \textit{Anabaena cylindrica} and 20 \textit{Oscillatoria agardhii} are prepared as the test group. The exoeriment result shows that the method distinguishes between \textit{Anabaena} and \textit{Oscillatoria} with an accuracy of over 90$\%$. In \cite{Javidi-2005-TIA}, a real-time microorganism detection method by applying texture feature and RGM method is presented. Experiment result shows that the proposed method has strong performance on moving objects and different conditions. In addition, in \cite{Zhai-2010-AIO}, texture features are combined with shape features for TB detection.
\par
\begin{figure}[http]
\centering
\includegraphics[width=0.95\linewidth]{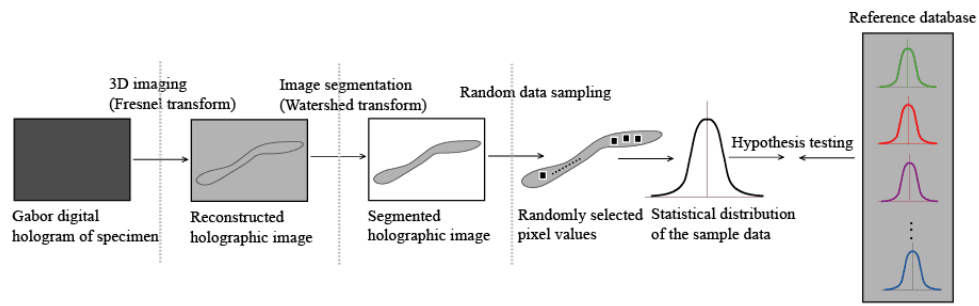}
\caption{The overview of proposed method mentioned in \cite{Moon-2010-ATM}. The figure corresponds to Fig.2 in the original paper.}
\label{fig:20} 
\end{figure}

Statistical features include mean, variance, energy, succession, etc. Statistical features are simple to compute and insensitive to the exact spatial distribution of color pixels. In \cite{Moon-2010-ATM,Javidi-2010-DIA}, a method based on statistical sampling methods is proposed for automated 3-D sensing and recognition of biological microorganisms. The specific steps are shown in Fig.\ref{fig:20}. The test result shows that by comparing the variances, biological microorganisms can be accurately detected. In \cite{Daneshpanah-2010-3HI}, escape-force measurement \cite{Schaa-l2009-MCD} is combined with other cell identifications for detecting, classifying and controlling microorganisms and cells. In \cite{Shin-2010-OSF}, a system, which consists of a microfluidic device, a digital holographic microscope and relevant statistical recognition algorithms, is described for 3-D sensing and detection of microorganisms. Firstly, microorganisms are provided by the microfluidic device for the system to process. The Fresnel diffraction pattern of microorganisms is then optically recorded as a digital hologram. Thirdly, sampling segment features are randomly extracted from the reconstructed wavefront data. At last, statistical recognition algorithm is applied for cell identification. Microorganisms used in the experiments are \textit{Euglena acus} and \textit{Chilomanas}. Experimental results suggest that the optical fluid 3-D sensing and recognition method is feasible. In \cite{Yourassowsky-2014-HTH}, a system consisting of a Digital Holographic Microscope (DHM) and a partial coherent source is proposed for detecting and extracting objects in prepared samples. First, a non-zero complex amplitude is applied for determining whether the object exists. After that, the inverse Fourier transform and threshold operation are performed to obtain the location of the object. To assess the performance of DHM, tests are performed with a \textit{Giardia lamblia} cysts image. Experiments show that the obtained phase and intensity images can be used for detection and classification.
\par
\subsubsection{Summary}
Based on the above analysis of relevant studies on the use of classification methods for microorganism detection, we can find the that:
\begin{itemize}
    \item The microorganism detection using classification methods has a development history of more than 30 years. Geometric features have been used as classification feature for microorganism detection in \cite{Fukuda-1989-ESD} as early as 1989. 
    \item Shape features and geometric features are widely chosen for microorganism detection due to their simplicity of calculation and ease of use.
    \item Using a combination of multiple features can achieve better detection results than a single feature. 
\end{itemize}

\par
\subsection{Table Analysis}
\label{Sect:3-3}
In the past four decades, classical image processing methods for microorganism detection are continuously developed. All related works can be grouped into segmentation based methods and classification based methods. For better statistical related works, Tab.\ref{tab:4} is designed, where the publication date, literature links, research objects and its domains, main methods and evaluation indicators are presented.
\begin{landscape} 
\begin{longtable}{c|c|c|c|c|c}
\caption{Summary for object detection methods based on classical image processing.}\\
\hline\hline
\endfirsthead
\caption{Continued}\\
\hline\hline
\endhead
\hline

Date & Refs & Microorganism & Domain & Method & Evaluation \\ \hline
\multicolumn{6}{c}{Segmentation based methods} \\ \hline
 1985 & \cite{Sieracki-1985-DEA} & Bacterial & Water-borne & Object label & Compared with visual count  \\
 1988 & \cite{Adams-1988-TUO} & Hyphae & Medical & Segmentation, skeletonization & \makecell[c]{The overall percent\\\ Standard errors = 0.25$\%$ } \\
 1990 & \cite{Packer-1990-MMO} & \makecell[c]{Filamentous\\ microorganisms} & Medical & \makecell[c]{Binarization, circularity test,\\ skeletonization} &\makecell[c]{ Compared with manual\\ analysis} \\
 1991 & \cite{Masuko-1991-ANM} & \makecell[c]{ Luminescent \\bacteria} & Environment & \makecell[c]{Detection of objects that\\ emit light}  & - \\
 1995 & \cite{Bloem-1995-FAD} & Bacteria & Environment &  \makecell[c]{Morphological operation, \\threshold segmentation}  & \makecell[c]{ Compared with visual \\counts   }       \\ 
 1998 & \cite{Baillieul-1998-ODO} & Microorganisms & Water-borne & Threshold & - \\ 
 2003 & \cite{Wang-2003-3AO,Wang-2003-3BO} & \makecell[c]{Cells from the\\ first colonic \\gland of rats} & Medical
 & \makecell[c]{Extracting nuclei, dividing \\image into regular\\ non overlapped cubes}   & - \\
2006  & \cite{Qing-2006-AOM} & Algae & Environment
 & Threshold segmentation &  Recognition rate $\ge$ 90$\%$ \\
 2007 & \cite{Li-2007-EDO} & Bacteria & Food & \makecell[c]{Histogram equalization, \\ morphological operations, \\edge detection} &   -  \\
2007 & \cite{Coltelli-2007-RMA} & Bacteria & Water-borne
  & \makecell[c]{Subtraction operation between\\ two successive frames}  & - \\
 2008 &  \cite{Costa-2008-AIO} &  TB  & Medical & \makecell[c]{Color channel based \\ difference operation, \\a global adaptive threshold} & Sensitivity = 76.5$\%$           \\
 2008 & \cite{Zhang-2008-AAB} & Bacterial colonies & Science  &  Dish/plate detection, Otsu      & Satisfaction rate = 96$\%$           \\
 Date & Refs & Microorganism & Domain & Method & Evaluation \\ \hline
 \multicolumn{6}{c}{Segmentation based methods} \\ \hline
 2008 &  \cite{Wang-2008-MMT} & Microorganisms &  Water-borne &  Multiple thresholds &  -    \\ 
 2008 & \cite{Fernandez-2008-MVA} & \makecell[c]{Microorganism\\ oocysts} & Water-borne
 &  \makecell[c]{Two threshold segmentation,\\ color base detection }     &   Correct detection rate = 100$\%$         \\
 2009  & \cite{Sotaquira-2009-DAQ} &  TB    & Medical &  \makecell[c]{Threshold segmentation, \\a logical AND}  & Average efficiency = 96.3$\%$         \\
  2010    & \cite{Vallotton-2010-SOD} & Bacteria & Science &  \makecell[c]{Marr Hildreth edge detector,\\ threshold segmentation}      &    Compared with manual counts        \\
   2010   &    \cite{Mukti-2010-DAC}       &  TB  & Medical &    Color thresholding segmentation    &     -       \\
   2011   &    \cite{Kemmler-2011-DOM}       &    bacteria, fungi & Medical &   Region-based level set     &  \makecell[c]{The average recognition\\ Rate of yeast = 95$\%$}         \\

    2011  &    \cite{Raof-2011-ISO}       &   TB   &  Medical &    \makecell[c]{Image enhancement, \\Color thresholding \\segmentation}   &    -    \\

   2012   &      \cite{Verikas-2012-AIA}     & P. minimum & Water-borne
 &   \makecell[c]{Threshold segmentation,\\ circular detection}     &  Detection rate = 93.25$\%$      \\
 2012 & \cite{Shi-2012-FBA} & Colonies & Food & Designing a circular filter & Recognition error $\le$10$\%$  \\
 2013 & \cite{Matuszewski-2013-PDA} & Microorganisms & Water-borne
 & \makecell[c]{Fourier transform, threshold, \\inverse Fourier transform} & - \\ 
    2013  &     \cite{Wang-2013-IFT}  &   Protozoa & Environment &   \makecell[c]{ Canny edge detection, \\ wavelet transform}  &     -   \\
   2013 & \cite{Rachna-2013-DOT}  &   TB&   & Otsu thresholding approach  &  Accuracy = 98$\%$      \\
  2014  &  \cite{Kowalski-2014-ASL} &  Nematode   & Science &  Adaptive thresholding   &     -   \\
 2014  &  \cite{Kurtulmucs-2014-DOD} &   \makecell[c]{Heterobacter\\ elegans}  & Medical
 & \makecell[c]{ Median filter, Otsu threshold,  \\skeletonization }  &    Recognition rates $\ge$ 85$\%$.    \\
 Date & Refs & Microorganism & Domain & Method & Evaluation \\ \hline
 \multicolumn{6}{c}{Segmentation based methods} \\ \hline
   2015 &  \cite{Farahi-2015-ASO} &  \makecell[c]{Leishman \\bodies}   & Medical & \makecell[c]{Morphological operation, \\CV level set    } &   \makecell[c]{ Mean of segmentation error\\ for global model = 10.90$\%$    } \\
  2015  &  \cite{Goyal-2015-ADO} & TB   & Medical &   Otsu global thresholding &   \makecell[c]{Compared with expert\\ manual detection }   \\
   2016 &  \cite{Shah-2016-ADA} &   TB &  Medical &  \makecell[c]{Threshold segmentation, \\watershed algorithm } &     \makecell[c]{Sensitivity = 93.3$\%$,\\ Specificity = 87$\%$  } \\
 2016   &  \cite{Zhou-2016-MCE} &  Microorganisms  &  Water-borne & \makecell[c]{ Adaptive thresholds, \\edge connection}  &    -    \\
 2017     &    \cite{Payasi-2017-DAC}       &    TB  & Medical &  \makecell[c]{Color space conversion, \\segmentation, denoising }     &  Counting accuracy = 90$\%$          \\ \hline

\multicolumn{6}{c}{Classification based methods} \\ \hline

 1989  &  \cite{Fukuda-1989-ESD} &  Microorganism & Industry & Basic shapes matching  &  Identification rate = 90$\%$  \\
 1994  &  \cite{Dubuisson-1994-SAC} &  Microorganism &  &  Pattern recognition &  -  \\
 1995  &  \cite{Thiel-1995-TAD} &  \makecell[c]{  \textit{Anabaena} ,\\ \textit{Oscillatoria}} & Water-borne & Texture feature based classification &  Accuracy $\ge$ 90$\%$  \\
2003& \cite{Forero-2003-AIT} & TB & Medical  & \makecell[c]{Canny operator,\\ classification tree} &  Specificity = 96.09$\%$ \\
 2004 & \cite{Perner-2004-ROA} & Fungi spores & Environment & \makecell[c]{Generating template,\\ template matching}  & Recognition rate = 98.2$\%$  \\
  2005 & \cite{Javidi-2005-TIA}  &  Alga & Water-borne &  Threshold, RGM &  -  \\
  2005 & \cite{Ogawa-2005-MDI}  &  Bacteria & Medical &  Colour based classification &  Compared with visual count  \\ 
  
  2006 & \cite{Javidi-2006-RA3,Yeom-2006-R3S}  &  \makecell[c]{ Sphacelaria alga,\\ tribonema alga}  & Water-borne & RGM, statistical algorithms  &  -  \\
 Date & Refs & Microorganism & Domain & Method & Evaluation \\ \hline
 \multicolumn{6}{c}{Classification based methods} \\ \hline
 2007 & \cite{Sklarczyk-2007-IAA} & Fungi & Air-borne & Case based matching & Recognition rate = 98.2$\%$  \\
 2007 & \cite{Tripathi-2007-DAI} & Bacterial & Science & \makecell[c]{Raman chemical imaging \\microscopy} & - \\ 
  2008 &  \cite{Kumar-2008-GAO} &  Bacteria  & Food &  \makecell[c]{ Analyzing geometrical\\ parameters and three\\ optical parameters}  &   - \\
  2008 &  \cite{Rizvandi-2008-ELB} &  \textit{C. elegans} & Science &  Binarization, skeletonization &  Correctly detected rate = 83$\%$  \\
 2008  &  \cite{Rizvandi-2008-SAO} &  \textit{C. elegans} & Science &  Binarization, skeletonization &  FRR = 7.9$\%$, FAR = 8.4$\%$ \\
 2008  &  \cite{Huang-2008-ADA} & Nematode  & Science &  \makecell[c]{ Adaptive threshold, morphological\\ closing, skeletonization } & -   \\

 2008  &  \cite{Rizvandi-2008-AID} & Nematode  & Science  &  Binarization, skeletonization &  Detected accuracy = 89$\%$  \\
  2008 &  \cite{Zhou-2008-UMV} &  Nematode   & Science &  Binarization, skeletonization  &  - \\
  2008 &  \cite{Fang-2008-AIO} & TB  & Medical &  \makecell[c]{Color based detection, \\geometric features based \\classification} &  \makecell[c]{Sensitivity = 97.8$\%$,\\Specificity = 72.4$\%$ } \\
 2010  &  \cite{Zhai-2010-AIO} & TB  & Medical & \makecell[c]{Combining two color space\\ segment result} & Detection accuracy $\ge$80$\%$   \\
  2010 &  \cite{Moon-2010-ATM,Javidi-2010-DIA} & Microorganisms  & Water-borne & \makecell[c]{Watershed transform, statistical \\sampling}   & -   \\
 2010 & \cite{Daneshpanah-2010-3HI} & Alga & Water-borne & Digital holographic microscopy & - \\
 2010 & \cite{Shin-2010-OSF} & Protozoa & Water-borne & \makecell[c]{Microfluidic device, digital\\ holographic microscope, \\statistical recognition algorithms} & - \\
 
 2011  & \cite{Hiremath-2011-SAI}  & Rotavirus-A  & Medical & Shape features based classification &  Identification rate = 98$\%$  \\
 Date & Refs & Microorganism & Domain & Method & Evaluation \\ \hline
 \multicolumn{6}{c}{Classification based methods} \\ \hline
  2013 &   \cite{Badsha-2013-ACA} & Protozoa & Water-borne & \makecell[c]{Threshold segmentation, roundness \\metric and eccentricity \\based classification} &  Identification rate = 97$\%$  \\
  2013 &  \cite{Coltelli-2013-AAR} & Microalgae  & Water-borne &\makecell[c]{Centroid distance spectrum, \\dissimilarity measure and \\color based classification }  &  Accuracy = 96.6$\%$  \\
 2014 & \cite{Yourassowsky-2014-HTH} &\makecell[c]{ \textit{Giardia lamblia}\\ cysts} & Water-borne & \makecell[c]{High throughput filtering,\\ Fourier transform,\\ inverse Fourier transform }&  -\\
 2014 & \cite{Liu-2014-AOI} &Bacteria & Water-borne &\makecell[c]{ Data of size, shape and \\refractive index based\\ classification }& - \\
 2014 & \cite{Yu-2014-DOI} & Virus & Medical & An optical imaging technique & - \\
  2015 &  \cite{Mader-2015-IST}  &  Fungal  &  Medical & \makecell[c]{Statistical features based \\classification }    &   \makecell[c]{ The total sensitivity = 83$\%$, \\Specificity =79$\%$} \\
  2015 & \cite{Jan-2015-DOT}  & TB  & Medical &  Threshold, edge detection,   & accuracy = 90$\%$   \\

\label{tab:4}  
\end{longtable}
\end{landscape} 

%% file: Traditional.tex
\section{Traditional Machine Learning Based Methods}
\label{Sect:4}
Traditional machine learning, a popular research domain in recent decades, achieves good results in many computer vision tasks. Methods based on traditional machine learning are extensively tried for object detection. In this section, related works on microorganism detection based on traditional machine learning are introduced in chronological order in Sect.\ref{Sect:4-1}. After that, to show the progress of this researches more clearly in recent years, a concise table is prepared in Sect.\ref{Sect:4-2}. 
\par
\subsection{Related Works}
\label{Sect:4-1}
In this subsection, related works based on traditional machine learning are introduced, including motivation, main methods, experimental data and results. In addition, some flowcharts are inserted to illustrate some of the research ideas better. 
\par
In \cite{Yin-2009-RMF}, for detecting bacteria in vegetables, a recognition algorithm  based on Back Propagation Neural Network (BPNN) is proposed. The first step is image segmentation based on iterative thresholds. The second step is to eliminate noise and extract object. Based on the extracted object, multiple morphological features of bacteria are obtained. At last, BPNN is applied for bacteria recognition by taking obtained feature data as input. The accuracy of presented algorithm is judged by comparison with the results of the conventional plate counting method. Experiment shows that in the test of 75 different vegetable samples, the correlation between the results of the two methods is 99.87$\%$. 
\par
In \cite{Ochoa-2010-AIO}, an automatic identification method for \textit{C. elegans} in population images is presented. A ridge segmentation method \cite{Steger-1998-AUD} is applied for image segmentation in the first step. Shape and appearance data are then obtained by applying the open contour. At last, a probabilistic classifier is employed for \textit{C. elegans} recognition. 687 \textit{C. elegans} in 2000 linear objects are prepared for testing. Result shows that the best TP is 95$\%$. 
\par
In \cite{Osman-2010-AGA}, a Genetic Algorithm-Neural Network (GA-NN) algorithm is proposed to assist pathologists in TB diagnosis. The first step is image segmentation, including color filter, moving $k$-means clustering and region growing. The second step is feature extraction based on Hu moment invariant technique and feature selection based on GA. The last step is classification using Neural Network (NN). 120 samples totally containing 360 TB and 600 possible TB are prepared. Among them, 200 TB and 200 possible TB are for training set. The others are equally grouped into the validation set and the test set. Result shows that the highest testing accuracy is 88.57$\%$. 
\par
\begin{figure}[http]
\centering
\includegraphics[width=0.95\linewidth]{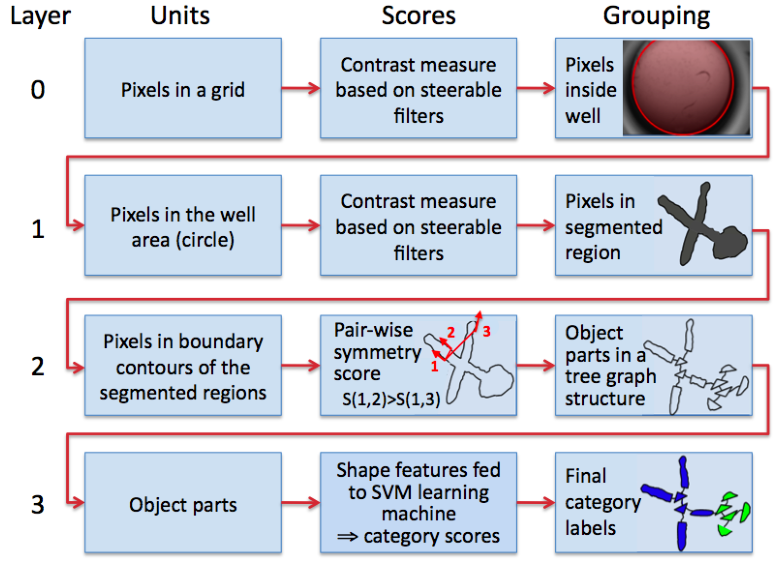}
\caption{The overview of proposed method mentioned in \cite{White-2010-RAA}. The figure corresponds to Fig.2 in the original paper.}
\label{fig:22} 
\end{figure}

In \cite{White-2010-RAA}, a hierarchical approach is presented o measure the number of embryos, larvae and adults in an image. The proposed method is composed of four layers, which are respectively for finding the area of interesting, filtering and segmentation, breaking regions into object parts and object categorization, seen in Fig.\ref{fig:22}. More than 1700 images of \textit{C. elegans} are prepared. Precision and recall for the segmentation and labelling of each developmental stage are evaluated. 
\par
In \cite{Osman-2010-DOM}, for identification of TB in ZN-stained tissue image, an automatic method is described. The first step is initial filter to remove all the colour range except red. The second step is segmentation using moving $k$-mean clustering. In step three, moment invariant feature is extracted. The last step is detection by applying Hybrid Multilayered Perceptron (HMLP) Network. 15 slides are prepared for experiment, each of which produces 30 to 50 images. Experiment results suggest that the proposed method achieve a accuracy of 98.07$\%$ and a specificity of 96.19$\%$.
\par
In \cite{Kumar-2010-RDO}, for identify pathogens in foods, an automatic and rapid detection method is proposed. The first step is background correction and isolating each cell of the treated image into the individual image. The second step is selecting the regions of interesting. After that, various geometrical, optical, and textural parameters of processed images are obtained. At last, the Probabilistic Neural Network (PNN) is designed and built for classifying the microorganisms. In the test on 155 images, PNN can classify the microorganisms using the nine best classification parameters with 100$\%$ accuracy.
\par
\begin{figure}[http]
\centering
\includegraphics[width=0.95\linewidth]{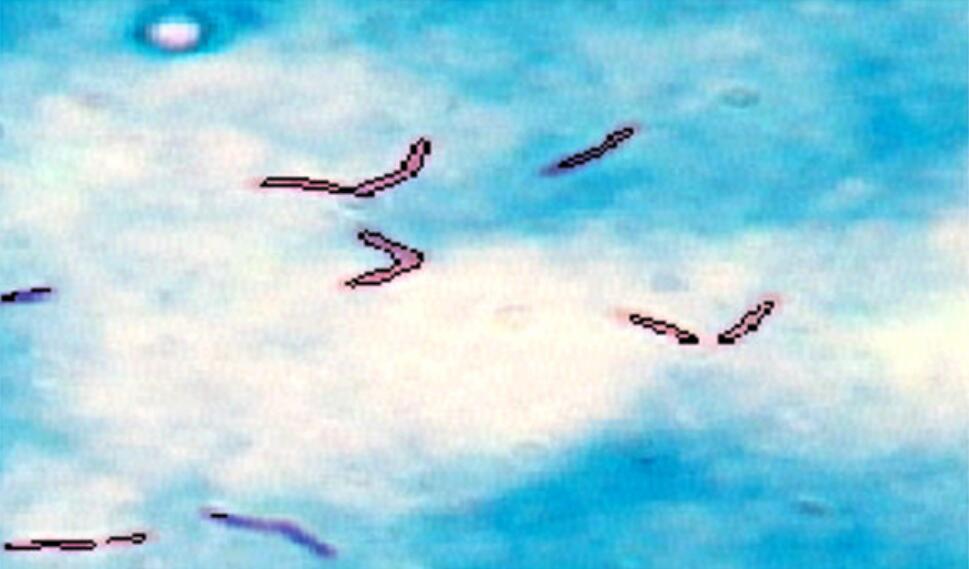}
\caption{The detection results of proposed method in \cite{Khutlang-2010-ADO}. The figure corresponds to Fig.1 in the original paper.}
\label{fig:23} 
\end{figure}

In \cite{Khutlang-2010-ADO}, for TB detection, an automatic and reliable detection method is presented. For detecting possible bacillus, one-class pixel classifier is employed in stage one, where outputs images with color of object. Results of the mixture of Gaussian pixel classifier is shown in Fig.\ref{fig:23}. The purpose of stage two is further segmenting image obtained in stage one. Eight samples containing 1064 positive objects and 1157 negative objects are prepared for test. Experiment result shows that the mixture of Gaussians classifiers performs best with the accuracy of 93.47$\%$.
\par
In \cite{Zhang-2010-AOS}, an efficient detection method is proposed to automatically detect bacterial in foods. The first step is preprocessing, including subtraction between original image and background image, median filtering, gray-level histogram equalization. The second step is binarization on Otsu algorithm. The third step is filling the hollow area by morphological algorithm. At last, a support vector machine (SVM) classifier is applyed to classify the objects into the three categories. 50 images of the first category, 30 images of the second category and 20 images of the third category are prepared as training samples. Result shows that the relative error of results between proposed method and visual counts is less than 3$\%$.
\par
\begin{figure}[http]
\centering
\includegraphics[width=0.95\linewidth]{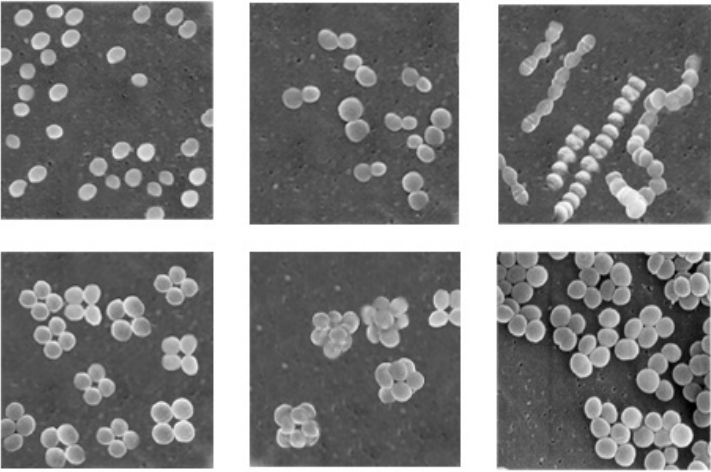}
\caption{The tested objects in \cite{Hiremath-2010-AIA}. The figure corresponds to Fig.2 in the original paper.}
\label{fig:24} 
\end{figure}

In \cite{Hiremath-2010-DIA}, to detection different kinds of cocci bacterial cells, an automatic method is proposed with low time-consuming and high accuracy. The first step is converting the original image into grayscale image and adjusting intensity values. The second step is image segmentation by active contour. The third step is labelling the segment image and computing geometric shape features for each labelled segment. The last step is applying 3$\sigma$ classifier, $k$-Nearest Neighbor ($k$-NN) classifier and NN classifier to the feature set and outputting the classification of identified cells. 100 color images of each phase of bacilli bacterial cell are prepared for testing. Samples of test images are shown in Fig.\ref{fig:24}. Result shows that the NN classifier yields 98$\%$ to 100$\%$ accuracy. In \cite{Hiremath-2010-AIA}, the same method is used by the same research group used for identifying and classifying the bacterial growth phases of bacilli cells. In addition, the Fuzzy classifier is also applied for this experiment. Result shows that the Fuzzy classifier yields 98$\%$ to 100$\%$ accuracy while the NN classifier yields 95$\%$ to 100$\%$ accuracy.
\par
In \cite{Kemmler-2011-DOM}, which is already mentioned in Sect.\ref{Sect:3-1}, a segment-based classification using Conditional Random Fields (CRFs) is proposed. A database including five different microbe species with 40 up to 470 microbes per class is prepared for testing.
\par
\begin{figure}[http]
\centering
\includegraphics[width=0.95\linewidth]{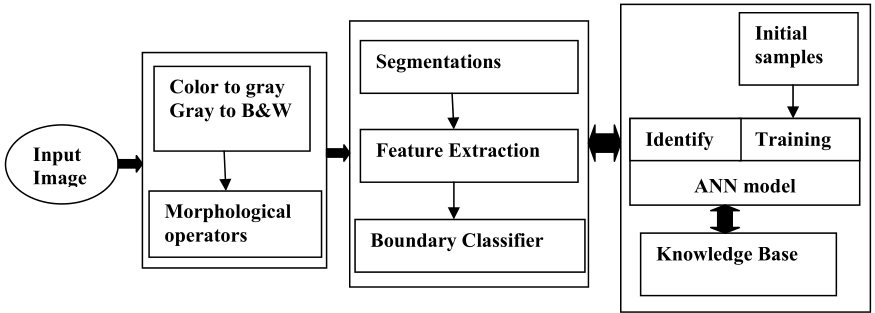}
\caption{The overview of proposed method mentioned in \cite{Mansoor-2011-ARS}. The figure corresponds to Fig.2 in the original paper.}
\label{fig:25} 
\end{figure}

In \cite{Mansoor-2011-ARS}, for automatic detection four types of cyanobacteria genera of freshwater tropical Putrajaya Lake, an artificial neural network (ANN) based system is proposed. Steps of automatic algae recognition system are shown in Fig.\ref{fig:25}. 400 samples of 4 cyanobacteria genera, namely \textit{Microcystis}, \textit{Oscillatoria}, \textit{Anabaena} and \textit{Chroococcus}, are prepared for experiment. Each genus has 100 samples that 80 samples for training and 20 samples for testing. The results illustrate as more than 95$\%$ success in identifying and classification the input samples of 4 genera of cyanobacteria.
\par
\begin{figure}[htbp]
 \centering
 \subfigure[]{
  \label{fig26-a}
  \includegraphics[width=11.5cm,height=2cm]{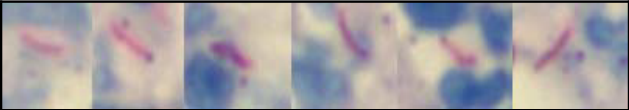}}
 \subfigure[]{
  \label{fig26-b}
  \includegraphics[width=11.5cm,height=2cm]{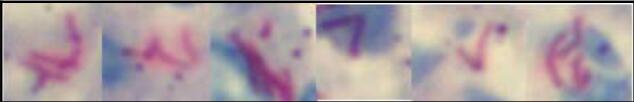}}

 \subfigure[]{
  \label{fig26-c}
  \includegraphics[width=11.5cm,height=2cm]{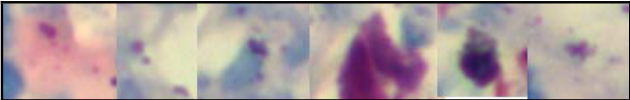}}
 
 \caption{The tested objects in \cite{Osman-2011-TBD}. The figure corresponds to Fig.3 in the original paper.}
  \label{fig:26}
\end{figure}

In \cite{Osman-2011-TBD}, for improving the detection performance and avoiding making false decision of computer-aided TB diagnosis system, an effective classifier is designed. The first step is image segmentation and $k$-means clustering algorithm based TB extraction. The second step is post-processing, including median filter, region growing and area based denoising. At last, a set of six affine moment invariants are calculated for every segmented region and are then fed into the Single Hidden Layer Feedforward Neural Network (SLFNN) for classifying the segmented regions into three classes: ‘TB’, ‘overlapped TB’ and ‘non-TB’. 1603 objects combined of ‘TB’, ‘overlapped TB’ or ‘non-TB’ in 500 images are used for experiment (seen in Fig.\ref{fig:26}), where 1000 objects are for training and 603 objects are for testing. Result shows that the SLFNN trained with the standard Extreme Learning Machine (ELM) method yields a high accuracy of 75.46$\%$.
\par
In \cite{Osman-2011-CSH}, a SLFNN trained using the standard ELM algorithm is compared with Compact-SLFNN ( C-SLFNN). Result shows that the classification accuracy of standard ELM is better than that of C-SLFN. However, standard ELM requires more hidden nodes than C-SLFN. In \cite{Osman-2011-HMP}, the classification performance of Modified Recursive Prediction Error-ELM (MRPE-ELM) trained HMLP Network, MRPE trained HMLP Network and ELM trained SLFNN are compared together. Results suggest that MRPE-ELM has a better classification performance compared to MRPE algorithm.
\par
In \cite{Hiremath-2012-SBC}, to detect different kinds of spiral bacteria, an automatic method is proposed with low time-consuming and high accuracy. The first step is converting the original image into grayscale image and adjusting intensity values.  The second step is image segmentation by active contour. The third step is labelling the segment image and computing geometric shape features for all marked objects. The last step is applying 3$\sigma$ classifier, $k$-NN classifier, NN and the Neuro Fuzzy classifier to the feature set and outputting the classification of identified cells. 300 color samples with three kinds of bacterial are prepares. Result shows that NN classifier as well as Neuro Fuzzy classifier yields a satisfied accuracy of 100$\%$ on prepared data.
\par
In \cite{Ding-2012-RDB}, a rapid detection system with low detection costs is designed to meet the demand for rapid detection of \textit{E. coli}. The first step is preprocessing, including creating a new memory area, grayscale the image, median filter denoising and threshold segmentation. Feature parameters are then extracted, including shape and color feature parameters. At last, Principal Component Neural Network (PCNN) is constructed and applied for \textit{E. coli} detection. The results suggest that PCNN achieves a prediction accuracy of 91.33$\%$ on non-training samples.
\par
In \cite{Mosleh-2012-APS}, by combining image processing and ANN, an automatic approach for detecting some special freshwater algae genera is proposed. The first step is image preprocessing, including image resizing, image enhancement and noise removing. The second step is image segmentation based on Canny edge detection. In step three, shape and texture features are extracted and then normalized by applying principal component analysis. At last, Multilayered Perceptron (MLP) trained with back propagation error algorithm ANN is employed for classification. Experiment shows that the proposed method successfully detects 93 of 100 prepared samples.
\par
\begin{figure}[http]
\centering
\includegraphics[width=0.95\linewidth]{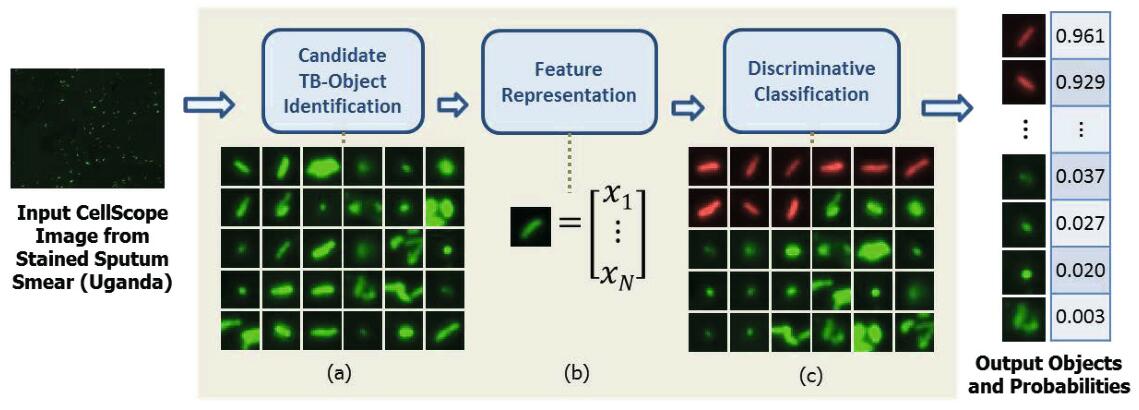}
\caption{The overview of proposed method mentioned in \cite{Chang-2012-ATD}. The figure corresponds to Fig.3 in the original paper.}
\label{fig:27} 
\end{figure}

In \cite{Chang-2012-ATD}, for detecting TB automatically, an algorithm with high accuracy is proposed, which is consisted of candidate TB identification step, feature representation step and discriminative classification step. Fig.\ref{fig:27}  shows the block diagram of designed algorithm. Experts mark TB objects in 92 of the 296 positive TB images, resulting in 1597 positive TB-objects. Each of the candidate objects is classified by an intersection kernel (IK)SVM, yielding an average precision of 91.3$\%$.
\par
\begin{figure}[http]
\centering
\includegraphics[width=0.95\linewidth]{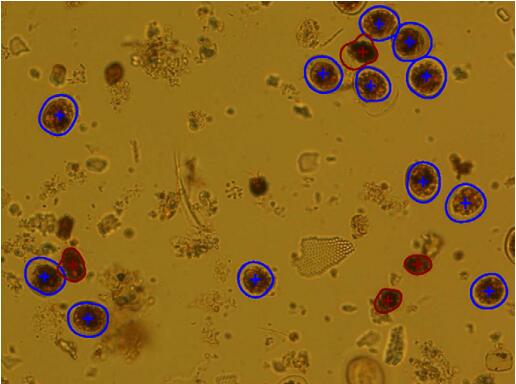}
\caption{The detection result of proposed method in \cite{Verikas-2012-PCD}. The figure corresponds to Fig.12 in the original paper.}
\label{fig:28} 
\end{figure}

In \cite{Verikas-2012-PCD}, a new technique is presented for detecting P. minimum species. In the first step, image edges is enhancement by applying phase congruency-based enhancement of image edges \cite{Kovesi-2000-PCA}. The second step is stochastic optimization-based object contour determination. At last, an automatic detection system combined of a Gaussian kernel SVM and a RF classifier is designed. Fig.\ref{fig:28} exhibits the final detection result, where all P. minimum cells are detected. Experiment on 2088 P. minimum in 114 images shows that the proposed system achieves an detection accuracy of 93.2$\%$.
\par
In \cite{Zhai-2012-ROM}, to reduce the workload of doctors in diagnosing TB, a color and gradient feature based image segmentation and recognition algorithm is proposed. The algorithm mainly consists of two steps. One is image segmentation, including pre-segmentation, adaptive segmentation and fusion segmentation. The other is image classification. Firstly, two gradient features and five shape features are extracted. Feature vector is then generated based on obtained features. At last, a Bayesian classifier is applied for classification. Experimental result shows that the object recognition rate of this algorithm can reach 91$\%$ on 100 images from different tuberculosis patients.
\par
In \cite{Li-2013-AMA}, for recognition of environmental microorganisms (EMs), a framework of content-based image analysis is proposed with low cost and low time consumption. The first step is image segmentation. By comparing results of six segmentation approaches and considering the actual work needs, a semi-automatic segmentation approach employing Sobel edge detector is finally selected for the first step. The second step is shape feature description based on edge histograms, fourier descriptors, extended geometrical features, as well as internal structure histograms. At last, a multi-class SVM is applied for classification based on above features. The experiment is tested on a dataset with ten classes of EMs containing 20 images each. Ten samples from each class are for training, and the left 10 are for testing. Result shows that the best classification rate is 89.7$\%$.
\par
In \cite{Santiago-2013-AAS}, an automatic screening systems is designed to detect agent in samples and make comprehensive decisions about subjects (e.g. ill/healthy) based on these tests. The system mainly consists of two stages. One is a classification stage. First, image is divided into many patches, which are then selected based on minimum green-color. Second, Canny edge detection is used for image segmentation. Third, a set of rotation and translation invariant features are extracted from each candidate object. At last, patches are classified by applying a SVM classifier. The other is a comprehensive decisions stage by applying a Bayesian methodology. The training set contains 34 TB-negative subjects and 11 TB-positive subjects while the test set contains 15 TB-negative subjects and 13 TB-positive subjects. Experiment result shows that the sensitivity of TB classifier is 73.53$\%$. 
\par
In \cite{Xu-2014-DOT}, to improve the efficiency of doctor diagnosis, a computer image processing based algorithm of TB detection is proposed. The first step is preprocessing based on $k$-means algorithm. The second step is feature extraction. At last, Gaussian mixture model is trained with maximum expectation algorithm and then applied for classifying samples. The classification results of all data show that the classification sensitivity is 66.7$\%$, and the accuracy is 96.2$\%$. 
\par
In \cite{Zetsche-2014-IDH}, a system combined DHM and imaging-in-flow is described for the detection and classification of planktonic organisms. The first step is object localization from phase images, including classical threshold based XY-localization, robust refocusing criterion based Z-localization and re-segmentation based refinement of XY-coordinates. The second step is feature extraction containing texture features from both intensity images and phase images and morphological features from phase images. The last step is classification based on an SVM classifier. Results suggest that the correct prediction rate over all test data is 92.4$\%$.
\par
\begin{figure}[http]
\centering
\includegraphics[width=0.95\linewidth]{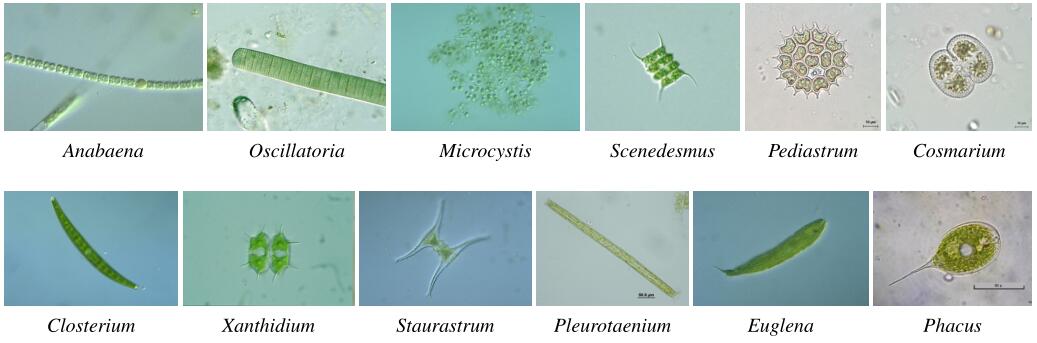}
\caption{The tested objects in \cite{Promdaen-2014-AMI}. The figure corresponds to Fig.1 in the original paper.}
\label{fig:31} 
\end{figure}

In \cite{Promdaen-2014-AMI}, a new method is proposed to detect 12 common microalgae in water resources of Thailand automatically. The first step is Sobel and Canny edge detection. Then, processed images are classified into rod-shaped and non-rod-shaped by sequential minimal optimization (SMO) classifier based on features of the axial ratio, the convex area and the area ratio. For rod-shaped algae image, the multi-resolution edge detection method is applied to re-segment. At last, SMO classifier is used for final classification. Experiment is performed on the data set that includes twelve genera of microalgae (seen in Fig.\ref{fig:31}), 60 images for each genus (45 for training, 15 for testing). Experiment results suggest that proposed algorithm yields an accuracy of 97.22$\%$.
\par
In \cite{Yang-2014-SCO}, for recognition of EMs, a framework of content-based image analysis is proposed with low cost and low time consumption. In addition, a novel 2-D feature descriptor is especially introduced for EM shapes. The first step is image segmentation by employing a Sobel edge detector. In the second step, the designed shape descriptor is applied for feature extraction. At last, a multi-class SVM is used for classification. The experiment is tested on a dataset with ten classes of EMs containing 20 images each. Ten samples from each class are for training, and the left 10 are for testing. Based on proposed feature descriptor, the overall classification rate increases by 2.8$\%$ from 89.7$\%$.
\par
In \cite{Verikas-2014-AIA}, an integrated approach is proposed for P. minimum detection. Firstly, histogram-based image binarization technique is applied to process the original images. Then, features for classification are determined, categorized and computed. At last, a committee combined of SVM and RF is applied to make a decision. The proposed method is performed on 158 images with 920 P. minimum cells. Result shows that the proposed method yields an overall recognition rate of 97$\%$ for P. minimum cells.
\par
In \cite{Nugroho-2015-FEA}, for detection of malaria parasite cell, a method based on image processing is proposed. The first step is image enhancement, which contains contrast stretching and median filter. The second step is image segmentation based on $k$-means algorithm. Histogram-based texture is then extracted for the last step, in where multilayer perceptron backpropagation algorithm is applied for final classification. The prepared data set contains of 60 images which are grouped into trophozoite, schizont and gametocyte. The results suggest that this algorithm achieves a good detection performance in the prepared data with accuracy of 87.8$\%$ and specificity of 90.8$\%$.
\par
\begin{figure}[http]
\centering
\includegraphics[width=0.95\linewidth]{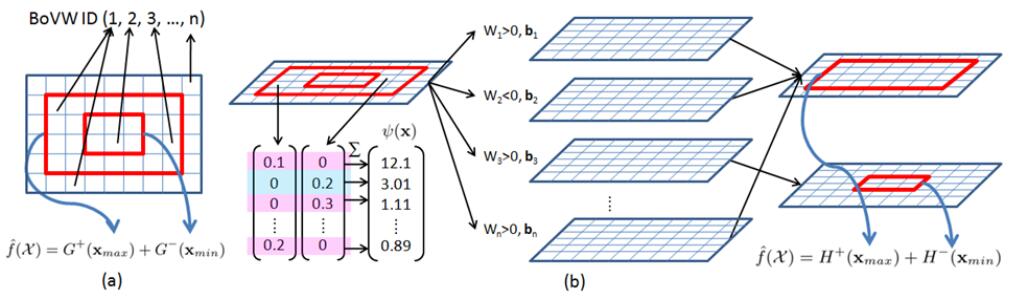}
\caption{The architecture of RBSVM mentioned in \cite{Li-2015-EMC}. The figure corresponds to Fig.2 in the original paper.}
\label{fig:32} 
\end{figure}

In \cite{Li-2015-EMC}, an EMs classification method is described, which solves the problem of small training data sets and noisy images. The presented algorithm contains three steps as following: sparse coding features extraction, region-based (RB)SVM designing and final localization and classification. Fig.\ref{fig:32} shows the comparison of basic and improved RBSVM. The database contains 15 classes of microbes, each with 20 images (10 for training, 10 for testing). Mean of average precisions (MAP) is chosen as the evaluation measure. Result shows that the MAP of (RBSVM$+$NNSC) is also higher than that of (RBSVM$+$BoVW).
\par
In \cite{Shan-2015-ATF}, an anisotropic tubular filtering (ATF) based algorithm is proposed for automatic detection of TB. Firstly, ATF is employed for image enhancement, which is mainly increasing the contrast between objects and background. The possible acid-fast bacillis (AFBs) are then segmented based on color threshold. Fourier descriptors and Hu’s moment of each candidate AFB are then calculated. Finally, a region-based function (RBF)SVM is employed to classify each candidate AFB according to calculated features. Among 300 samples, 161 are marked as positive and the other 139 are marked as negative by experts. In addition, 180 images are grouped into training set while others are grouped into testing set. Comparative result shows that proposed method with Fourier descriptor has the highest sensitivity, F1-Score and accuracy.
\par
In \cite{Dannemiller-2015-ANM}, a new method based on Retinex and SVM is presented for the segmentation of alga images. The proposed algorithm mainly contains two steps. One is improving image quality by applying the Retinex filtering technique. The other is segmenting algae in the image based on SVM. In addition, 100 samples containing only algae and 100 samples containing only background are obtained from prepared 32 images. Result shows that presented algorithm yields an overall detection rate of over 95$\%$.
\par
In \cite{Sajedi-2019-ASR}, for detecting actinobacterial species, an algorithms based on principle component analysis (PCA) and MLP is presented. The first step is data augmentation by applying Gaussian blurring. In the second step, feature vector is generated, including wavelet transform and dimension reduction. At last, feature vectors obtained above are classified by MLP. The experiment is performed on the database, named UTMC.V2.DB. In addition, classes with less than eight images are not considered. Results suggest that the accuracy is about 80.5$\%$.
\par
In \cite{Dhindsa-2020-AEM}, for identification of algae in water bodies, some classification algorithms are used, separately. Firstly, the gray scale images are subjected to the pixel clustering. Boundaries of the microbes are then extracted by using Otsu method and Kirsch filter. Thirdly, the object features corresponding to different classifiers are extracted. At last, the following classifiers are used for final classification: comparing Classification and Regression Trees (CART), $k$-NN, Gaussian Naive Bayes, Linear Regression, Linear discriminant analysis and SVM. The experiment to test the performance of the classifiers is carried out on 10 kinds of algae. Experimental evaluation and classification algorithm research show that the CART algorithm is the most suitable. 
\par
\subsection{Summary}
In recent decade, detection methods based on traditional machine learning are increasingly used for microorganism detection. The total related works are shown in Tab.\ref{tab:5}, where the publication date, literature links, research objects, main methods and evaluation indicators are displayed.
\label{Sect:4-2}

\begin{landscape} 
\begin{longtable}{c|c|c|c|c}

\caption{Summary for object detection method based on traditional machine learning.}\\
\hline\hline
\endfirsthead
\caption{Continued}\\
\hline\hline
\endhead

\hline

Date & Refs & Microorganism & Method & Evaluation \\ \hline

2009 & \cite{Yin-2009-RMF} & Bacteria & \makecell[c]{ Iterative thresholds, \\BPNN} &  Correlation = 0.99.87$\%$ \\
 2010 & \cite{Ochoa-2010-AIO} & \textit{C. elegans} & \makecell[c]{ Open contour, \\aprobabilistic classifier} &  TP = 95$\%$ \\
2010  & \cite{Osman-2010-AGA} & TB & \makecell[c]{ $k$-means clustering, \\GA-NN }& Accuracy = 88.57$\%$  \\
 2010 &  \cite{White-2010-RAA} & \textit{C. elegans} & A hierarchical approach & Precision vs. recall plots  \\
  2010& \cite{Osman-2010-DOM} & TB & \makecell[c]{ Initial filter, moving\\ $k$-means clustering, \\HLMP Network} &  \makecell[c]{ Accuracy =98.07$\%$,\\ Sensitivity = 100$\%$,\\Specificity = 96.19$\%$} \\
 2010 & \cite{Kumar-2010-RDO} & \makecell[c]{Five\\ microorganisms} & \makecell[c]{ Background correction, \\PNN} &  Accuracy=100$\%$ \\
 2010 & \cite{Zhang-2010-AOS}  & Bacterial & Otsu algorithm, SVM &  Relative error $\le$3$\%$ \\
 2010 & \cite{Khutlang-2010-ADO} & TB & \makecell[c]{ Mixture of Gaussian\\ classifier} &  Accuracy = 93.47$\%$ \\
 2010 & \cite{Hiremath-2010-DIA} & Cocci bacterials & \makecell[c]{ Active contour, 3$ \sigma $ classifier,\\ $k$-NN classifier and NN}  &  \makecell[c]{ Accuracy of NN classifier \\= 98$\%$ to 100$\%$ }\\

 2010 & \cite{Hiremath-2010-AIA} & Bacilli cells & \makecell[c]{ Active contour, 3$ \sigma $ classifier,\\ $k$-NN classifier, NN and\\ Fuzzy classifier} & \makecell[c]{ Accuracy of fuzzy \\classifier = 98$\%$ to 100$\%$} \\
 2011 & \cite{Mansoor-2011-ARS} & \makecell[c]{Four types of \\cyanobacteria}  & ANN model & Classification rate = 95$\%$   \\
2011 & \cite{Osman-2011-TBD,Osman-2011-CSH,Osman-2011-HMP} & TB & \makecell[c]{$k$-means clustering,region\\ growing, SLFNN, ELM,\\ C-SLFNN, MRPE, HMLP} & Accuracy \\

Date & Refs & Microorganism & Method & Evaluation \\ \hline

2011& \cite{Kemmler-2011-DOM} & Microorganisms & CRF &  \makecell[c]{Average recognition\\ rate of \textit{ E.coli} = 85$\%$} \\

2012& \cite{Hiremath-2012-SBC} & \makecell[c]{ Different spiral\\ bacterial cells} & \makecell[c]{ Active contour, 3$ \sigma $ classifier,\\ $k$-NN classifier, NN and\\ Neuro Fuzzy classifier}  &  \makecell[c]{ Accuracy of NN\\ classifier = 100$\%$,  \\Accuracy of Neuro Fuzzy\\ classifier = 100$\%$ }\\
2012& \cite{Ding-2012-RDB} & \textit{E. coli} & \makecell[c]{ Median filter denoising,\\ threshold, PCNN} & Prediction accuracy= 91.33$\%$  \\
2012& \cite{Chang-2012-ATD} & TB & \makecell[c]{ Template matching,\\ IKSVM} & Average precision = 91.3$\%$  \\
2012& \cite{Mosleh-2012-APS} & \makecell[c]{ Freshwater \\algae genera} & \makecell[c]{ Image enhancement,\\ Canny edge detection,\\ MLP, ANN} & Accuracy = 93$\%$ \\
2012& \cite{Verikas-2012-PCD} & P. minimum & \makecell[c]{ Edge enhancement,\\ SVM, RF}  & Accuracy = 93.2$\%$ \\
2012& \cite{Zhai-2012-ROM} & TB & \makecell[c]{ Adaptive segmentation, \\fusion segmentation, \\Bayesian classifier} & Recognition rate =	 91$\%$ \\
2013& \cite{Li-2013-AMA} & EMs & \makecell[c]{Sobel edge detector,\\ multi-class SVM} &  Classification rate = 89.7$\%$ \\
2013& \cite{Santiago-2013-AAS} & TB & \makecell[c]{Threshold,\\ Canny edge detection,\\Bayesian methodology}  & Sensitivity = 73.53$\%$.   \\
2014& \cite{Xu-2014-DOT} & TB & \makecell[c]{$k$-means, \\Gaussian mixture Model} & \makecell[c]{Sensitivity = 66.7$\%$, \\Accuracy = 96.2$\%$ }\\

Date & Refs & Microorganism & Method & Evaluation \\ \hline

2014& \cite{Zetsche-2014-IDH} & \makecell[c]{Planktonic\\ organisms} & \makecell[c]{Robust Refocusing \\Criterion, SVM} &  Prediction rate = 92.4$\%$\\
2014& \cite{Yang-2014-SCO} & EMs & \makecell[c]{Sobel edge detector, \\a new feature descriptor,\\ multi-class SVM}  & Classification rate = 92.5$\%$  \\
2014 & \cite{Promdaen-2014-AMI} & \makecell[c]{12 common\\ microalgae} & \makecell[c]{Sobel edge detection,\\ Canny edge detection,\\ SMO classifier} &  Accuracy = 97.22$\%$ \\
2014& \cite{Verikas-2014-AIA} & P. minimum & \makecell[c]{Histogram-based \\binarization, SVM, RF} & \makecell[c]{Overall recognition\\ rate = 97$\%$ } \\
2015& \cite{Nugroho-2015-FEA} & Malaria parasite cell & $k$-means, MLP & \makecell[c]{Accuracy = 87.8$\%$,\\ Sensitivity = 81.7$\%$, \\Specificity = 90.8$\%$. } \\
2015& \cite{Li-2015-EMC} &EMs & \makecell[c]{RBSVM, sparse coding \\features} &  MAP \\
2015& \cite{Shan-2015-ATF} & TB & ATF, RBFSVM & \makecell[c]{Sensitivity, Specificity, \\F1-Score, Accuracy}  \\
2015& \cite{Dannemiller-2015-ANM} & Algae & \makecell[c]{Retinex filtering technique,\\ SVM }&   \makecell[c]{Overall detection \\rate $\ge$ 95$\%$ }\\
2019& \cite{Sajedi-2019-ASR} &\makecell[c]{Actinobacterial \\strains}  & PCA, MLP & Accuracy = 80.5$\%$.  \\
2020& \cite{Dhindsa-2020-AEM} & Algae & \makecell[c]{Otsu, Kirsch filter , multiple\\ machine learning classifiers} & \makecell[c]{Accuracy, Recall,\\ Precision, F1-score}  \\

\label{tab:5}  
\end{longtable}
\end{landscape} 

%% file: Deep.tex
\section{Deep Learning Based Methods}
\label{Sect:5}
Recently, deep learning develops rapidly and achieves a series of excellent results in image analysis. In this section, we first introduce the advantages of deep learning methods. Then, works related to microorganism detection based on deep learning are summarized. After that, a concise table is designed to show the progress of deep learning in microorganism detection more clearly.
\subsection{Advantages of Deep Learning Methods}
Compared with traditional machine learning methods, deep learning methods have a wide range of applications and strong applicability. Deep learning methods build its network framework by combining multiple simple but nonlinear modules, which enables it to design corresponding module combinations to achieve function mapping according to different problems \cite{Lecun-2015-DP}. In addition, In the feature extraction step of detection processing, traditional machine learning methods apply handcrafted feature engineering methods, which is labor-intensive and time-consuming. Deep learning methods cannot only achieve automatic learning of features through its advanced network structure, but also learn complex features from simple features. Next, the advantages of deep learning for object detection are better represented by introducing some backbones of deep learning networks.
\par
\begin{figure}[htbp]
\centering
\includegraphics[scale=0.35]{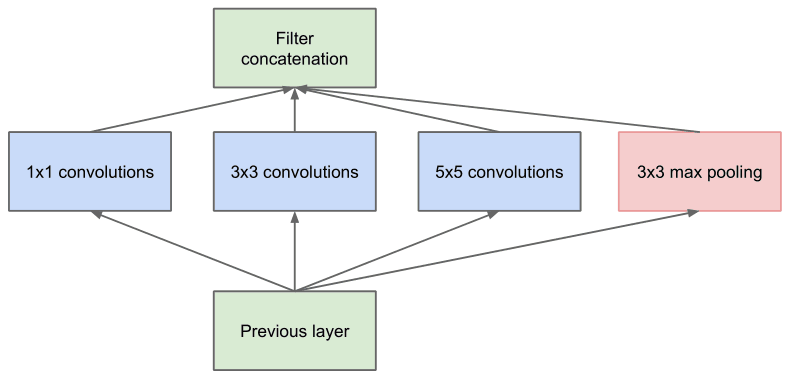}
 \caption{Inception module mentioned in \cite{Szegedy-2015-GDW}. The figure corresponds to Fig.2 in the original paper.}
 \label{fig:3-2}
\end{figure}
\begin{figure}[htbp]
\centering
\includegraphics[scale=0.7]{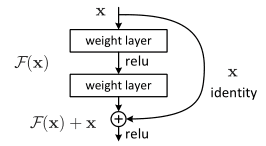}
\caption{Residual block mentioned in~\cite{He-2016-DRL}. 
The figure corresponds to Fig.2 in the original paper.}
\label{fig:3-1}
\end{figure}

In \cite{Simonyan-2014-VDC}, the VGG backbone is proposed, which has two structures: VGG-16 and VGG-19. The deepening of network layers enables VGG to obtain larger feature maps. In addition, the superposition of multiple small convolution kernels enhances the feature learning ability of VGG. Relevant comparative experiments show that VGG can learn more complex and more feature information than the previous traditional machine learning. VGG is the backbone of single shot detector (SSD) \cite{Liu-2016-SSS}, a object detection model with high speed and high accuracy. In \cite{Szegedy-2015-GDW}, GoogLeNet is presented, which is based on Inception modules shown in Fig.\ref{fig:3-2}. The feasibility of obtaining optimal sparse structure with ready-made dense building blocks is confirmed by GoogLeNet. GoogLeNet not only deepens the depth and width of the network, but also reduces the amount of parameters. In \cite{He-2016-DRL}, a residual learning framework called ResNet is proposed to solve the problem that training difficulty will increase with the deepening of network layer. By employing residual block, shown in Fig.\ref{fig:3-1}, the problem of gradient disappearance that occurs when training deeper network layers is well solved. With the backbone of ResNet, both Faster R-CNN \cite{Ren-2015--FRT} and Mask R-CNN \cite{He-2017-MR} achieve great detection results.
\par
\subsection{Related Works}
In this subsection, related works based on deep learning are introduced, including motivation, main methods, experimental data and results. In addition, some flowcharts are inserted to illustrate some of the research ideas better. 
\label{Sect:5-1}
\par
\begin{figure}[http]
\centering
\includegraphics[width=0.95\linewidth]{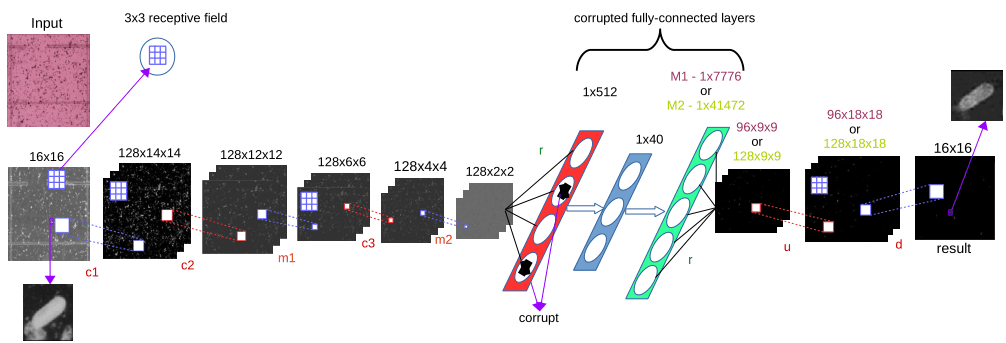}
\caption{The architecture of proposed model mentioned in \cite{Akintayo-2016-AEC}. The figure corresponds to Fig.3 in the original paper.}
\label{fig:33} 
\end{figure}

In \cite{Akintayo-2016-AEC}, a new selective autoencoder algorithm based on deep convolutional networks is designed to detect soybean cyst nematode eggs. Firstly, the proposed deep convolutional network is trained with training patches, where error back-propagation algorithm is applied. secondly, the trained network is tested with test patches and then outputs test results. Convolutional autoencoder architecture for two alternative model structures are shown in Fig.\ref{fig:33}. The bounding box of every soybean cyst nematode egg in 644 images are extracted and stored for experiment, of which 80$\%$ are used for training and 20$\%$ are used for verification. Result shows that the average detection accuracy of proposed approach is 94.33$\%$.
\par
\begin{figure}[http]
\centering
\includegraphics[width=0.95\linewidth]{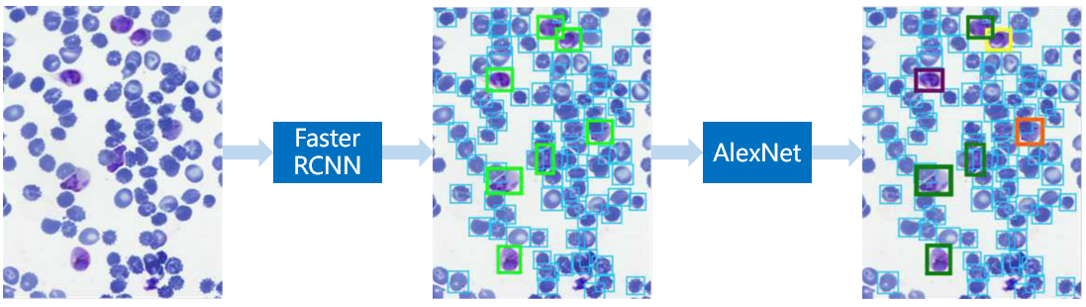}
\caption{The overview and detection result of proposed method mentioned in \cite{Hung-2017-AFR}. The figure corresponds to Fig.4 in the original paper.}
\label{fig:34} 
\end{figure}

In \cite{Hung-2017-AFR}, a faster region-convoluational neural network (Faster R-CNN) is firstly employed for detecting cell in blood and determining their stages. Firstly, a traditional segmentation method and machine learning are employed for this task. Based on their performance, the baseline is established. After that, a two-stage detection and classification approach is selected, seen in Fig.\ref{fig:34}. In stage one, Faster R-CNN is applied for classifying bounding boxes around objects as red blood cell (RBC) or non-RBC. In stage two, non-RBCs detected in stage one are sent to AlexNet for further classification. 1300 images containing around 100000 single labelled cell is prepared. Experiment results suggest that the two-stage algorithm yields a total accuracy of 98$\%$.
\par
\begin{figure}[http]
\centering
\includegraphics[width=0.95\linewidth]{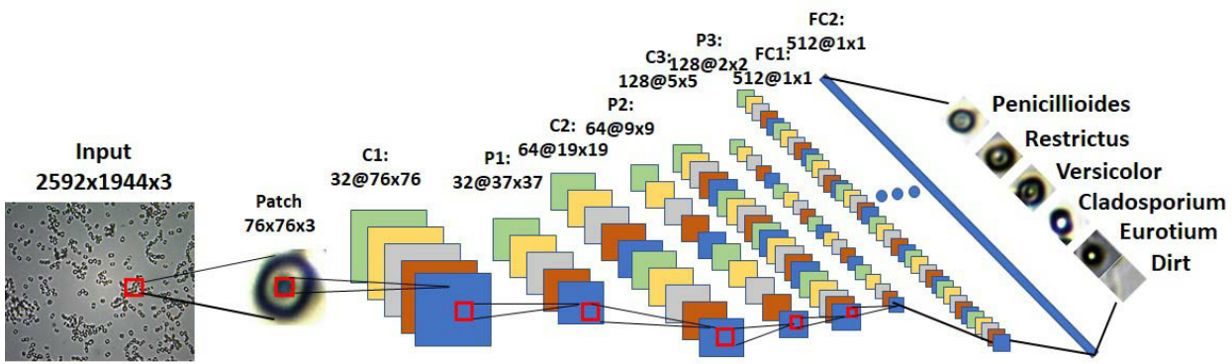}
\caption{The architecture of CNN mentioned in \cite{Tahir-2018-AFS}. The figure corresponds to Fig.5 in the original paper.}
\label{fig:35} 
\end{figure}

In \cite{Tahir-2018-AFS}, a CNN-based approach for detecting fungus and distinguishing different types of fungus is proposed. In addition, a novel fungus dataset consisting of five different types of fungus spores and dirt is developed. The proposed CNN architecture for detecting fungus is summarized in Fig.\ref{fig:35}. The presented method is trained with 30000 images (5000 images for each class) and test on 10,800 images (1800 images for each class). Result shows that the accuracy of proposed method is 94.8$\%$.
\par
\begin{figure}[htbp]
 \centering
 \subfigure[]{
  \label{fig36-a}
  \includegraphics[height=2.95cm,width=11.5cm]{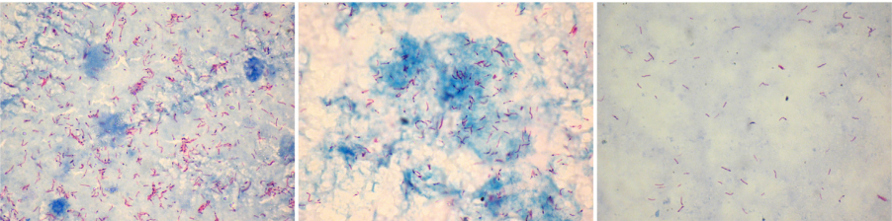}}
 \subfigure[]{
  \label{fig36-b}
  \includegraphics[height=2.95cm,width=11.5cm]{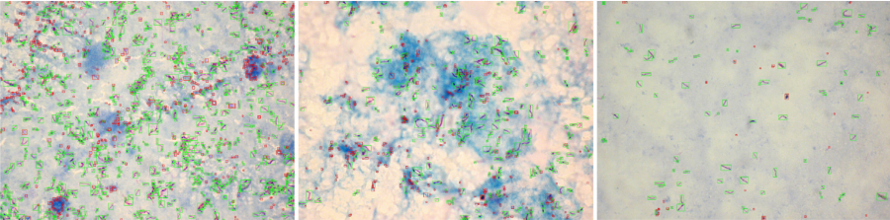}}
\subfigure[]{
  \label{fig36-c}
  \includegraphics[height=2.95cm,width=11.5cm]{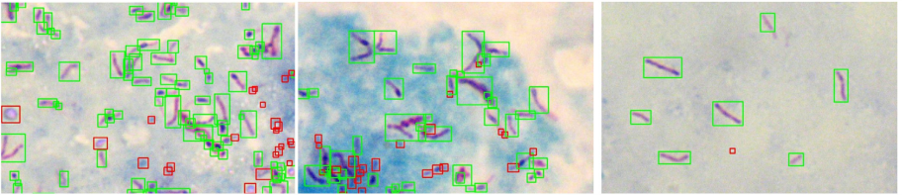}} 
 \caption{The detection results of proposed method in \cite{Panicker-2018-ADO}. The figure corresponds to Fig.5 in the original paper.}
  \label{fig:36}
\end{figure}

 
In \cite{Panicker-2018-ADO}, for detecting TB in microscopic sputum smear images, a CNN-based detection method is presented. The proposed method mainly consists of two stages. One is an image binarization stage by using Otsu threshold algorithm. The other is a pixel classification stage by applying CNN to determine the class of regions extracted from stage one. 22 microscopic images containing high-density and low density images are prepared. Segmentation and classification results of TB shown in Fig.\ref{fig:36}suggest that the presented method yields a recall of 97.13$\%$, a precision of 78.4$\%$ and a F-score of 86.76$\%$.
\par
In \cite{Treebupachatsakul-2019-BCU}, a CNN based system for bacteria detection is proposed to reduce the analysis time and increase the accuracy of diagnostic process. 400 sample images of \textit{Staphylococcus aureus} (\textit{S. aureus}) and 400 sample images of \textit{Lactobacillus delbrueckii} (\textit{L. delbruekii}) are collected. Each type images are separated into 20$\%$ training datasets and 80$\%$ test datasets. Result shows that the best validation accuracy is 96$\%$.
\par
\begin{figure}[htbp]
 \centering
 \subfigure[]{
  \label{fig37-a}
  \rotatebox{90}{\includegraphics[scale=0.35]{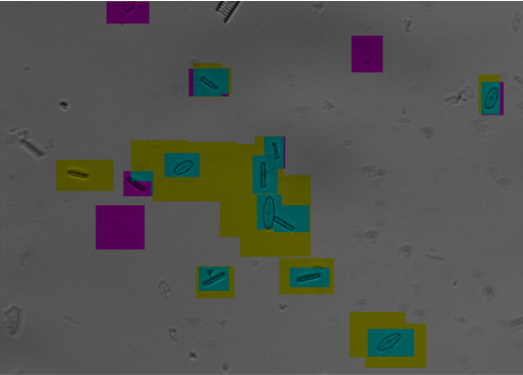}}
  }
 \subfigure[]{
  \label{fig37-b}
  \rotatebox{90}{\includegraphics[scale=0.35]{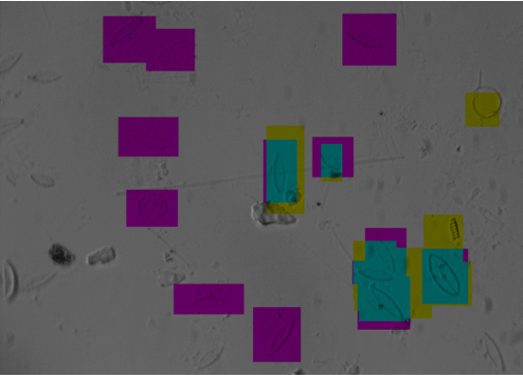}}
  }\subfigure[]{
  \label{fig37-c}
  \rotatebox{90}{\includegraphics[scale=0.35]{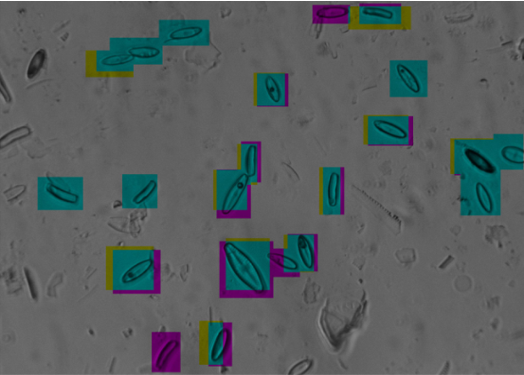}}}
   \caption{The detection results of proposed method in \cite{Pedraza-2018-LAP}. The figure corresponds to Fig.6 in the original paper.}
  \label{fig:37}
\end{figure}

In \cite{Pedraza-2018-LAP}, the performance of R-CNN and you only look once (YOLO) in diatom detection is compared. The R-CNN based method consists of four fundamental steps as following, generating edge boxes for region proposal, rejecting proposed regions, classification of regions and box merging. The YOLO based detection method also contains four steps. Firstly, the input image is divided into a cell matrix and each cell proposes a fixed number of candidate regions. The second step is moving each box to fit to a candidate object. After that, each box is classified and provided with a confidence score. At last, most of boxes are rejected using a threshold. The two methods are test on 11000 images from the 10 species. Detection results of RCNN and YOLO are shown in Fig.\ref{fig:37}. Result shows that YOLO has better performance in diatom detection with a top F-measure of 84$\%$ compared to R-CNN. 
\par
In \cite{Sajedi-2019-ASR}, two deep learning based method is proposed to detect actinobacterial species on solid culture plates. In the first method, CNN is applied for actinobacterial strains detection. The second method consists of two steps. One is a data augmentation step, including blurring, cropping, and horizontal rotation. The other is a classification step based on ResNet, which training mechanism is transfer learning. Two dataset respectively named UTMC.V1.DB and UTMC.V2.DB are prepared. UTMC.V1.DB contains 703 images from 55 different classes. UTMC.V2.DB contains 1303 images from 97 different classes. Result shows that the first method achieves up to 80.81$\%$ and 84.81$\%$ accuracy on UTMC.V1.DB and UTMC.V1.DB, respectively. The second method yields a accuracy of 90.24$\%$ in UTMC.V1.DB and a accuracy of 85.96$\%$ in UTMC.V2.DB.
\par
\begin{figure}[http]
\centering
\includegraphics[width=0.95\linewidth]{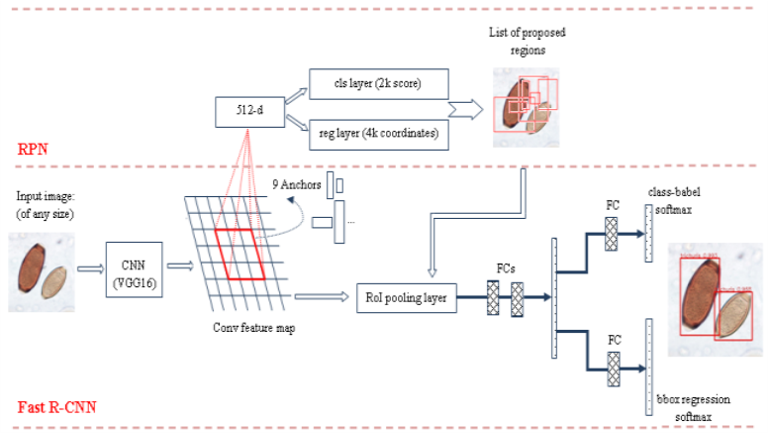}
\caption{The architecture of Faster R-CNN mentioned in \cite{Viet-2019-PWE}. The figure corresponds to Fig.2 in the original paper.}
\label{fig:38} 
\end{figure}

In \cite{Viet-2019-PWE}, to detect parasite eggs, an automatic algorithm by applying Faster R-CNN is proposed. Fig.\ref{fig:38} displays the architecture of Faster R-CNN. In addition, the decision metric of bounding box regression is that the intersection over union must be greater than 70$\%$. Images obtained from stool samples of patients are prepared for testing. Result shows that the mAP of Faster R-CNN based detection is very high with 97.67$\%$. 
\par
\begin{figure}[http]
\centering
\includegraphics[width=0.95\linewidth]{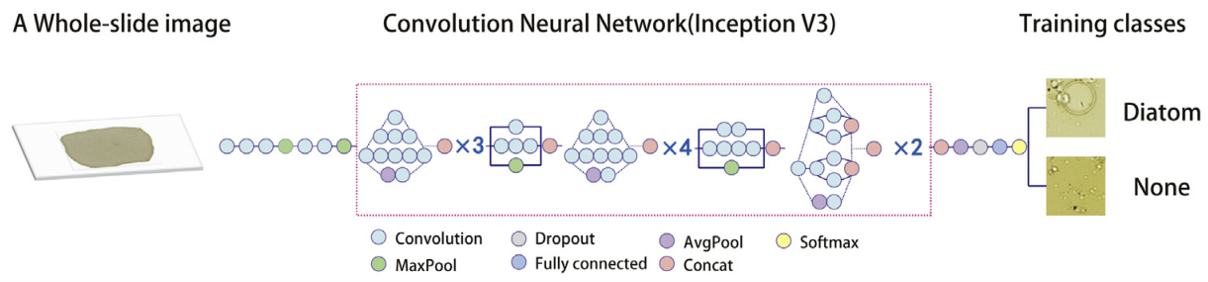}
\caption{The architecture of Inception V3 mentioned in \cite{Zhou-2019-DWI}. The figure corresponds to Fig.1 in the original paper.}
\label{fig:41} 
\end{figure}

In \cite{Zhou-2019-DWI}, an automatic detection system is proposed for diatoms classification. For the proposed method, a GoogLeNet Inception V3 architecture is applied and trained to identify diatoms. The architecture of employed model is shown in Fig.\ref{fig:41}. The sensitivity and specificity are employed to evaluate the performance of proposed method. In this experiment, 43 slide images are collected for training and 10 slide images are collected for validation. Result shows that the successfully identification rate of region of interesting (ROI) reaches 89.6$\%$.
\par
\begin{figure}[http]
\centering
\includegraphics[width=0.95\linewidth]{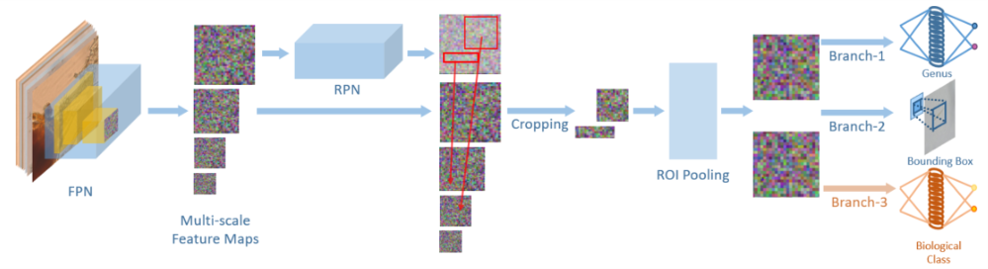}
\caption{The architecture of model mentioned in \cite{Qian-2020-MDL}. The figure corresponds to Fig.2 in the original paper.} 
\label{fig:39} 
\end{figure}
\begin{figure}[http]
\centering
\includegraphics[width=0.9\linewidth]{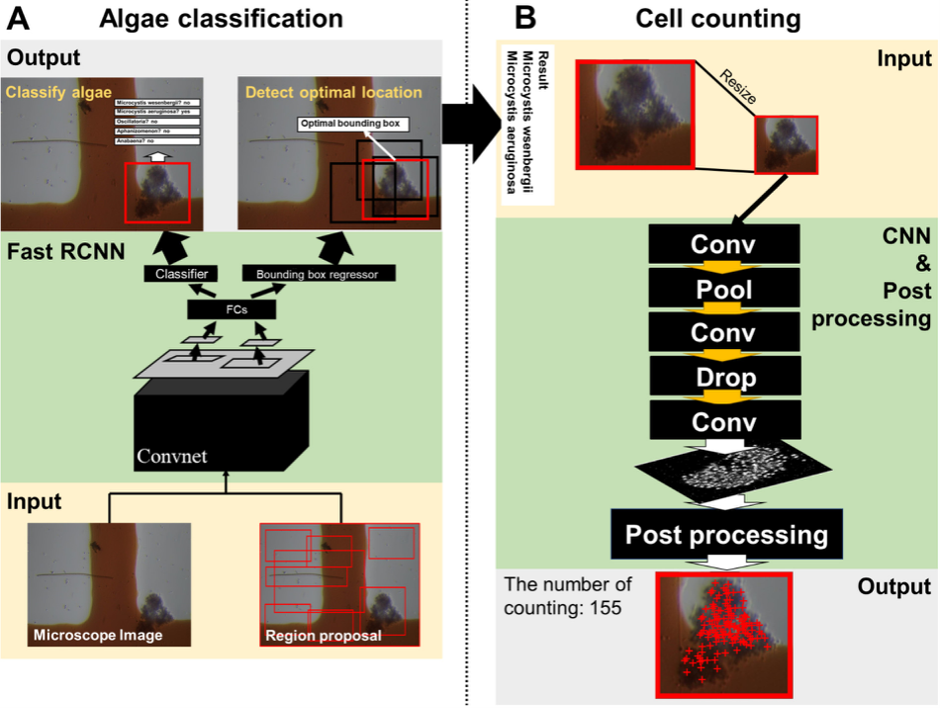}
\caption{The overview of proposed method mentioned in \cite{Baek-2020-IAE}. The figure corresponds to Fig.2 in the original paper.}
\label{fig:40} 
\end{figure}

In \cite{Qian-2020-MDL}, an automatic method based on Faster R-CNN is presented for algal detection. The proposed framework is depicted in Fig.\ref{fig:40}. As seen in Fig.\ref{fig:39}, the origin Faster R-CNN is extended based on adding extra classification branches. A data set containing 1859 samples of 37 algae in six biological categories as well as annotations of genera and classes is prepared. In this experiment, algae no more than 10 images are all classified as other genus. Therefore, 27 genera of algae is applied. 80$\%$ images of each genera is grouped into training set, while the remaining images is grouped into testing set. Result shows that the proposed method achieve 74.64$\%$ mAP on detection at genus level, and 81.17$\%$ mAP at class level. 
\par
In \cite{Baek-2020-IAE}, a Fast R-CNN and CNN based system is deigned to detect five cyanobacteria. The proposed system combines with classification part and counting part. The overview of proposed method in algae detection is shown in Fig.\ref{fig:40}. Two hundred images are obtained for each of the five cyanobacteria (\textit{Microcystis aeruginosa}, \textit{Microcystis wesenbergii}, \textit{Dolichospermum}, \textit{Oscillatoria}, and \textit{Aphanizomenon}). The precision–recall index is applied for evaluating classification performance, while the coefficient of determination and root-meansquared error are selected to determine the accuracy of the cell counting performance. Results show that the proposed model yields the highest AP values of 0.973 on \textit{Microcystis wesenbergii} of five cyanobacteria.

\par
\begin{figure}[http]
\centering
\includegraphics[width=0.95\linewidth]{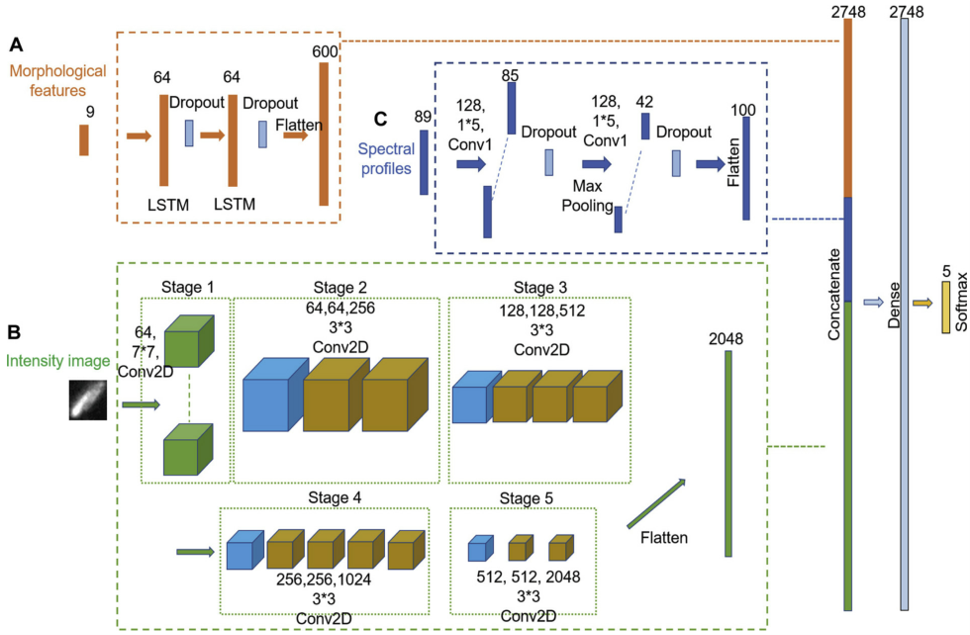}
\caption{The architecture of Fusion-Net mentioned in \cite{Kang-2020-SCO}. The figure corresponds to Fig.2 in the original paper.}
\label{fig:42} 
\end{figure}

In \cite{Kang-2020-SCO}, a hybrid deep learning framework named as ``Fusion-Net” is proposed for foodborne pathogens detection, consisting of long-short term memory  network, ResNet and 1D-CNN. The Fusion-Net is generated by hyperparameter optimization, multiple deep learning architecture selection, and Fusion-Net construction. The framework of Fusion-Net is shown in Fig.\ref{fig:42}. 5000 bacterial cells of five common foodborne bacterial cultures are prepared, which are randomly grouped into training dataset (72$\%$), validation dataset (18$\%$) and test dataset (10$\%$). Based on Fusion-Net, the classification accuracy is improved up to 98.4$\%$.
\par
In \cite{Salido-2020-ALA}, for choosing the most suitable algorithm for diatom detection, the performances of YOLO and SegNet are compared. The prepared dataset contains 80 species of diatoms, each of which contains dozens to hundreds of images. Through comprehensive comparison and analysis of the specificity, sensitivity and precision of the detection results of the two models, The YOLO model is finally selected and integrated into the microscope software to provide a live diatom detection.
\par
In \cite{Ruiz-2020-SVI}, the performances of semantic segmentation and instance segmentation in detecting diatom are compared. SegNet is selected as the semantic segmentation model to detect diatom. Mask R-CNN is selected as instance segmentation model. A total of 126 images of ten different taxa are analysed. Among them, 105 samples are for training and others are for validation. Compare with semantic segmentation method, the proposed instance segmentation achieves a better performance with a sensitivity of 85$\%$ and a specificity of 91$\%$. In addition, without taking into account the reduction in specificity and precision, semantic segmentation achieves an average sensitivity of 95$\%$.
\par
\subsection{Summary}
With the rapid development of deep learning, more and more researches employ deep learning methods for microorganism detection. The related works are shown in Tab.\ref{tab:6}, where the publication date, literature links, research objects, main methods and evaluation indicators are displayed.
\label{Sect:5-2}

\begin{landscape} 
\begin{longtable}{c|c|c|c|c}

\caption{Summary for object detection method based on deep learning.}\\
\hline\hline
\endfirsthead
\caption{Continued}\\
\hline\hline
\endhead

\hline

Date & Refs & Microorganism & Method & Evaluation \\ \hline
2016 & \cite{Akintayo-2016-AEC} & \makecell[c]{Soybean cyst \\nematode eggs} & \makecell[c]{An end-to-end convolutional\\ selective autoencoder} &  Average accuracy = 94.33$\%$\\
2017&  \cite{Hung-2017-AFR}& RBC & Faster RCNN, ALexNet & Total accuracy = 98$\%$. \\
2018& \cite{Tahir-2018-AFS} & Fungus & CNN &  Accuracy = 94.8$\%$ \\
2018& \cite{Panicker-2018-ADO} & TB &Otsu, CNN  & \makecell[c]{Recall = 97.13$\%$ , \\Precision =78.4$\%$,\\ F-score = 86.76$\%$} \\
2018 & \cite{Pedraza-2018-LAP} & Diatoms & R-CNN, YOLO &  F-measure of YOLO = 84$\%$\\
2019 & \cite{Viet-2019-PWE} & \makecell[c]{Human parasite \\eggs} & Faster R-CNN & MAP = 97.67$\%$ \\
2019&\cite{Treebupachatsakul-2019-BCU}  &Bacteria  & CNN & Accuracy = 96$\%$\\
2019& \cite{Sajedi-2019-ASR} & \makecell[c]{Actinobacterial \\species} & \makecell[c]{CNN, ResNet, transfer\\ learning} & Accuracy \\
2019& \cite{Zhou-2019-DWI} & Diatoms & Inception V3 & Identification rate = 89.6$\%$. \\
2020& \cite{Qian-2020-MDL} & Algal & Faster R-CNN & MAP \\
2020& \cite{Baek-2020-IAE} &Cyanobacteria  &Fast R-CNN, CNN  & AP \\
2020& \cite{Kang-2020-SCO} & \makecell[c]{Foodborne \\pathogens} & Fusion-Net & Classification accuracy = 98.4$\%$. \\
2020& \cite{Salido-2020-ALA} & Diatoms & YOLO, SegNet & \makecell[c]{Specificity, sensitivity, \\precision} \\
2020& \cite{Ruiz-2020-SVI} & Diatoms & SegNet, Mask R-CNN & \makecell[c]{Precision, sensitivity, \\specificity} \\

\label{tab:6}  
\end{longtable}
\end{landscape} 

%% file: Methodology.tex
\section{Methodology Analysis}
\label{Sect:6}
In this section, 
A deep analysis of classical image processing based methods, traditional machine learning based methods and deep learning based methods are respectively compiled in this section. Then, methods that are not limited to the visual transformer can be employed directly or indirectly in microorganism detection. After that, other research fields, where detection methods mentioned in this review have potential to be employed in, are introduced and analysed. Finally a small summary of this section is presented.
\par
\subsection{Analysis of Common Methods}
\label{Sect:6-1}
In this subsection, further analysis of classical image processing based methods, traditional machine learning based methods and deep learning based methods are compiled. For each category of methods, a summary chart is equipped to illustrate the commonly used methods.
\subsubsection{Analysis of Classical Image Processing Based Methods}
\label{Sect:6-1-1}
\begin{figure}[http!]
\centering
\includegraphics[width=0.85\linewidth]{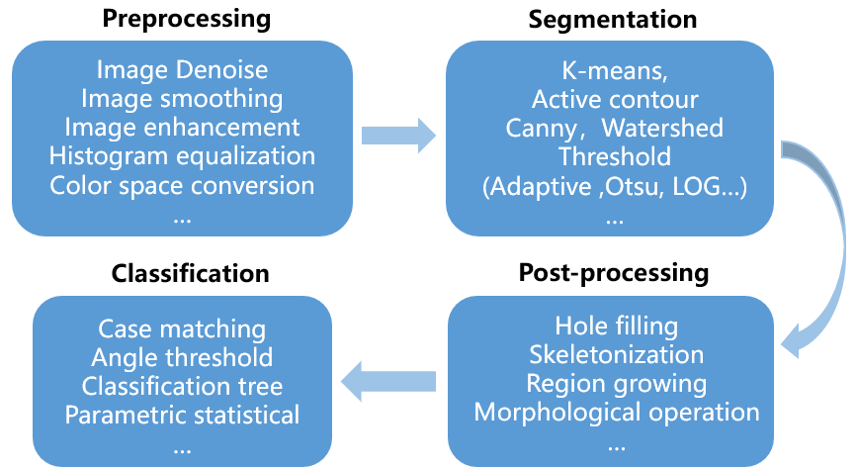}
\caption{Main processing flow and commonly used algorithms of the first 
category of detection methods.}
\label{fig:43} 
\end{figure}

According to the survey on classical image processing based methods for microorganism detection, the detection process of this methods mainly includes four steps: preprocessing, segmentation, post-processing and classification. Fig.\ref{fig:43} shows the main processing flow and the commonly used algorithms. The most widely used segmentation algorithm for detection is threshold segmentation, which has simple calculation and high computational efficiency. The papers involved in based on threshold are \cite{Bloem-1995-FAD}, \cite{Baillieul-1998-ODO}, \cite{Qing-2006-AOM}, \cite{Costa-2008-AIO}, \cite{Zhang-2008-AAB}, \cite{Rizvandi-2008-ELB}, \cite{Rizvandi-2008-SAO}, \cite{Huang-2008-ADA}, \cite{Rizvandi-2008-AID}, \cite{Zhou-2008-UMV}, \cite{Wang-2008-MMT}, \cite{Fernandez-2008-MVA}, \cite{Sotaquira-2009-DAQ}, \cite{Zhai-2010-AIO}, \cite{Raof-2011-ISO}, \cite{Shi-2012-FBA}, \cite{Badsha-2013-ACA}, \cite{Kowalski-2014-ASL}, \cite{Payasi-2017-DAC}, \cite{Javidi-2005-TIA}, \cite{Fernandez-2006-MDI}.
There are many types of thresholds to choose from \cite{Sezgin-2004-SOI}. The more commonly used threshold selection method is Otsu threshold. Its main idea is selecting an optimal threshold automatically from a gray level histogram by a discriminant criterion  \cite{Otsu-1979-ATS}. The papers involved in based on Otsu threshold are \cite{Zhang-2008-AAB}, \cite{Badsha-2013-ACA}, \cite{Rachna-2013-DOT}, \cite{Kurtulmucs-2014-DOD}, \cite{Goyal-2015-ADO}.
\par
The calculation process of the Otsu threshold is simple and has strong generality \cite{Otsu-1979-ATS}. Moreover, satisfactory results can be obtained by employing the Otsu threshold, even if the pixel values of two classes is close \cite{Sezgin-2004-SOI}. However, when the object area is much smaller than the background area, the Otsu threshold cannot provide a good segmentation result \cite{Lee-1990-COA}. The large variances of the object and the background intensities and the small mean difference are responsible for the Otsu threshold to degrade its performance \cite{Lee-1990-ACP}.
\subsubsection{Analysis of Traditional Machine Learning Based Methods}
\label{Sect:6-1-2}
\begin{figure}[http!]
\centering
\includegraphics[width=0.95\linewidth]{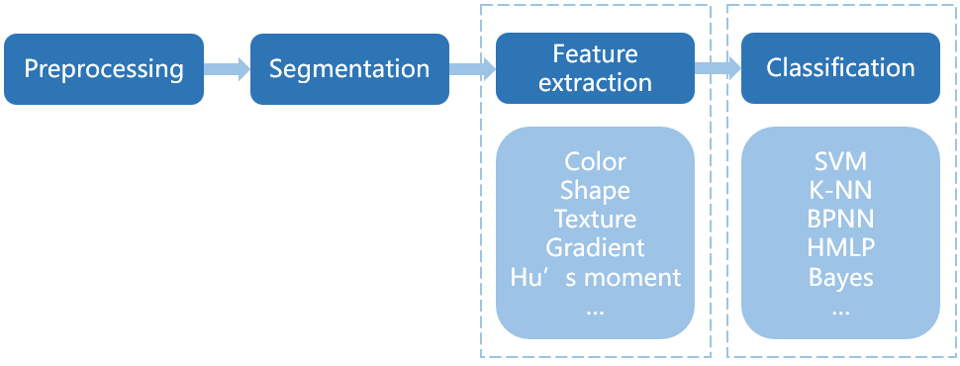}
\caption{Main processing flow and widely used features ang classification algorithms of the second category of detection methods.}
\label{fig:44} 
\end{figure}

Common traditional machine learning method includes SVM, CRF, perceptron, $k$-NN, decision tree and logistic regression model. The main processing flow, widely used features and classification algorithms are shown in Fig.\ref{fig:44}. According to the survey on traditional machine learning methods for microorganism detection, the most widely applied classification model is SVM, mentioned in \cite{ White-2010-RAA}, \cite{Khutlang-2010-ADO }, \cite{Chang-2012-ATD }, \cite{Verikas-2012-PCD }, \cite{Li-2013-AMA}, \cite{Zetsche-2014-IDH }, \cite{Santiago-2013-AAS}, \cite{Li-2015-EMC}, \cite{ Verikas-2014-AIA}, \cite{ Shan-2015-ATF}. SVM is a computer algorithm that learns to assign labels to objects by example \cite{Boser-1992-ATA}. It constructs an optimal separating hyperplane in a higher dimensional space mapped the data into \cite{Suykens-1999-LSS}. In addition, the most widely extracted feature is shape feature, used in \cite{Ochoa-2010-AIO}, \cite{Kumar-2010-RDO}, \cite{Zhang-2010-AOS}, \cite{Hiremath-2010-DIA}, \cite{Mansoor-2011-ARS}, \cite{Kemmler-2011-DOM}, \cite{Osman-2011-TBD}, \cite{ Hiremath-2012-SBC}, \cite{Ding-2012-RDB}, \cite{Chang-2012-ATD}, \cite{Mosleh-2012-APS}, \cite{Verikas-2012-PCD}, \cite{Zhai-2012-ROM}, \cite{Zetsche-2014-IDH}, \cite{Verikas-2014-AIA}. In fact, shape feature is regarded as the most useful traditional feature in the detection of most microorganisms. 

SVM can effectively use smaller training samples. This enables SVM to achieve higher classification accuracy on a smaller training set \cite{Mercier-2003-SVM}. In addition, the idea of selecting the best hyperplane makes SVM have excellent generalization ability \cite{Tzotsos-2008-SVM}. However, SVM is not suitable for solving multi-classification tasks \cite{Noble-2006-WIA}. Moreover, SVM is sensitive to the selection of parameters and kernel functions \cite{Chang-2011-LAL}. This means that different choices will have a great impact on the final classification accuracy of SVM.

\subsubsection{Analysis of Deep Learning Based Methods}
\label{Sect:6-1-3}
\begin{figure}[http!]
\centering
\includegraphics[width=0.95\linewidth]{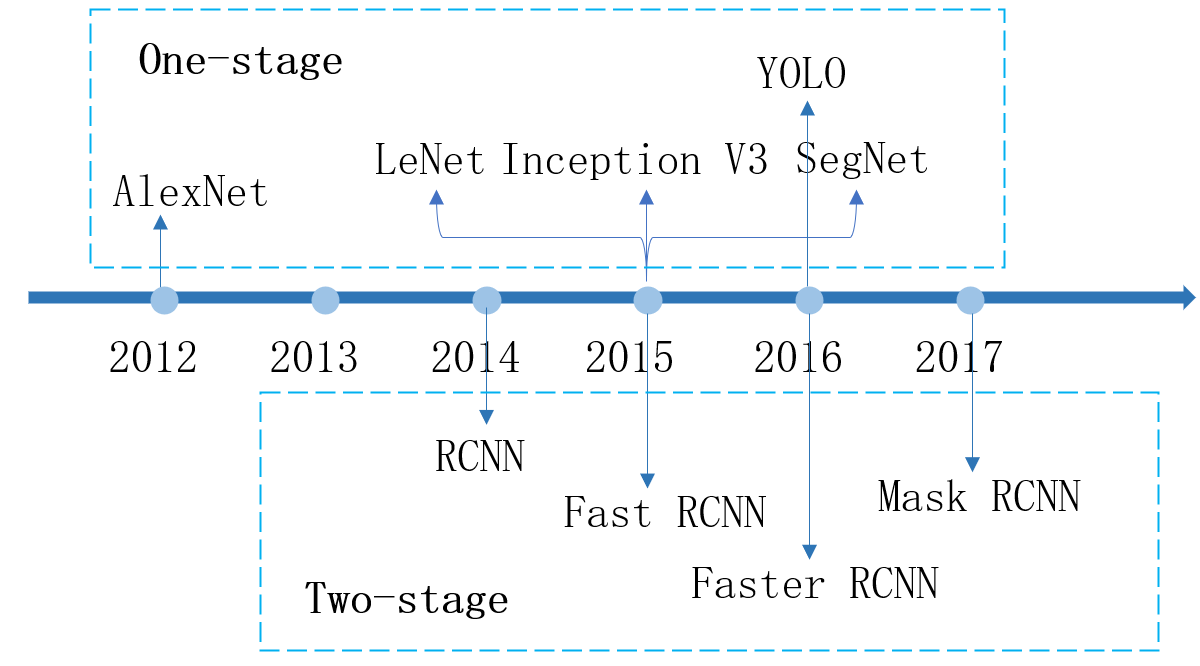}
\caption{Deep learning models used in microorganism detection. The horizontal arrow in the middle is the timeline. Models above time line is one-stage detection model, and the following is two-Stage detection model.}
\label{fig:45} 
\end{figure}

Deep learning is good at handling more complex tasks such as image recognition, object detection, speech recognition, and trend prediction. Fig.\ref{fig:45} shows deep learning models used in microorganism detection and the publication year of each model. As one of the classical deep learning models, CNN and its derivative models are often used in the task of microorganism detection, such as CNN mentioned in \cite{Panicker-2018-ADO}, \cite{Tahir-2018-AFS}, \cite{Sajedi-2019-ASR}, R-CNN mentioned in \cite{Pedraza-2018-LAP}, Faster R-CNN mentioned in \cite{Hung-2017-AFR}, \cite{Viet-2019-PWE}, \cite{Baek-2020-IAE}, \cite{Qian-2020-MDL}, YOLO mentioned in \cite{Pedraza-2018-LAP}, \cite{Salido-2020-ALA}, Mask R-CNN mentioned in \cite{Ruiz-2020-SVI}. Among them, Faster R-CNN is the most commonly used method. Compared to Fast R-CNN, Faster R-CNN is combined with region proposal network (RPN) to break through the bottleneck of regional proposal calculation \cite{Ren-2016-FRT}.
\par

The use of RPN enables Faster R-CNN to not only improve the overall detection efficiency and accuracy, but also truly realize end-to-end detection \cite{Ren-2016-FRT}. Although the detection performance of Faster R-CNN is very good, it cannot perform real-time detection of object. In addition, there is a deviation in the mapping between the feature map of Faster R-CNN and the original image, which is solved by RoIAlign layer in Mask R-CNN \cite{He-2017-MR}. Moreover, the RPN-based region proposal extraction method requires a large amount of calculation.

\subsection{Potential Methods for Microorganism Detection}
\label{Sect:6-2}
The subsection pays attention to some new methods that have not been used in microorganism detection but have the potential. In Sect.\ref{Sect:6-6-1}, the visual transformer is firstly briefly introduced, which has received much attention from computer vision. Moreover, the feasibility of applying computer vision to microorganism detection is analysed. In Sect.\ref{Sect:6-6-2}, other potential methods are introduced and analysed.

\subsubsection{Potential Visual Transformer Methods}
\label{Sect:6-6-1}
The transformer is firstly employed in natural language processing \cite{Devlin-2018-BPO}. Considering the series of achievements made by the transformer architecture in natural language processing, transformer is tried to be applied to computer vision. In fact, due to have a more robust ability of global information representation than the CNNs, it can replace the role of CNNs in vision applications. It turns out that visual transformer methods have achieved good results in many fields of computer vision, such as image segmentation \cite{Chen-2021-TTM,Liang-2020-PDP}, image classification \cite{Chen-2021-CCM,Liu-2020-AUR}, image detection \cite{Aubreville-2017-AGS,Sun-2021-RTS}. Although visual transformer methods show great potential, it have not yet been applied to the field of microorganism detection. Next, some Transformers that have potential applications in microorganism detection will be introduced and analyzed.

\begin{figure}[http!]
\centering
\includegraphics[width=0.95\linewidth]{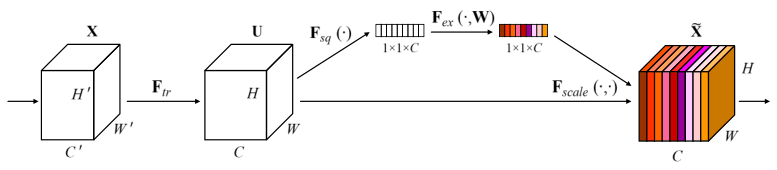}
\caption{The architecture of mentioned in \cite{Hu-2018-SN}. The figure corresponds to Fig.1 in the original paper.}
\label{fig:52} 
\end{figure}

In \cite{Hu-2018-SN}, a new block called squeezeand-excitation (SE) based on channel domain attention mechanism is designed to recalibrate channel-wise feature responses, shown in Fig.\ref{fig:52}. Based on SE block, a SENet is proposed with great generalization. Experiment result shows that SE is able to provide significant performance improvements for existing deep architectures with minimal additional computational cost.

\begin{figure}[http!]
\centering
\includegraphics[width=0.95\linewidth]{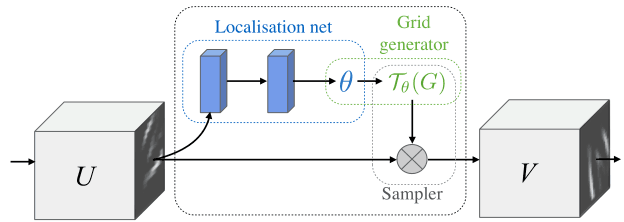}
\caption{The architecture of a spatial transformer module mentioned in \cite{Jaderberg-2015-STN}. The figure corresponds to Fig.2 in the original paper.}
\label{fig:48} 
\end{figure}

In \cite{Jaderberg-2015-STN}, a new learn-able module called spatial transformer based on spatial domain attention mechanism is proposed, which can perform explicit spatial transformations of features in NN. The spatial transformer module can provide an appropriate spatial transformation to each input sample. The architecture of a spatial transformer module is shown in Fig.\ref{fig:48}. Experimental result demonstrates that the participation of spatial transformer can effectively improve the detection performance of the CNNs tested. Based on this, spatial transformer is possible to improve the performance of CNNs in microorganism detection.

In \cite{Devries-2017-DAI}, a dataset augmentation method is proposed, which has nothing to do with domain. The key idea of is constructing a feature space by sequence auto-encoder and then transforming the encoded data by adding noise, interpolating or extrapolating. Experiment result shows that extrapolation is well suited for dataset augmentation in feature space. In addition, the method  provides an idea of data augmentation in microorganism detection
\par
\begin{figure}[http]
\centering
\includegraphics[width=0.82\linewidth]{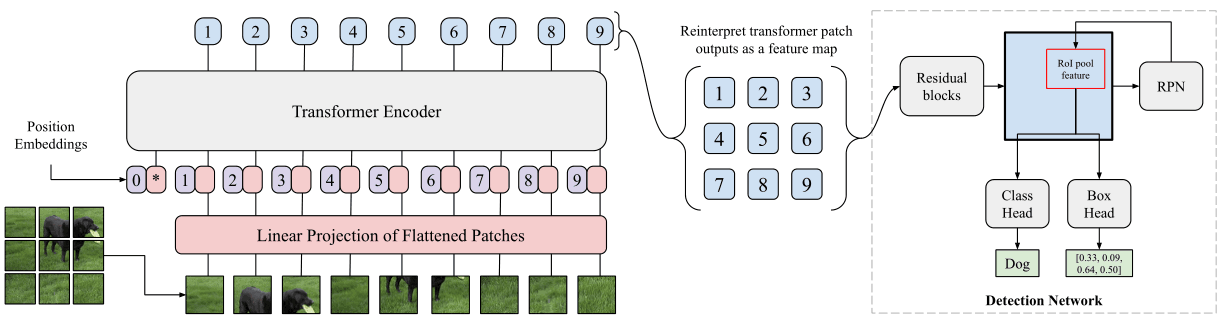}
\caption{The architecture of ViT-FRCNN mentioned in \cite{Beal-2020-TTO}.The figure corresponds to Fig.1 in the original paper.}
\label{fig:49} 
\end{figure}

In \cite{Beal-2020-TTO}, a Vision Transformer-Faster RCNN (ViT-FRCNN) is proposed to confirm the feasibility of Vision Transformer to perform object detection tasks. The key idea of ViT-FRCNN is that ViT is firstly applied to generate a a spatial feature map, which is then fed to Faster R-CNN for final detection. The architecture of ViT-FRCNN is depicted in Fig.\ref{fig:49}. Experiment results on COCO 2017 validation set suggest that ViT-FRCNN can achieve reliable performance on the detection task. This means that ViT-FRCNN cannot only be used for microorganism detection, but also hope to achieve good results. 
\par
In \cite{Woo-2018-CCB}, a convolutional block attention module (CBAM) based on both channel and spatial domain attention mechanism is presented. After obtaining a feature map, CBAM respectively infers attention maps on both channel and spatial domain, which are then delivered to the input feature map. Results suggest that CBAM cannot only integrate into any CNN architectures easily, but also improve the detection performances.
\par
\begin{figure}[http]
\centering
\includegraphics[width=0.95\linewidth]{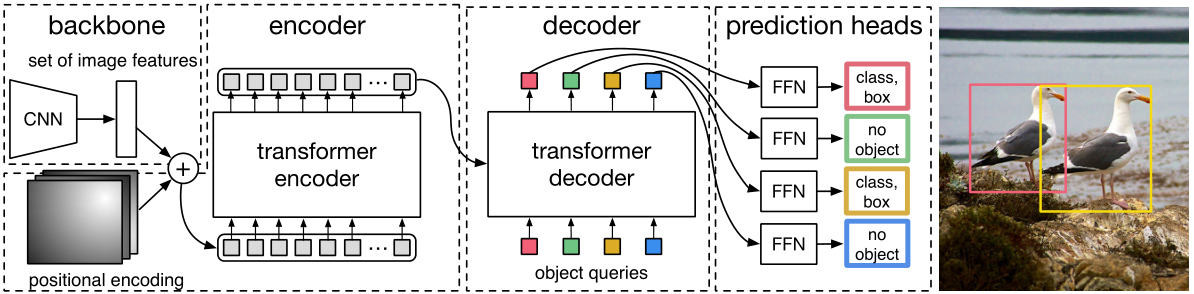}
\caption{The architecture of DETR mentioned in \cite{Carion-2020-EOD}. The figure corresponds to Fig.2 in the original paper.}
\label{fig:50} 
\end{figure}

In \cite{Carion-2020-EOD}, a novel framework called DEtection TRansformer (DETR) is proposed to achieve end-to-end object detection. The main architecture of DETR consists of a CNN backbone for feature extraction, an encoder-decoder transformer and a detection prediction network based on feed forward network. The specific framework of DETR is shown in the Fig.\ref{fig:50}. Experiment results on COCO indicate that the results of DETR and Faster R-CNN are comparable. It means DETR has the potential to be employed in microorganism detection.
\par
\begin{figure}[http]
\centering
\includegraphics[width=0.95\linewidth]{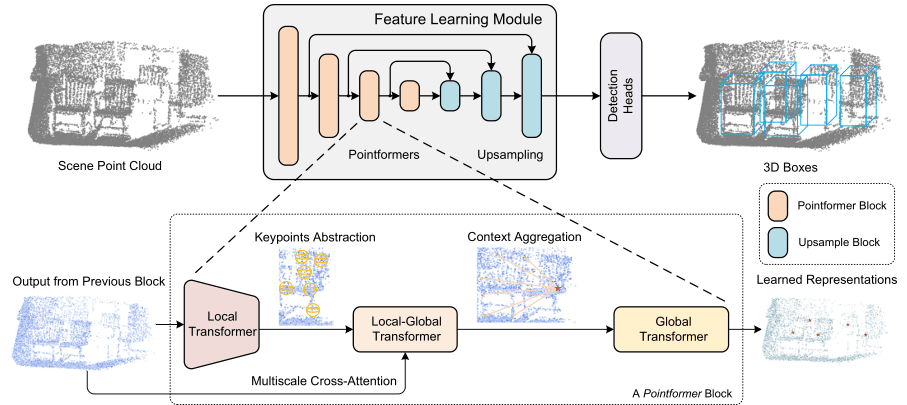}
\caption{The architecture of mentioned in \cite{Pan-2020-3OD}. The figure corresponds to Fig.2 in the original paper.}
\label{fig:51} 
\end{figure}

In \cite{Pan-2020-3OD}, a Transformer backbone designed for 3-D point clouds called Pointformer is presented, which has the ability of learning features effectively. Fig.\ref{fig:51} shows the backbone of Pointformer, which is consisted of a Local Transformer, a Local-Global Transformer and a Global Transformer. The experiment result of detection models with Pointformer show significant performance improvements. This suggests that with the assistance of Pointformer, the performance of 3-D microorganism detection is possible to improve.
\par
\begin{figure}[http]
\centering
\includegraphics[width=0.95\linewidth]{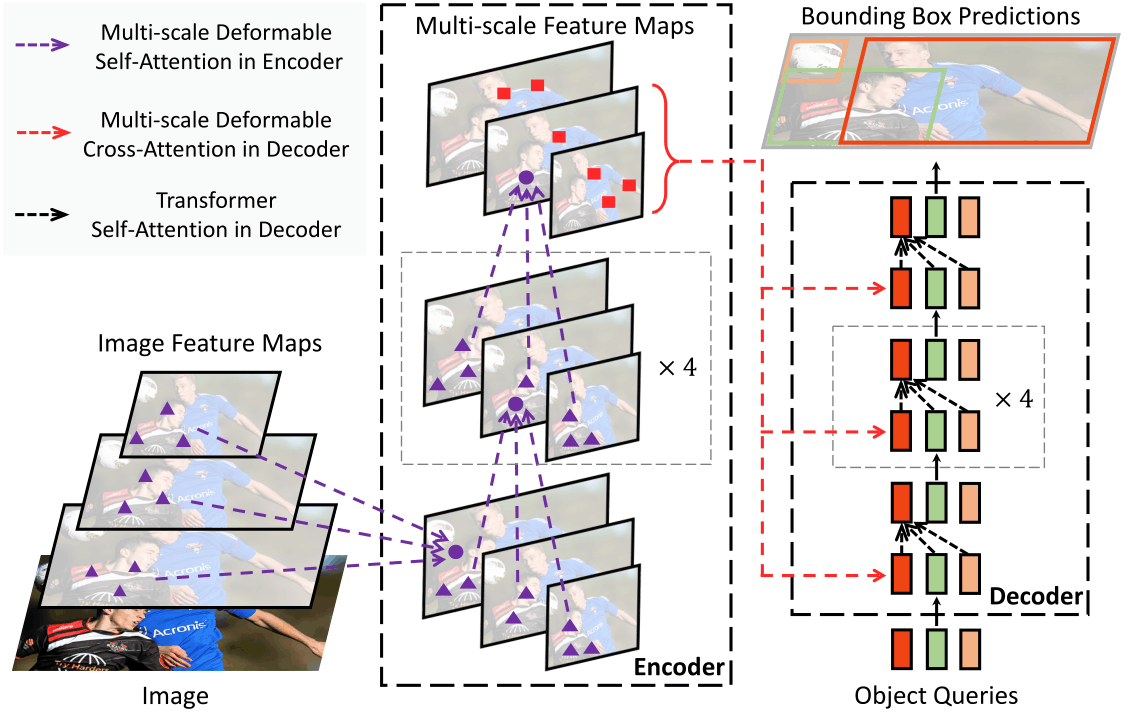}
\caption{The architecture of Deformable DETR mentioned in \cite{Zhu-2021-DDD}. The figure corresponds to Fig.1 in the original paper.}
\label{fig:53} 
\end{figure}

In \cite{Zhu-2021-DDD}, to overcome the limitations of DETR's slow convergence and limited feature spatial resolution when processing image features, a Deformable DETR is presented. The framework of Deformable DETR is shown in Fig.\ref{fig:53}. As an end-to-end object detector, the attention modules of Deformable DETR only focuses on a small set of key sampling points. This makes Deformable DETR more efficient than DETR. Experiment results on the COCO indicate that Deformable DETR is reliable in object detection. Therefore, DETR is able to be applied in microorganism detection with good performance.	
\par
\begin{figure}[http]
\centering
\includegraphics[width=0.93\linewidth]{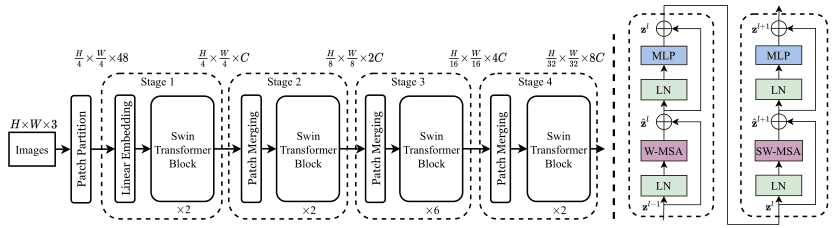}
\caption{ Model architecture of a Swin Transformer (Swin-T) and two Swin Transformer blocks mentioned in \cite{Liu-2021-STH}. The figure corresponds to Fig.3 in the original paper.}
\label{fig:6-2} 
\end{figure}

In \cite{Liu-2021-STH}, a noval transformer named Swin Transformer is designed to better employ transformers applied in language field in computer vision field. Fig.\ref{fig:6-2} illustrates the structure of a Swin Transformer (Swin-T) and two Swin Transformer blocks in detail. By designing a hierarchical feature calculated by Shifted windows, Swin Transformer can efficiently solve the problem of large size and high resolution of images. Experiments show that this method achieves excellent detection results on the COCO test-dev set. In \cite{Yang-2021-FSF}, a better transformer called Focal Transformer is proposed, which has similar structure to Swin Transformer. Unlike Swin Transformer, Focal Transformer replaces self-attention with Focal self-attention. This enables the Focal Transformer to achieve as many attention regions as the Swim Transformer at a lower cost. In addition, Focal self-attention can combine local information and global information more efficiently. Comparative experiments show that Focal Transformer can achieve better detection results than Swin Transformer. 
\begin{figure}[http]
\centering
\includegraphics[width=0.93\linewidth]{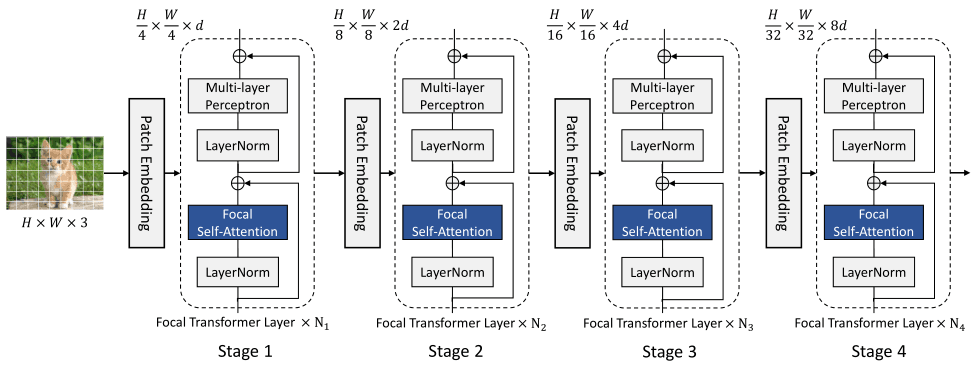}
\caption{ The architecture of Focal Transformer mentioned in \cite{Yang-2021-FSF}. The figure corresponds to Fig.2 in the original paper.}
\label{fig:6-3} 
\end{figure}
\par
\begin{figure}[http]
\centering
\includegraphics[width=0.9\linewidth]{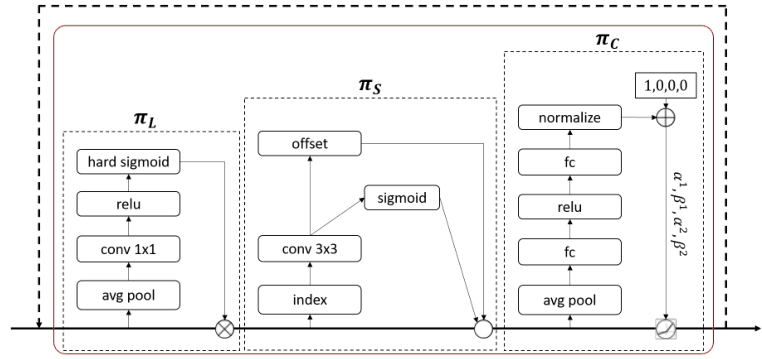}
\caption{ The detailed implementation of dynamic head mentioned in \cite{Dai-2021-DHU}. The figure corresponds to Fig.2 in the original paper.}
\label{fig:6-1} 
\end{figure}

In \cite{Dai-2021-DHU}, for unifying object detection heads with attentions, a dynamic head framework based on combining scale-aware, spatial-aware, and task-aware attentions is proposed. The detailed implementation of dynamic head can be seen in Fig.\ref{fig:6-1}. As a network module, the dynamic head can be easily combined with existing detection models to obtain better detection results. 
\par
\subsubsection{Other Potential Methods}
\label{Sect:6-6-2}
For each related methods, a brief introduction is firstly made and the feasibility of used in microorganism detection is then analysed. Finally, a summary table of methods mentioned in this subsection is presented in Tab.\ref{tab:7}, where the publish date, reference links, categories, main methods and potential contribution of relate works are displayed.
\par
In \cite{Dai-2016-ROD}, an accurate and reliable object detection model is proposed, which is a region-based, fully convolutional networks (R-FCN). R-FCN is an improved design on the basis of ResNet-101. Compared to previous region-based detectors, R-FCN is fully convolutional, which means all required computation is shared on the entire image. The proposed method is tested on the PASCAL VOC 2007 datasets with 83.6$\%$ mAP. In addition, R-FCN requires far less detection time than Faster R-CNN. It means R-FCN has the potential to reduce the time of microorganism detection. 
\par
\begin{figure}[http]
\centering
\includegraphics[width=0.95\linewidth]{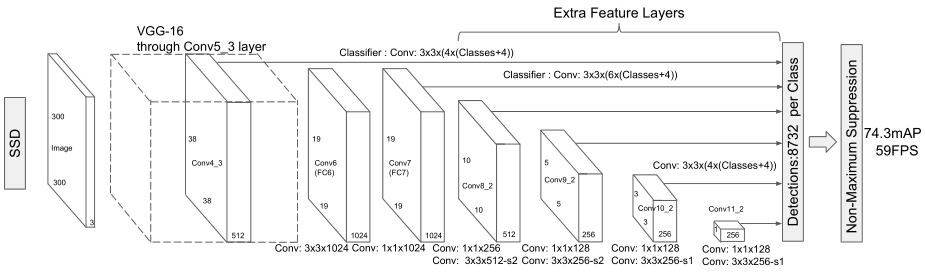}
\caption{The architecture of SSD mentioned in \cite{Liu-2016-SSM}. The figure corresponds to Fig.2.}
\label{fig:46} 
\end{figure}

In \cite{Liu-2016-SSM}, SSD is proposed for object detection. The network architecture of SSD is shown in the Fig.\ref{fig:46}. Compared to Faster R-CNN, the detection efficiency of SSD is higher after eliminating the process of bounding box candidate and feature up-sampling. Experimental result shows that SSD can reduce the detection time under the condition that the detection accuracy is not lower than that of Faster R-CNN. Therefore, SSD is possible to realize high precision and high efficiency detection of microorganisms.
\par
In \cite{Shrivastava-2016-TRO}, an on-line hard example mining (OHEM) algorithm is proposed to train region-based ConvNet detectors, such as Faster R-CNN, Mask R-CNN. The key ides of OHEM to simplify training is automatically selecting hard examples, eliminating several common heuristics and hyper parameters. Result shows that OHEM can effectively improve the convergence of training and the final detection accuracy. With the assistance of OHEM, the performance of existing region-based ConvNet detectors in microorganism detection is able to improve. 
\par
In \cite{Li-2017-FFF}, a novel feature fusion single shot multibox detector (FSSD) on the basic of SSD is prooposed. Compared to SSD, FSSD enables to fuse the features easily from different scales while SSD cannot. In the feature fusion stage, FSSD makes features from different layers with different scales concatenate together. Results on the PASCAL VOC 2007 suggest that the performance of FSSD is not only superior to SSD but also superior to Faster R-CNN and YOLO. FSSD offers the possibility of obtaining better microorganism detection results.
\par
\begin{figure}[http]
\centering
\includegraphics[width=0.95\linewidth]{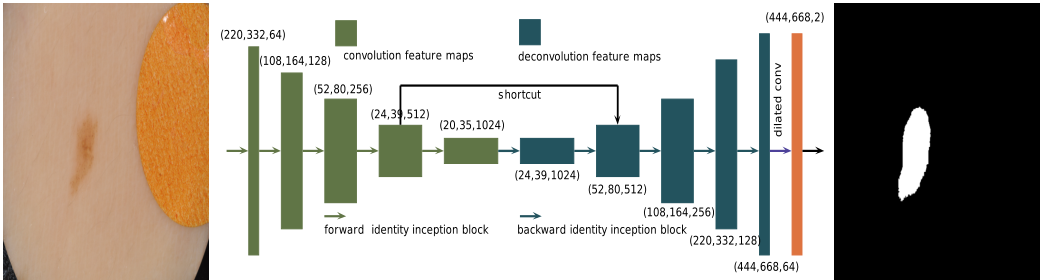}
\caption{The architecture of II-FCN mentioned in \cite{Wen-2017-IF}. The figure corresponds to Fig.2 in the original paper.}
\label{fig:47} 
\end{figure}

In \cite{Wen-2017-IF}, a symmetrical identity inception fully convolution network (II-FCN) is designed to detect Lesion in Dermoscopy image, which is low contrast image. The network architecture of II-FCN is shown in Fig.\ref{fig:47}. Results suggest that II-FCN is able to provide a accurate segmentation result even with low contrast image. It means II-FCN has the potential to improve the performance of microorganism detection for some low contrast image.
\par
In \cite{Coletta-2019-CCA}, a meaningful image detector to detect new classes that is not labelled is designed. The proposed detector is an iterative one by combining SVM and clustering algorithms. Results suggest that the presented detector has the ability of detecting new classes over time using unlabelled instances. Therefore, it provides a new idea in the detection new classes of microorganism.
\par
In \cite{Mittal-2019-AE}, a reliable edge detection algorithm based on multiple threshold approaches (B-Edge) is introduced for overcoming edge connectivity and edge thickness in edge detection. B-Edge mainly consists of three phases: graythresh and threshold computation phase, intensity adjustment phase and grayscale conversion phase. Result shows that B-Edge can detect thin edges with less noise proportion. With the assistance of B-Edge, improving the detection accuracy of microorganism is possible.
\par
In \cite{Sert-2019-BTS}, an approach called neutrosophic set-expert maximum fuzzy-sure entropy (NS-EMFSE) is proposed to for Glioblastoma, which is the most difficult in brain tumor segmentation. Experiment results prove that the performance of NS-EMFSE is batter than others. It means NS-EMFSE has the potential to improve the performance of microorganism detection by improving the accuracy of segmentation.
\par
In \cite{Liu-2020-STS}, an automatic small traffic sign detection method in large traffic scenes is proposed based on a deconvolution region-based-CNN (DR-CNN). Experimental result shows that DR-CNN can well cope with the challenge of detecting small objects in a large scene. It means when the area of the target microorganism is much smaller than the microscopic image in which it is located, the difficulty of microorganism detection can be solved by referring to the idea of DR-CNN.
\par
In \cite{Xu-2020-AEF}, an enhanced framework of generative adversarial networks (EF-GANs) is proposed to deal with image augmentation problem of small datasets. EF-GANs mainly consists three steps: color space augmentation, GANs and image rotation. The experimental results suggest that the classification accuracy of corresponding classifier is improved to different degrees based on EF-GANs data expansion.
This means that it is possible for EF-GANs to solve the problem of too few training samples encountered in microorganism detection, which is also a common problem in this field.
\par
In \cite{Abdel-2020-OED}, an optimized edge detection technique is presented by applying a genetic algorithm. First, the image features are improved by balance contrast enhancement technique. After that, the fine edges is detected by combining proposed genetic algorithm and a appropriate training dataset. Through comparative experiments, the performance of the proposed method in magnetic resonance image edge detection is proved to be superior to the classical edge detection techniques. It means when applying this method in microorganism detection, it is possible to provide a better edge detection result.
\par
In \cite{Nsaif-2021-FCF}, a Faster R-CNN with Gabor filters and naive Bayes (FRCNN-GNB) model is proposed to improve the accuracy of eye detection. Compared to existing methods, the proposed model can solve the problem of occlusion or reflections from glass in eye detection. Faster R-CNN is employed to detect the initial bounding boxes of the eye region. Among this bounding boxes, Gabor filters and the naïve Bayes model are then used for determining which of them belongs to the eye region. Results suggest that the introduced algorithm is able to detect eyes with high accuracy. The idea of this method is possible to improve the accuracy of microorganism detection.
\par
\begin{figure}[htbp]
 \centering
 \subfigure[]{
  \label{fig:6-4-a}
  \includegraphics[scale=0.3]{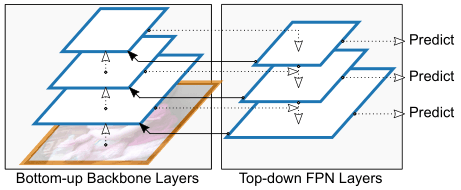}}
 \subfigure[]{
  \label{fig:6-4-b}
  \includegraphics[scale=0.3]{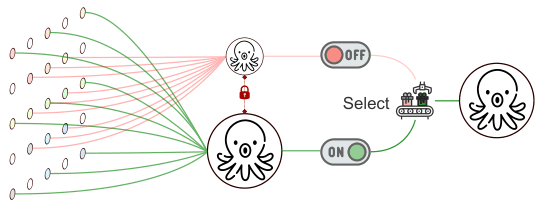}}
   \caption{(a):Recursive Feature Pyramid; (b): Switchable Atrous Convolution. The figure corresponds to Fig.1 in \cite{Qiao-2021-DDO}.}
  \label{fig:6-4}
\end{figure}

In \cite{Qiao-2021-DDO}, DetectoRS, a detector with improvements at both macro and micro levels, is proposed. Recursive Feature Pyramid is employed in macro level for extracting features from images that are more suitable for object detection. In addition, the employing of Switchable Atrous Convolution in micro levels enables DetectoRS to choose the training scale more flexibly and effectively. The structures of Recursive Feature Pyramid and Switchable Atrous Convolution are shown in Fig.\ref{fig:6-4}. Results on COCO dataset indicate that  DetectoRS yields a high box AP of 55.7$\%$ box AP for object detection.
\par
So far, only YOLO in the YOLO series is employed in microorganism detection with good results. Compared with YOLO, YOLO9000 \cite{Redmon-2017-YBF} can predict more accurately and identify more objects on the basis of ensuring that the processing speed does not decrease by applying multi-dataset joint training, new backbone network (DarkNet19) and $k$-means clustering algorithm to generate anchor box. YOLOv3 \cite{Redmon-2018-YAI} achieves a more accurate detection with a slight loss of detection speed by applying multi-level feature fusion and new backbone network (DarkNet53), compared with YOLO9000. YOLOv4 \cite{Bochkovskiy-2020-YOS} not only achieves fast and accurate detection, but also requires less performance on the GPU during training and use. 
In \cite{Ge-2021-YEY}, YOLOX is proposed as an excellent detector. Compared with previous detectors in the YOLO series, YOLOX adopts decoupled head to speed up convergence and increase AP value of detection results. YOLOR is proposed in \cite{Wang-2021-YOL} with a novel improvement perspective of combining explicit knowledge learning with implicit knowledge learning. The experimental results show that the introduction of implicit knowledge learning can improve the detection accuracy by a small margin.
According to the performances of all YOLO series in addition to the original YOLO, it all have the potential to improve the performance of microorganism detection. 
\par
\begin{landscape} 
\begin{longtable}{c|c|c|c|c}

\caption{Summary of potential other methods for microorganism detection.}\\
\hline\hline
\endfirsthead
\caption{Continued}\\
\hline\hline
\endhead
\hline

Date & Refs & Category  & Method  & Potential Contribution \\ \hline
  
  2020   &  \cite{Xu-2020-AEF} & Data Augmentation & EF-GANs & Extending data set  \\\hline

   2019  & \cite{Mittal-2019-AE} & \multirow{4}{*}{\begin{tabular}[c]{@{}c@{}} \\ \\Image\\ Segmentation\end{tabular}} &B-Edge & Providing a better edge detection  \\
 2019  &   \cite{Sert-2019-BTS}   &        &   NS-EMFSE      & Improving accuracy of segmentation  \\
2020	&  \cite{Abdel-2020-OED}&    & \makecell[c]{Genetic algorithm \\based edge detection}   & \makecell[c]{Providing a better \\edge detection} \\ \hline
2016  & \cite{Dai-2016-ROD} &\multirow{10}{*}{\makecell[c]{\\ \\ \\ \\ \\ \\ \\Detector} }  & R-FCN  & Reducing time \\ 
    2016 &\cite{Liu-2016-SSM} &      & SSD & Improving precision and efficiency \\
    2017 & \cite{Li-2017-FFF} &      & FSSD & Improving precision and efficiency \\
2017 & \cite{Wen-2017-IF} &      & II-FCN & \makecell[c]{Detecting objects in low \\contrast image} \\
2017 & \cite{Redmon-2017-YBF} &      & YOLO9000 & Improving accuracy \\
2018 & \cite{Redmon-2018-YAI} &      & YOLOv3 & Improving accuracy \\
2019 & \cite{Coletta-2019-CCA}&      & An iterative detector & Detecting new classes \\ 
2020  & \cite{Liu-2020-STS} &  & DR-CNN & \makecell[c]{Detecting small object in \\a large scene} \\
2020  & \cite{Bochkovskiy-2020-YOS} & & YOLOv4 &Improving precision and efficiency  \\ 
2021  & \cite{Nsaif-2021-FCF}&  & FRCNN-GNB & Improving accuracy \\
2021 & \cite{Qiao-2021-DDO}&      & DetectoRS & \makecell[c]{Providing more efficient feature \\information} \\ 
2021& \cite{Ge-2021-YEY}&      &  YOLOX & Speeding up convergence \\ 
2021& \cite{Wang-2021-YOL}&      & YOLOR & \makecell[c]{Providing more efficient feature \\information}
                      
\label{tab:7}  
\end{longtable}
\end{landscape} 
\par
\subsection{Potential Application Fields}
\label{Sect:6-3}
After an in-depth study of microorganism detection methods, the reviewed methods can be applied to some other fields, such as ecosystem monitoring \cite{Coppin-2004-RAC}, on-road vehicle detection \cite{Sun-2006-OVD}, fabric defect detection\cite{Ngan-2011-AFD}, crack detection \cite{Mohan-2018-CDU}, forest fire detection \cite{Alkhatib-2014-ARO}, pedestrian detection \cite{Enzweiler-2008-MPD}, object detection in optical remote sensing images \cite{Cheng-2016-ASO}, remote sensing digital image analysis \cite{Richards-1999-RSD}, histopathological image analysis \cite{Ai-2021-ASR,Chen-2022-IAI,Li-2022-ACR,Li-2021-ACRO,Xue-2020-APO,Zhou-2020-ACR,Li-2020-ARF,Li-2019-CHI,Sun-2020-GHI}, cytopathological image analysis \cite{Rahaman-2021-DAD,Rahaman-2020-ASF,Li-2017-FCA}, video analysis \cite{Chen-2022-SDA,Li-2020-FFF,Shen-2015-ITO} and COVID-19 image analysis \cite{Rahaman-2020-IOC,Li-2020-ASM}. In general, the analyzed methods presented in this article provide new research ideas and better detection results for many other research fields involving detection.
\par

%% file: Conclusion.tex
\section{Conclusion and Future Work}
\label{Sect:7}
This review presents a state-of-the-art survey for object detection technologies in microorganism image analysis: from classical image processing and traditional machine learning to current deep learning and potential visual transformer methods. First, we introduce the basic information of microorganism detection in Sect.\ref{Sect:1}, including research motivation and research status. Second, we outline the development of early and current stages of microorganism detection methods in chronological order, including standard evaluation criteria in Sect.\ref{Sect:2}.
Third, we have summarized related works in classical image processing, traditional machine learning and deep learning-based methods in Sect.\ref{Sect:3}, Sect.\ref{Sect:4} and Sect.\ref{Sect:5}, respectively. For each category of detection methods, we present the relevant research in chronological order. For some of the outstanding research, we show its flowchart or result diagram to better understand. 
Finally, Sect.\ref{Sect:6} is the method analysis part, where we analyze all the three categories of methods mentioned above and further analyze the advantages and disadvantages of some of the methods based on performance. Potential methods in microorganism detection are also introduced. Among them, methods based on visual transformer show great potential in microorganism detection.
\par
Based on the analysis of the development of microorganism detection methods, the future development trend and challenges are predicted. The most promising development in this field can be the combination of existing detection models and visual transformer. Due to the successful applications of transformer in natural language processing, many researchers are trying to apply visual transformer to multiple computer vision tasks. Therefore, methods by combining visual transformer and others have the potential to yield good performance in the field of microorganism detection. One of the challenges can be real-time detection. With the development of related technologies, the requirements for detection speed are getting higher and higher. In many aspects such as microbial tracking, pathogen detection and wild microbial monitoring, the detection speed will significantly affect its practical application. Moreover, the difficulty of obtaining good quality microorganism data and high processing costs leads to obtaining insufficient dataset with poor quality, which encounters the experimental results. The problem of insufficient data sets and poor quality is a major challenge for microorganism detection.
However, the Environmental Microorganism Image Dataset Seventh Version (EMDS-7) is available publicly \cite{Yang-2021-EEM}. This dataset has 41 types of environmental microorganisms. As a multi-object dataset, it has 13216  objects in 2365 images. Furthermore, all 2365 images have corresponding labeling files in ``.XML". Considering the huge development prospect in the field of microorganism detection, it is believed that more and more teams will produce high quality databases like EMDS-7.

\par